\documentclass[10pt,twocolumn,letterpaper]{article}

\usepackage[pagenumbers]{cvpr}

\usepackage{graphicx}
\usepackage{amsmath}
\usepackage{amssymb}
\usepackage{booktabs}

\usepackage[pagebackref,breaklinks,colorlinks]{hyperref}

\usepackage[utf8]{inputenc} 
\usepackage[T1]{fontenc}    
\usepackage{algorithmic}
\usepackage{algorithm}
\usepackage[table]{xcolor}
\usepackage{wrapfig}
\usepackage{tabularx}
\usepackage{multicol}
\usepackage{multirow}
\usepackage{caption}
\usepackage{marvosym}

\usepackage{listings}

\usepackage[capitalize]{cleveref}
\crefname{section}{Sec.}{Secs.}
\Crefname{section}{Section}{Sections}
\Crefname{table}{Table}{Tables}
\crefname{table}{Tab.}{Tabs.}

\definecolor{forestgreen}{rgb}{0.133, 0.545, 0.133}
\definecolor{yellowyellow}{rgb}{0.133, 0.545, 0.133}

\definecolor{correct}{RGB}{173, 173, 173}
\definecolor{incorrect}{RGB}{192, 0, 0}

\newlength\savewidth

\usepackage{xspace}
\usepackage{amsmath}

\newcommand*{\lidar}{LiDAR\@\xspace}
\newcommand*{\name}{LaserMix}

\newcommand*{\Xcrop}{X_\mathrm{in}}
\newcommand*{\Ycrop}{Y_\mathrm{in}}
\newcommand*{\Xcomp}{X_\mathrm{out}}

\newcommand*{\ev}{\mathbb{E}}
\newcommand*{\evemp}{\hat{\mathbb{E}}}
\newcommand*{\Pemp}{P}
\newcommand*{\Hemp}{\hat{H}}

\usepackage{setspace}
\definecolor{mygray}{gray}{0.5}

\newcommand\blfootnote[1]{%
\begingroup
\renewcommand\thefootnote{}{}\footnote{#1}%
\addtocounter{footnote}{-1}%
\endgroup
}

\usepackage[accsupp]{axessibility}  

\def\confName{CVPR}
\def\confYear{2023}

\begin{document}

\title{LaserMix for Semi-Supervised LiDAR Semantic Segmentation}

\author{Lingdong Kong$^{1,2,3,*}$ \quad Jiawei Ren$^{1,*}$ \quad Liang Pan$^{1}$ \quad Ziwei Liu$^{1,\textrm{\Letter}}$
\\[1ex]
{$^{1}$S-Lab, Nanyang Technological University} ~~ {$^{2}$National University of Singapore} ~~ {$^{3}$CNRS@CREATE}
\\
{\tt\small \{lingdong001,jiawei011\}@e.ntu.edu.sg} \quad {\tt\small \{liang.pan,ziwei.liu\}@ntu.edu.sg}
}

\twocolumn[{%
\renewcommand\twocolumn[1][]{#1}%
\maketitle
\begin{center}
    \centering
    \vspace{-8pt}
    \captionsetup{type=figure}
    \includegraphics[width=\textwidth]{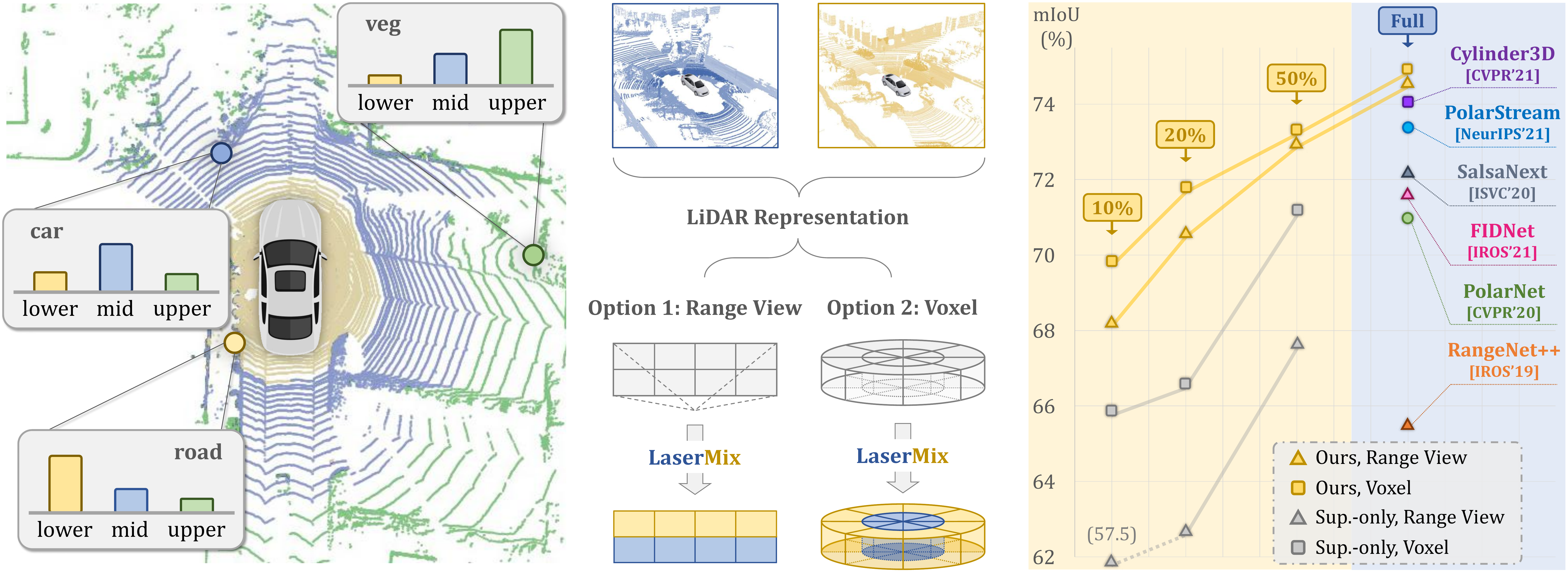}
    \vspace{-14pt}
    \captionof{figure}{\textbf{Left:} The LiDAR point cloud contains strong spatial prior. Objects and backgrounds around the ego-vehicle have a patterned distribution on different (lower, middle, upper) laser beams. \textbf{Middle:} Following the scene structure, the proposed LaserMix blends beams from different LiDAR scans, which is compatible with various popular LiDAR representations. \textbf{Right:} We achieve superior results over SoTA methods in both low-data ($10\%$, $20\%$, and $50\%$ semantic labels) and high-data (full semantic labels) regimes on nuScenes~\cite{Panoptic-nuScenes}.}
    \label{fig:LaserMix}
    \vspace{6.26pt}
\end{center}
}]

\blfootnote{${(*)}$~Lingdong and Jiawei contributed equally to this work. ${(\textrm{\Letter})}$~Ziwei serves as the corresponding author. E-mail: \texttt{ziwei.liu@ntu.edu.sg}.}

\begin{abstract}
    \vspace{-0.125cm}
   Densely annotating LiDAR point clouds is costly, which often restrains the scalability of fully-supervised learning methods. In this work, we study the underexplored semi-supervised learning (SSL) in LiDAR semantic segmentation. Our core idea is to leverage the strong spatial cues of LiDAR point clouds to better exploit unlabeled data. We propose \textbf{LaserMix} to mix laser beams from different LiDAR scans and then encourage the model to make consistent and confident predictions before and after mixing. Our framework has three appealing properties. \textbf{1) Generic:} LaserMix is agnostic to LiDAR representations (\textit{e.g.}, range view and voxel), and hence our SSL framework can be universally applied. \textbf{2) Statistically grounded:} We provide a detailed analysis to theoretically explain the applicability of the proposed framework. \textbf{3) Effective:} Comprehensive experimental analysis on popular LiDAR segmentation datasets (nuScenes, SemanticKITTI, and ScribbleKITTI) demonstrates our effectiveness and superiority. Notably, we achieve competitive results over fully-supervised counterparts with $2\times$ to $5\times$ fewer labels and improve the supervised-only baseline significantly by relatively $10.8\%$. We hope this concise yet high-performing framework could facilitate future research in semi-supervised LiDAR segmentation. Code is publicly available\footnote{\url{https://github.com/ldkong1205/LaserMix}.}.
   \vspace{-0.35cm}
\end{abstract}

\section{Introduction}
\label{sec:introduction}

LiDAR segmentation is one of the fundamental tasks in autonomous driving perception~\cite{Survey-LiDAR}. It enables autonomous vehicles to semantically perceive the dense 3D structure of the surrounding scenes~\cite{nunes2022segcontrast, Autonomous-Driving,Geiger2012CVPR}. However, densely annotating LiDAR point clouds is inevitably expensive and labor-intensive~\cite{ScribbleKITTI,SQN,ConDA}, which restrains the scalability of fully-supervised \lidar segmentation methods. Semi-supervised learning (SSL) that directly leverages the easy-to-acquire unlabeled data is hence a viable and promising solution to achieve scalable LiDAR segmentation \cite{Survey-LiDAR-Data-Hungry,gebrehiwot2022teachers}.

Yet, semi-supervised \lidar segmentation is still underexplored. Modern SSL frameworks are mainly designed for image recognition~\cite{MixMatch, FixMatch, ReMixMatch} and semantic segmentation~\cite{CCT,GCT,CPS} tasks, which only yield sub-par performance on \lidar data due to the large modality gap between 2D and 3D. Recent research~\cite{GPC} proposed to consider semi-supervised point cloud semantic segmentation as a fresh task and proposed a point contrastive learning framework. However, their solutions do not differentiate indoor and outdoor scenes and therefore overlook the intrinsic and important properties that only exist in \lidar point clouds.

In this work, we explore the use of spatial prior for semi-supervised \lidar segmentation. Unlike the general 2D/3D segmentation tasks, the spatial cues are especially significant in \lidar data. In fact, LiDAR point clouds serve as a perfect reflection of real-world distributions, which is highly dependent on the spatial areas in the \lidar-centered 3D coordinates. As shown in \cref{fig:LaserMix}~(left), the top laser beams travel outward long distance and perceive mostly \textit{vegetation}, while the middle and bottom beams tend to detect \textit{car} and \textit{road} from the medium and close distances, respectively.
To effectively leverage this strong spatial prior, we propose \textbf{LaserMix} to mix laser beams from different LiDAR scans, and then encourage the LiDAR segmentation model to make consistent and confident predictions before and after mixing. Our SSL framework is statistically grounded, which consists of the following components:

\noindent\textbf{\textit{1)}} Partitioning the \lidar scan into low-variation areas. We observe a strong distribution pattern on laser beams as shown in \cref{fig:LaserMix}~(left) and thus propose the laser partition.

\noindent\textbf{\textit{2)}} Efficiently mixing every area in the scan with foreign data and obtaining model predictions. We propose LaserMix to manipulate the laser-grouped areas from two \lidar scans in an intertwining way as depicted in \cref{fig:LaserMix}~(middle) and serves as an efficient \lidar mixing strategy for SSL.

\noindent\textbf{\textit{3)}} Encouraging models to make confident and consistent predictions on the same area in different mixing. We hence propose a mixing-based teacher-student training pipeline.

Despite the simplicity of our overall pipeline, it achieves competitive results over the fully supervised counterpart using $2\times$ to $5\times$ fewer labels as shown in \cref{fig:LaserMix}~(right) and significantly outperforms all prevailing semi-supervised segmentation methods on nuScenes~\cite{Panoptic-nuScenes} (up to $+5.7\%$ mIoU) and SemanticKITTI~\cite{SemanticKITTI} (up to $+3.5\%$ mIoU). Moreover, \name{} directly operates on point clouds so as to be agnostic to different \lidar representations, \eg, range view~\cite{RangeNet++} and voxel~\cite{Cylinder3D}. Therefore, our pipeline is highly compatible with existing state-of-the-art (SoTA) \lidar segmentation methods under various representations~\cite{PolarNet,KPConv,FIDNet}. Besides, our pipeline achieves competitive performance using very limited annotations on weak supervision dataset \cite{ScribbleKITTI}: it achieves $54.4\%$ mIoU on SemanticKITTI~\cite{SemanticKITTI} using only $0.8\%$ labels, which is on-par with PolarNet~\cite{PolarNet} ($54.3\%$), RandLA-Net~\cite{RandLa-Net} ($53.9\%$), and RangeNet++~\cite{RangeNet++} ($52.2\%$) using $100\%$ labels. Spatial prior is proven to play a pivotal role in the success of our framework through comprehensive empirical analysis.
To summarize, this work has the following key contributions:
\begin{itemize}
    \vspace{-0.18cm}
    \item We present a statistically grounded SSL framework that effectively leverages the spatial cues in \lidar data to facilitate learning with semi-supervisions.
    \vspace{-0.18cm}
    \item  We propose LaserMix, a novel and representation-agnostic mixing technique that strives to maximize the ``strength'' of the spatial cues in our SSL framework.
    \vspace{-0.18cm}
    \item  Our overall pipeline significantly outperforms previous SoTA methods in both low- and high-data regimes. We hope this work could lay a solid foundation for semi-supervised LiDAR segmentation.
    \vspace{-0.18cm}
\end{itemize}

\section{Related Work}
\label{sec:related_work}

\noindent\textbf{LiDAR Segmentation}. Various approaches from different aspects have been proposed for LiDAR scene segmentation, \ie, range view \cite{RangeNet++,SqueezeSegV3,SalsaNext,FIDNet}, bird's eye view \cite{PolarNet}, voxel \cite{SPVNAS,Cylinder3D}, and multi-view \cite{AMVNet,RPVNet} methods. Although appealing results have been achieved, these fully-supervised methods rely on large-scale annotated LiDAR datasets and their performance would degrade severely in the low-data regime \cite{Survey-LiDAR-Data-Hungry}. Recent works seek weak \cite{SQN,PSD}, scribble \cite{ScribbleKITTI}, and box \cite{Box2Seg} supervisions or activate learning \cite{LiDAR-SSL-2019,LESS} to ease the annotation cost. We tackle this problem from the perspective of semi-supervised learning (SSL), aiming at directly leveraging the easy-to-acquire unlabeled data to boost the LiDAR semantic segmentation performance.

\noindent\textbf{SSL in 2D}. Well-known SSL algorithms are prevailing in handling image recognition problems \cite{Pi-Model,MeanTeacher,MixMatch,ReMixMatch,FixMatch}. In the context of semantic segmentation, CutMix-Seg \cite{CutMix-Seg} and PseudoSeg \cite{PseudoSeg} apply perturbations on inputs and hope the decision boundary lies in the low-density region. CPS \cite{CPS} and GCT \cite{GCT} enforce consistency between two perturbed networks \cite{PS-MT}. These perturbations \cite{Simple-Baseline,ClassMix,Strong-Weak-Net}, however, are either inapplicable or only yield sub-par results in 3D. Another line of research is entropy minimization. Methods like CBST \cite{CBST} and ST++ \cite{ST++} generate pseudo-labels \cite{lee2013pseudo} offline per round during self-training. The extra storage needed might become costly for large-scale LiDAR datasets \cite{nuScenes,SemanticKITTI,Panoptic-nuScenes}. Our framework encourages both consistency regularization and entropy minimization and does not require extra overhead, which better maintains scalability.

\noindent\textbf{SSL in 3D}. Most works focus on developing SSL for object-centric point clouds \cite{SRN,SSL-ShapeSeg} or indoor scenes \cite{Superpoint-SemiSeg,SSPC-Net,nunes2022segcontrast,xie2021exploring}, whose scale and diversity are much lower than the outdoor LiDAR point clouds \cite{nuScenes,SemanticKITTI}. Some other works \cite{TGNN,DetMatch,Offboard-3D-OD} try to utilize SSL for 3D object detection on LiDAR data. A recent work~\cite{GPC} tackles semi-supervised point cloud semantic segmentation using contrastive learning, but it still mainly focused on indoor scenes and does not distinguish between the uniformly distributed indoor point clouds and the spatially structured LiDAR point clouds. We are one of the first works to explore SSL for LiDAR segmentation. Our work also establishes comprehensive SSL benchmarks upon popular autonomous driving databases~\cite{Panoptic-nuScenes,SemanticKITTI,ScribbleKITTI}.

\section{Approach}
\label{sec:methods}

In this section, we first introduce our SSL framework that leverages the spatial prior of \lidar data by encouraging confidence and consistency in predictions (\cref{sec:ssl-framework}). We then present LaserMix which strives to maximize the ``strength'' of the spatial prior and mixes \lidar scans in an efficient manner (\cref{sec:lasermix}). Finally, we elaborate on the overall pipeline (\cref{sec:pipeline}) and the pseudo-code (Algo.~\ref{alg:lasermix}).

\subsection{Leveraging the Spatial Prior for SSL}
\label{sec:ssl-framework}

\noindent\textbf{Spatial Prior Formulation}. The distribution of real-world objects/backgrounds has a strong correlation to their spatial positions in the \lidar scan. Objects/backgrounds inside a specified spatial area of a \lidar point cloud follow similar patterns, \eg, the close-range area is most likely \textit{road} while the long-range area consists of \textit{building}, \textit{vegetation}, etc. In another word, there exists a spatial area $a \in A$ where LiDAR points and semantic labels inside the area (denoted as $\Xcrop$ and $\Ycrop$, respectively) will have relatively low variations. Formally, the conditional entropy $H(\Xcrop, \Ycrop|A)$ is smaller. Therefore, when estimating the parameter $\theta$ of the segmentation network $\mathcal{G}_{\theta}$, we would expect:
\begin{align}\label{eq:entropy}
    \ev_\theta[H(\Xcrop, \Ycrop|A)] =c\,,
\end{align}
where $c$ is a small constant. Similar to the classic entropy minimization \cite{grandvalet2004semi}, the constraint in \cref{eq:entropy} can be converted to a prior on the model parameter $\theta$ using the principle of maximum entropy:
\begin{equation}
\begin{split}
\label{eq:structure_prior}
    P(\theta) &\propto \exp(-\lambda H(\Xcrop, \Ycrop|A)) \\ &\propto \exp(-\lambda H(\Ycrop|\Xcrop, A))\,,
\end{split}
\end{equation}
where $\lambda>0$ is the Lagrange multiplier corresponding to constant $c$; $H(\Xcrop| A)$ has been ignored for being independent of the model parameter $\theta$. We consider \cref{eq:structure_prior} as the formal formulation of the spatial prior and discuss how to empirically compute it in the following sections.

\noindent\textbf{Marginalization}.
To utilize the spatial prior defined in \cref{eq:structure_prior}, we empirically compute the entropy $H(\Ycrop|\Xcrop, A)$ of the LiDAR points \textit{inside} area $A$ as follows:
\begin{equation}
\begin{split}\label{eq:empirical-entropy}
    &\Hemp(\Ycrop|\Xcrop, A) =\\ &\evemp_{\Xcrop, \Ycrop, A}[\Pemp(\Ycrop|\Xcrop, A)\log \Pemp(\Ycrop|\Xcrop, A)]\,,
\end{split}
\end{equation}
where $\hat{.}$ denotes the empirical estimation. The end-to-end LiDAR segmentation model $\mathcal{G}_{\theta}$ usually takes full-sized data as inputs during inference. Therefore, to compute $\Pemp(\Ycrop|\Xcrop, A)$ in \cref{eq:empirical-entropy}, we first pad the data \textit{outside} the area to obtain the full-sized data. Here we denote the data \textit{outside} the area as $\Xcomp$; we then let the model infer $P(\Ycrop|\Xcrop, \Xcomp, A)$, and finally marginalize $\Xcomp$ as:
\begin{align}\label{eq:marginalization}
    \Pemp(\Ycrop|\Xcrop, A) = \evemp_{\Xcomp}[\Pemp(\Ycrop|\Xcrop, \Xcomp, A)]\,.
\end{align}
The generative distribution of the padding $P(\Xcomp)$ can be directly obtained from the dataset.

\noindent\textbf{Training}.
Finally, we train the segmentation model $\mathcal{G}_{\theta}$ using the standard maximum-a-posteriori (MAP) estimation. We maximize the posterior that can be computed by \cref{eq:structure_prior}, \cref{eq:empirical-entropy} and \cref{eq:marginalization}, which is formulated as follows:
\vspace{-0.05cm}
\begin{equation}
\label{eq:our-ssl}
\begin{split}
    &C(\theta) = L(\theta) - \lambda \Hemp(\Ycrop|\Xcrop, A) 
                = L(\theta) \\ &- \lambda\evemp_{\Xcrop, \Ycrop, A}[\Pemp(\Ycrop|\Xcrop, A)\log \Pemp(\Ycrop|\Xcrop, A)].
\end{split}
\end{equation}
Here, $L(\theta)$ is the likelihood function computed using labeled data, \ie, the conventional supervised learning. Minimizing $\Hemp(\Ycrop|\Xcrop, A)$ requires the marginal probability $P(\Ycrop|\Xcrop, A)$ to be confident, which further requires $P(\Ycrop|\Xcrop, \Xcomp, A)$ to be both confident and consistent with respect to different outside data $\Xcomp$.

In summary, our proposed SSL framework in \cref{eq:our-ssl} encourages the segmentation model to make confident and consistent predictions at a predefined area, regardless of the data outside the area. The predefined area set $A$ determines the ``strength'' of the prior. When setting $A$ to the full area (\ie, the whole point cloud), our framework degrades to the classic entropy minimization framework \cite{grandvalet2004semi}.

\noindent\textbf{Implementation}.
There are three key steps for implementing our framework: 
\begin{itemize}
    \vspace{-0.12cm}
    \item \textit{Step 1)}: Select a proper partition set $A$ which maintains strong spatial prior;
    \vspace{-0.12cm}
    \item \textit{Step 2)}: Efficiently compute the marginal probability, \ie, $\Pemp(\Ycrop|\Xcrop, A)$;
    \vspace{-0.12cm}
    \item \textit{Step 3)}: Efficiently minimize the marginal entropy, \ie, $\Hemp(\Ycrop|\Xcrop, A)$.
    \vspace{-0.12cm}
\end{itemize}
We propose a simple yet effective implementation following these steps in the next section.

\subsection{LaserMix}
\label{sec:lasermix}

\begin{figure}
    \begin{center}
    \includegraphics[width=0.98\linewidth]{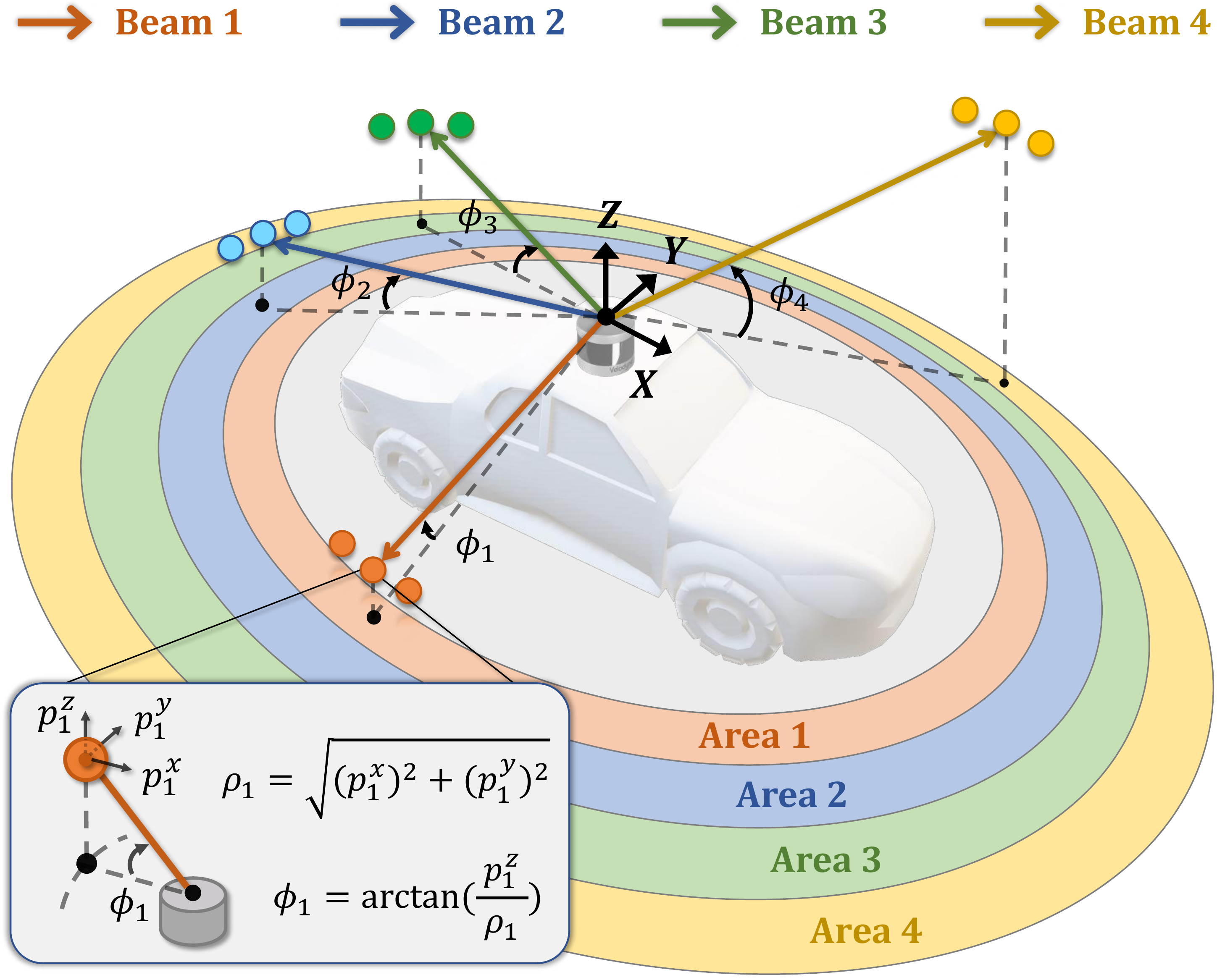}
    \end{center}
    \vspace{-0.5cm}
    \caption{\textbf{Laser partition example}. We group LiDAR points $(p^x_i, p^y_i, p^z_i)$ whose inclinations $\phi_i$ are within the same inclination range into the same area, as depicted in the color regions.}
    \vspace{-0.1cm}
    \label{fig:inclination}
\end{figure}

\noindent\textbf{Partition}.
LiDAR sensors have a fixed number (\eg, $32$, $64$, and $128$) of laser beams which are emitted isotropically around the ego-vehicle with predefined inclination angles as shown in \cref{fig:inclination}. To obtain a proper set of spatial areas $A$, we propose to partition the LiDAR point cloud based on laser beams. Specifically, points captured by the same laser beam have a unified inclination angle to the sensor plane. For point $i$, its inclination $\phi_i$ is defined as follows:
\begin{align}
   \phi_i = \mathrm{arctan}(\frac{p^z_i}{\sqrt{(p^x_i)^2 + (p^y_i)^2 }})\,,
\end{align}
where $(p^x, p^y, p^z)$ is the Cartesian coordinates of the \lidar points.
Given two \lidar scans $x_1$ and $x_2$, we first group all points from each scan by their inclination angles. Concretely, to form $m$ non-overlapping areas, a set of $m+1$ inclination angles $\Phi=\{\phi_0, \phi_1, \phi_2, ..., \phi_m\}$ will be evenly sampled within the range of the minimum and maximum inclination angles in the dataset (defined by sensor configurations), and the area set $A=\{a_1, a_2, ..., a_m\}$ can be formed by bounding area $a_i$ in the inclination range $[\phi_{i-1}, \phi_i)$.

\textit{Role in our framework:} Laser partition effectively ``excites'' a strong spatial prior in the \lidar point cloud, as described by \textit{Step 1} in our framework. As shown in \cref{fig:LaserMix}~(left), we find an overt pattern in semantic classes detected by each laser beam. More concrete evidence on this aspect has been included in the Appendix. Despite being an empirical choice, we will show in later sections that laser partition significantly outperforms other partition choices, including random points (\textit{MixUp}-like partition~\cite{MixUp}), random areas (\textit{CutMix}-like partition~\cite{CutMix}), and other heuristics like azimuth $\alpha$ (sensor horizontal direction) or radius $r$ (sensor range direction) partitions.

\noindent\textbf{Mixing}. To this end, we propose LaserMix, a simple yet effective \lidar mixing strategy that can better control the ``strength'' of the spatial prior. LaserMix mixes the aforementioned laser partitioned areas $A$ from two scans in an intertwining way, \ie, one takes from odd-indexed areas $A_1=\{a_1, a_3, ...\}$ and the other takes from even-indexed areas $A_2=\{a_2, a_4, ...\}$, so that each area's neighbor will be from the other scan:
\begin{equation}
\begin{split}
\tilde{x}_1, \tilde{x}_2 = \textrm{LaserMix}(x_1, x_2)\,, \\
\tilde{x}_1 = x_1^{a_1} \cup x_2^{a_2} \cup x_1^{a_3} \cup\cdots, \\
\tilde{x}_2 = x_2^{a_1} \cup x_1^{a_2} \cup x_2^{a_3} \cup\cdots, \\
\end{split}
\end{equation}
where $x_i^{a_j}$ is the data crop of $x_i$ in the area $a_j$. The semantic labels are mixed in the same way. 
LaserMix is directly applied to the point clouds and is thus agnostic to the various LiDAR representations~\cite{RandLa-Net,RangeNet++,PolarNet,Cylinder3D}. We show LaserMix's instantiations with the \textit{range view} and \textit{voxel} representations as in \cref{fig:LaserMix}~(middle), since they are currently the  most efficient and the best-performing options, respectively.

\textit{Role in our framework:} LaserMix helps to efficiently compute the marginal probability $P(\Ycrop|\Xcrop, A)$, as described by \textit{Step 2} in our framework. The cost for directly computing the marginal probability in \cref{eq:marginalization} on real-world LiDAR data is prohibitive; we need to iterate through all areas in $A$ and all outside data in $\Xcomp$, which requires $|A|\cdot|\Xcomp|$ predictions in total. To reduce the training overhead, we take advantage of the fact that a prediction in an area will be largely affected by its neighboring areas and let $\Xcomp$ fill only the neighbors instead of all the remaining areas. LaserMix mixes two scans by \textit{intertwining} the areas so that the neighbors of each area are filled with data from the other scan. As a result, we obtain the prediction on all areas $A$ of two scans from only two predictions, which on average reduces the cost from $|A|$ to $1$. The scan before and after mixing counts as two data fillings, therefore $|\Xcomp|=2$. Overall, the training overhead is reduced from $|A|\cdot|\Xcomp|$ to $2$: only one prediction on original data and one additional prediction on mixed data are required for each \lidar scan. During training, the memory consumption for a batch will be $2\times$ compared to a standard SSL framework, and the training speed will not be affected.

\begin{figure*}[t]
    \begin{center}
    \includegraphics[width=0.99\linewidth]{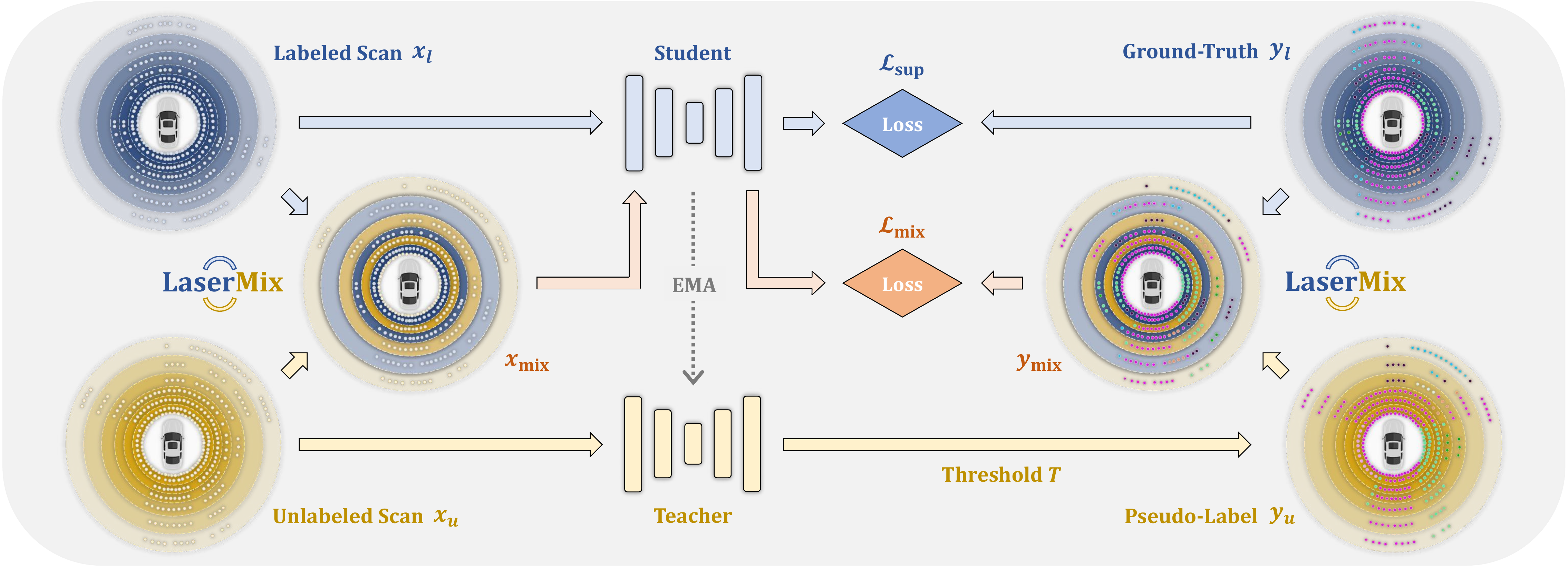}
    \end{center}
    \vspace{-0.5cm}
    \caption{\textbf{Framework overview}. The labeled scan $x_{l}$ is fed into the Student net to compute the supervised loss $\mathcal{L}_{\text{sup}}$ (\textit{w/} ground-truth $y_{l}$). The unlabeled scan $x_{u}$ and the generated pseudo-label $y_{u}$ are mixed with $(x_{l},y_{l})$ via LaserMix (\cref{sec:lasermix}) to produce mixed data sample $(x_{\text{mix}},y_{\text{mix}})$, which is then fed into the Student net to compute the mixing loss $\mathcal{L}_{\text{mix}}$. Additionally, we adopt the EMA update in~\cite{MeanTeacher} for the Teacher net and compute the mean teacher loss $\mathcal{L}_{\text{mt}}$ over Student net's and Teacher net's predictions.}
    \label{fig:framework}
    \vspace{-0.1cm}
\end{figure*}

\subsection{Overall Pipeline}
\label{sec:pipeline}

We show the overall framework in \cref{fig:framework} and the pseudo-code in Algo.~\ref{alg:lasermix}. There are two branches in our pipeline, one Student net $\mathcal{G}_{\theta}^{s}$ and one Teacher net $\mathcal{G}_{\theta}^{t}$. During training, a batch is composed of half labeled data and half unlabeled data. We collect the predictions from both $\mathcal{G}_{\theta}^{s}$ and $\mathcal{G}_{\theta}^{t}$, and produce pseudo-labels from Teacher net's prediction with a predefined confidence threshold $T$. For labeled data, we compute the cross-entropy loss between the Student net's prediction and the ground-truth as $\mathcal{L}_{\text{sup}}$. 
For unlabeled data, LaserMix blends every scan with a random labeled scan, together with their pseudo-label or ground-truth. Then, we let $\mathcal{G}_{\theta}^{s}$ predict on the mixed data and compute the cross-entropy loss $\mathcal{L}_{\text{mix}}$ (\textit{w/} mixed labels). The point-wise cross-entropy loss for a scan $x$ and its corresponding ground-truth/pseudo-label $y$ on the segmentation net $\mathcal{G}_{\theta}$ is defined as:
\begin{align}
    \mathcal{L}_{\text{ce}} =\frac{1}{|x|}\sum_{i=1}^{|x|}\text{CrossEntropy}(y^{(i)}, \mathcal{G}^{(i)}_\theta(x))\,,
\end{align}
where $(i)$ denotes the $i$-th point. Moreover, we adopt the mean teacher idea in~\cite{MeanTeacher} and use Exponential Moving Average (EMA) to update the weights of $\mathcal{G}_{\theta}^{t}$ from $\mathcal{G}_{\theta}^{s}$, and compute the L$2$ loss between their predictions as $\mathcal{L}_{\text{mt}}$:
\begin{align}
    \mathcal{L}_{\text{mt}} = ||\mathcal{G}_{\theta}^{s}(x) - \mathcal{G}_{\theta}^{t}(x)||^2_2\,,
\end{align}
where $||\cdot||^2_2$ is the L$2$ norm. The overall loss function is $\mathcal{L} = \mathcal{L}_{\text{sup}} + \lambda_\textrm{mix} \mathcal{L}_{\text{mix}} + \lambda_\textrm{mt}\mathcal{L}_{\text{mt}}$,
where $\lambda_\textrm{mix}$ and $\lambda_\textrm{mt}$ are loss weights. We use the Teacher net during inference as it empirically gives more stable results. There will be no extra inference overhead in our framework.

\textit{Role in our framework:} Our overall pipeline minimizes the marginal entropy, as described by \textit{Step 3} in our framework. Since the objective for minimizing the entropy has a hard optimization landscape, pseudo-labeling is a common resort in practice~\cite{lee2013pseudo}. Unlike conventional pseudo-label optimization in SSL which only aims to encourage the predictions to be confident, minimizing the marginal entropy requires all predictions to be both confident and consistent. Therefore, we use the ground-truth and pseudo-label as an anchor and encourage the model's predictions to be confident and consistent with these supervision signals.

\begin{algorithm}[!t]
\setstretch{1.1}
        \caption{Pseudo-code for one training iteration.}
   \label{alg:lasermix}
\begin{algorithmic}[1]
  \footnotesize
   \STATE {\bfseries Input:} Shuffled labeled batch $(X_l, Y_l)=\{(x^{(b)}_l, y^{(b)}_l); b \in (1, \ldots, B)\}$, shuffled unlabeled batch $X_u=\{x^{(b)}_u; b \in (1, \ldots, B)\}$, threshold $T$, loss weights $\lambda_\textrm{mix}$ and $\lambda_\textrm{mt}$, Student net and Teacher net. \\
   \FOR{$b = 1$ \TO $B$}
   \STATE $x^{(2b-1)}_\text{mix}$, $x^{(2b)}_\text{mix}$ = LaserMix($x^{(b)}_l$, $x^{(b)}_u$) \COMMENT{\textit{LaserMix data}} \\
   \ENDFOR
   \STATE $X_\text{mix} = \{{x}^{(i)}_\text{mix}; i \in (1, \ldots, 2B)\}$ 
   \STATE $S_l$, $S_u$, $S_\text{mix}$ = Student$\big($Concat($X_l$, $X_u$, $X_\text{mix}$)$\big)$ \COMMENT{\textit{Student pred}} \\
   \STATE $\hat{S}_l$, $\hat{S}_u$ = Teacher$\big($Concat($X_l$, $X_u$)$\big)$  \COMMENT{\textit{Teacher pred}} \\
   \STATE $Y_u$ = PseudoLabel($\hat{S}_u$, $T$) 
   \COMMENT{\textit{Pseudo-label generation process}} \\
   \FOR{$b = 1$ \TO $B$}
   \STATE $y^{(2b-1)}_\text{mix}$, $y^{(2b)}_\text{mix}$ = LaserMix($y^{(b)}_l$, $y^{(b)}_u$) \COMMENT{\textit{LaserMix label}}\\
   \ENDFOR
   \STATE $Y_\text{mix} = \{{y}^{(i)}_\text{mix}; i \in (1, \ldots, 2B)\}$
   \STATE $L_\text{sup}$ = CrossEntropy($S_l$, $Y_l$) \COMMENT{\textit{Supervised loss}}\\
   \STATE $L_\text{mix}$ = CrossEntropy($S_\text{mix}$, $Y_\text{mix}$) \COMMENT{\textit{Mixing loss}}\\
   \STATE $L_\text{mt}$ = L2$\big($Concat($S_l$,  $S_u$), Concat($\hat{S}_l$, $\hat{S}_u$)$\big)$ \COMMENT{\textit{MeanTeacher loss}}\\
   \STATE ${L} = {L}_{\text{sup}} + \lambda_\textrm{mix} {L}_{\text{mix}} + \lambda_\textrm{mt}{L}_{\text{mt}}$ \COMMENT{\textit{Overall loss}}\\
   \STATE Backward({L}), Update(Student), UpdateEMA(Teacher) \\
\end{algorithmic}
\end{algorithm}

\section{Experiments}
\label{sec:experiments}

\begin{table*}[t]
\caption{Benchmarking results among different SSL methods with the LiDAR \textit{range view} (top) and \textit{voxel} (bottom) representations. All mIoU scores are given in percentage ($\%$). The \textbf{best} and \underline{second best} score for each data split is highlighted in \textbf{bold} and \underline{underline}.}
\vspace{-0.2cm}
\setlength{\tabcolsep}{4pt}
\centering\scalebox{0.79}{
\begin{tabular}{c|p{85pt}<{\centering}|p{23pt}<{\centering}p{23pt}<{\centering}p{23pt}<{\centering}p{23pt}<{\centering}|p{23pt}<{\centering}p{23pt}<{\centering}p{23pt}<{\centering}p{23pt}<{\centering}|p{23pt}<{\centering}p{23pt}<{\centering}p{23pt}<{\centering}p{23pt}<{\centering}}
\toprule
\multirow{2}{*}{Repr.} & \multirow{2}{*}{Method} &
\multicolumn{4}{c|}{nuScenes~\cite{Panoptic-nuScenes}} & \multicolumn{4}{c|}{SemanticKITTI~\cite{SemanticKITTI}} & \multicolumn{4}{c}{ScribbleKITTI~\cite{ScribbleKITTI}}
\\
& & $1\%$ & $10\%$ & $20\%$ & $50\%$ & $1\%$ & $10\%$ & $20\%$ & $50\%$ & $1\%$ & $10\%$ & $20\%$ & $50\%$
\\\midrule\midrule
\multirow{8}{*}{\rotatebox{90}{Range View}} &
\cellcolor{yellow!12.5}\textit{Sup.-only} & \cellcolor{yellow!12.5}$38.3$ & \cellcolor{yellow!12.5}$57.5$ & \cellcolor{yellow!12.5}$62.7$ & \cellcolor{yellow!12.5}$67.6$ & \cellcolor{yellow!12.5}$36.2$ & \cellcolor{yellow!12.5}$52.2$ & \cellcolor{yellow!12.5}$55.9$ & \cellcolor{yellow!12.5}$57.2$ & \cellcolor{yellow!12.5}$33.1$ & \cellcolor{yellow!12.5}$47.7$ & \cellcolor{yellow!12.5}$49.9$ & \cellcolor{yellow!12.5}$52.5$
\\\cmidrule{2-14}
& MeanTeacher~\cite{MeanTeacher} & $42.1$ & $60.4$ & \underline{$65.4$} & $69.4$ & $37.5$ & $53.1$ & $56.1$ & $57.4$ & $34.2$ & $49.8$ & $51.6$ & $53.3$
\\
& CBST~\cite{CBST} & $40.9$ & $60.5$ & $64.3$ & $69.3$ & \underline{$39.9$} & $53.4$ & $56.1$ & $56.9$ & $35.7$ & \underline{$50.7$} & $52.7$ & \underline{$54.6$}
\\
& CutMix-Seg~\cite{CutMix-Seg} & \underline{$43.8$} & \underline{$63.9$} & $64.8$ & \underline{$69.8$} & $37.4$ & \underline{$54.3$} & \underline{$56.6$} & \underline{$57.6$} & \underline{$36.7$} & \underline{$50.7$} & \underline{$52.9$} & $54.3$
\\
& CPS~\cite{CPS} & $40.7$ & $60.8$ & $64.9$ & $68.0$ & $36.5$ & $52.3$ & $56.3$ & $57.4$ & $33.7$ & $50.0$ & $52.8$ & \underline{$54.6$}
\\\cmidrule{2-14}
& \textbf{LaserMix~(Ours)} & $\mathbf{49.5}$ & $\mathbf{68.2}$ & $\mathbf{70.6}$ & $\mathbf{73.0}$ & $\mathbf{43.4}$ & $\mathbf{58.8}$ & $\mathbf{59.4}$ & $\mathbf{61.4}$ & $\mathbf{38.3}$ & $\mathbf{54.4}$ & $\mathbf{55.6}$ & $\mathbf{58.7}$
\\
& $\Delta$ $\uparrow$ & \textcolor{brown}{\small{$\mathbf{+11.2}$}} & \textcolor{brown}{\small{$\mathbf{+10.7}$}} & \textcolor{brown}{\small{$\mathbf{+7.9}$}} & \textcolor{brown}{\small{$\mathbf{+5.4}$}} & \textcolor{brown}{\small{$\mathbf{+7.2}$}} & \textcolor{brown}{\small{$\mathbf{+6.6}$}} & \textcolor{brown}{\small{$\mathbf{+3.5}$}} & \textcolor{brown}{\small{$\mathbf{+4.2}$}} & \textcolor{brown}{\small{$\mathbf{+5.2}$}} & \textcolor{brown}{\small{$\mathbf{+6.7}$}} & \textcolor{brown}{\small{$\mathbf{+5.7}$}} & \textcolor{brown}{\small{$\mathbf{+6.2}$}}
\\
\midrule\midrule
\multirow{7}{*}{\rotatebox{90}{Voxel}} &
\cellcolor{yellow!12.5}\textit{Sup.-only} & \cellcolor{yellow!12.5}$50.9$ & \cellcolor{yellow!12.5}$65.9$ & \cellcolor{yellow!12.5}$66.6$ & \cellcolor{yellow!12.5}$71.2$ & \cellcolor{yellow!12.5}$45.4$ & \cellcolor{yellow!12.5}$56.1$ & \cellcolor{yellow!12.5}$57.8$ & \cellcolor{yellow!12.5}$58.7$ & \cellcolor{yellow!12.5}$39.2$ & \cellcolor{yellow!12.5}$48.0$ & \cellcolor{yellow!12.5}$52.1$ & \cellcolor{yellow!12.5}$53.8$
\\\cmidrule{2-14}
& MeanTeacher~\cite{MeanTeacher} & $51.6$ & $66.0$ & $67.1$ & $71.7$ & $45.4$ & $57.1$ & $59.2$ & $60.0$ & $41.0$ & $50.1$ & $52.8$ & $53.9$
\\
& CBST~\cite{CBST} & \underline{$53.0$} & \underline{$66.5$} & $69.6$ & $71.6$ & \underline{$48.8$} & $58.3$ & $59.4$ & $59.7$ & \underline{$41.5$} & $50.6$ & $53.3$ & $54.5$
\\
& CPS~\cite{CPS} & $52.9$ & $66.3$ & \underline{$70.0$} & \underline{$72.5$} &  $46.7$ & \underline{$58.7$} & \underline{$59.6$} & \underline{$60.5$} & $41.4$ & \underline{$51.8$} & \underline{$53.9$} & \underline{$54.8$}
\\\cmidrule{2-14}
& \textbf{LaserMix~(Ours)} & $\mathbf{55.3}$ & $\mathbf{69.9}$ & $\mathbf{71.8}$ & $\mathbf{73.2}$ & $\mathbf{50.6}$ & $\mathbf{60.0}$ & $\mathbf{61.9}$ & $\mathbf{62.3}$ & $\mathbf{44.2}$ & $\mathbf{53.7}$ & $\mathbf{55.1}$ & $\mathbf{56.8}$
\\
& $\Delta$ $\uparrow$ & \textcolor{brown}{\small{$\mathbf{+4.4}$}} & \textcolor{brown}{\small{$\mathbf{+4.0}$}} & \textcolor{brown}{\small{$\mathbf{+5.2}$}} & \textcolor{brown}{\small{$\mathbf{+2.0}$}} & \textcolor{brown}{\small{$\mathbf{+5.2}$}} & \textcolor{brown}{\small{$\mathbf{+3.9}$}} & \textcolor{brown}{\small{$\mathbf{+4.1}$}} & \textcolor{brown}{\small{$\mathbf{+3.6}$}} & \textcolor{brown}{\small{$\mathbf{+5.0}$}} & \textcolor{brown}{\small{$\mathbf{+5.7}$}} & \textcolor{brown}{\small{$\mathbf{+3.0}$}} & \textcolor{brown}{\small{$\mathbf{+3.0}$}}
\\
\bottomrule
\end{tabular}}
\label{tab:benchmark}
\vspace{-0.cm}
\end{table*}
\begin{table*}[t]
\begin{minipage}[t]{0.38\linewidth}
\caption{Comparison to the state-of-the-art 3D SSL method~\cite{GPC} on the \textit{val} set of SemanticKITTI \cite{SemanticKITTI}. All mIoU scores are given in percentage ($\%$).}
\vspace{-0.2cm}
\centering\scalebox{0.79}{
\begin{tabular}{c|p{20pt}<{\centering}p{20pt}<{\centering}p{20pt}<{\centering}p{20pt}<{\centering}p{20pt}<{\centering}}
\toprule
Method & $5\%$ & $10\%$ & $20\%$ & $30\%$ & $40\%$ 
\\\midrule
\cellcolor{yellow!12.5}GPC~\cite{GPC} & \cellcolor{yellow!12.5}$41.8$ & \cellcolor{yellow!12.5}$49.9$ & \cellcolor{yellow!12.5}$58.8$ & \cellcolor{yellow!12.5}$59.4$ & \cellcolor{yellow!12.5}$59.9$
\\\midrule
\textbf{Ours (RV)} & $54.6$ & $58.8$ & $59.4$ & $60.1$ & $60.8$
\\
$\Delta$ $\uparrow$ & \textcolor{brown}{\small{$\mathbf{+12.8}$}} & \textcolor{brown}{\small{$\mathbf{+8.9}$}} & \textcolor{brown}{\small{$\mathbf{+0.6}$}} & \textcolor{brown}{\small{$\mathbf{+0.7}$}} & \textcolor{brown}{\small{$\mathbf{+0.9}$}}
\\\midrule
\textbf{Ours (Voxel)} & $56.7$ & $60.0$ & $61.9$ & $62.1$ & $62.3$
\\
$\Delta$ $\uparrow$ & \textcolor{brown}{\small{$\mathbf{+14.9}$}} & \textcolor{brown}{\small{$\mathbf{+10.1}$}} & \textcolor{brown}{\small{$\mathbf{+3.1}$}} & \textcolor{brown}{\small{$\mathbf{+1.7}$}} & \textcolor{brown}{\small{$\mathbf{+1.4}$}}
\\\bottomrule
\end{tabular}}
\vspace{-0.1cm}
\label{tab:benchmark-sota}
\end{minipage}
~~~~~~
\begin{minipage}[t]{0.58\linewidth}
\caption{Ablation results on the \textit{val} set of nuScenes~\cite{Panoptic-nuScenes}. (1) Baseline results~\cite{MeanTeacher}; (2) Results with Student net supervision (SS); (3) Results with Teacher net supervision (TS). All mIoU scores are given in percentage ($\%$).}
\vspace{-0.2cm}
\centering\scalebox{0.79}{
\begin{tabular}{c|cccc|ccccc}
\toprule
\# & $\mathcal{L}_{\text{mt}}$ & $\mathcal{L}_{\text{mix}}$ & SS & TS & $1\%$ & $10\%$ & $20\%$ & $50\%$ 
\\\midrule
$(1)$ & \cellcolor{yellow!12.5}\checkmark & \cellcolor{yellow!12.5} & \cellcolor{yellow!12.5} & \cellcolor{yellow!12.5} & \cellcolor{yellow!12.5}$42.1$ & \cellcolor{yellow!12.5}$60.4$ & \cellcolor{yellow!12.5}$65.4$ & \cellcolor{yellow!12.5}$69.4$
\\\midrule
\multirow{2}{*}{$(2)$} & & \checkmark & \checkmark & & $45.6$\textcolor{brown}{\small{$\mathbf{(+3.5})$}} & $64.3$\textcolor{brown}{\small{$\mathbf{(+3.9})$}} & $67.8$\textcolor{brown}{\small{$\mathbf{(+2.4})$}} & $71.6$\textcolor{brown}{\small{$\mathbf{(+2.2})$}}
\\
& \checkmark & \checkmark & \checkmark & & $47.0$\textcolor{brown}{\small{$\mathbf{(+4.9})$}} & $65.5$\textcolor{brown}{\small{$\mathbf{(+5.1})$}} & $69.5$\textcolor{brown}{\small{$\mathbf{(+4.1})$}} & $72.0$\textcolor{brown}{\small{$\mathbf{(+2.6})$}}
\\\midrule
\multirow{2}{*}{$(3)$} & & \checkmark & & \checkmark & $46.0$\textcolor{brown}{\small{$\mathbf{(+3.9})$}} & $64.1$\textcolor{brown}{\small{$\mathbf{(+3.7})$}} & $69.5$\textcolor{brown}{\small{$\mathbf{(+4.1})$}} & $72.3$\textcolor{brown}{\small{$\mathbf{(+2.9})$}}
\\
& \checkmark & \checkmark & & \checkmark & $49.5$\textcolor{brown}{\small{$\mathbf{(+7.4})$}} & $68.2$\textcolor{brown}{\small{$\mathbf{(+7.8})$}} & $70.6$\textcolor{brown}{\small{$\mathbf{(+5.2})$}} & $73.0$\textcolor{brown}{\small{$\mathbf{(+3.6})$}}
\\\bottomrule
\end{tabular}}
\label{tab:ablation-supervision}
\end{minipage}
\vspace{-0.1cm}
\end{table*}

\subsection{Settings}
\noindent\textbf{Protocol}. We follow the Realistic Evaluation Protocol \cite{Realistic-Evaluation} when building the benchmark. Specifically, all experiments share the same backbones and are within the same codebase. All configurations and augmentations are unified to ensure a fair comparison among different SSL algorithms.

\noindent\textbf{Data}. We build three SSL benchmarks upon nuScenes \cite{Panoptic-nuScenes}, SemanticKITTI \cite{SemanticKITTI}, and ScribbleKITTI \cite{ScribbleKITTI}. nuScenes \cite{Panoptic-nuScenes} and SemanticKITTI \cite{SemanticKITTI} are the two most popular LiDAR segmentation datasets, with $29130$ and $19130$ training scans and $6019$ and $4071$ validation scans, respectively. ScribbleKITTI \cite{ScribbleKITTI} is a recent variant of SemanticKITTI \cite{SemanticKITTI}, which contains the same number of scans but is annotated with line scribbles (approximately $8.06\%$ valid semantic labels) rather than dense annotations. For all three LiDAR segmentation datasets, we uniformly sample $1\%$, $10\%$, $20\%$, and $50\%$ labeled training scans and assume the remaining ones as unlabeled. This is in line with the conventional settings from the semi-supervised image segmentation community~\cite{CCT,GCT,CPS}. We also conduct experiments with sequential splits and show the results in the Appendix.

\noindent\textbf{Implementation Details}. We adopt FIDNet \cite{FIDNet} and Cylinder3D \cite{Cylinder3D} as the segmentation backbones for the \textit{range view} and the \textit{voxel} options, respectively. The input resolution of range images is set as $64\times2048$ for SemanticKITTI \cite{SemanticKITTI} and ScribbleKITTI \cite{ScribbleKITTI}, and $32\times1920$ for nuScenes \cite{Panoptic-nuScenes}. The voxel resolution is fixed as $[240, 180, 20]$ for all three sets. The number of spatial areas $m$ in LaserMix is uniformly sampled from $2$ to $6$ areas. We denote the supervised-only baseline as \textit{sup.-only}. Due to the lack of LiDAR SSL works \cite{GPC}, we also compare SoTA consistency regularization \cite{MeanTeacher,CPS,CutMix-Seg} and entropy minimization \cite{CBST} methods from semi-supervised image segmentation. We report the intersection-over-union (IoU) over each semantic class and the mean IoU (mIoU) scores over all classes in our benchmarks. All experiments are implemented using PyTorch on NVIDIA Tesla V100 GPUs with 32GB RAM. For additional details, please refer to our Appendix.

\begin{figure*}[t]
    \begin{center}
    \includegraphics[width=0.999\textwidth]{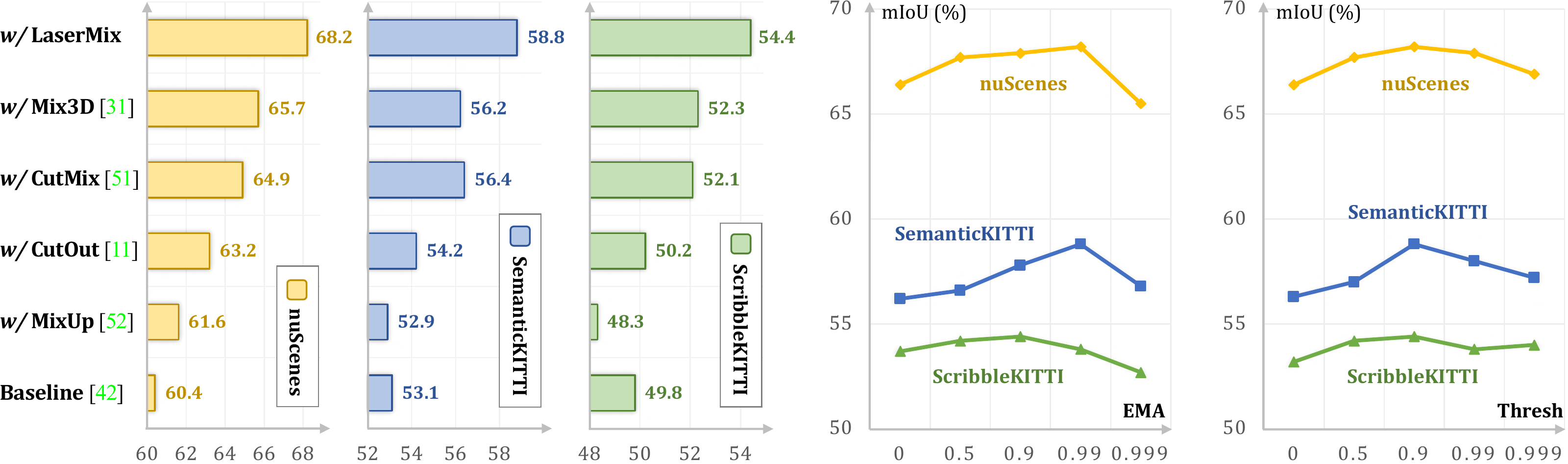}
    \end{center}
    \vspace{-0.5cm}
    \caption{Ablation studies on: \textbf{Left:} Different mixing-based techniques used in point partition \& mixing; \textbf{Middle:} Different EMA decay rates between the Teacher net and the Student net; \textbf{Right:} Different confidence thresholds $T$ used in the pseudo-label generation.}
    \label{fig:ablation-stats}
\end{figure*}

\subsection{Comparative Study}

\noindent\textbf{Improvements over Baseline}. \cref{tab:benchmark} benchmarks results on nuScenes \cite{Panoptic-nuScenes}, SemanticKITTI \cite{SemanticKITTI}, and ScribbleKITTI \cite{ScribbleKITTI}. For all three sets under different data splits, we observe significant improvements in our approach over the \textit{sup.-only} baseline. Such gains are especially evident in \textit{range view}, which reach up to $11.2\%$ mIoU. We also observe constant improvements for the \textit{voxel} option, which provide on average $4.1\%$ mIoU gains over all splits across all sets. The results verify the effectiveness of our framework and further highlight the importance of leveraging unlabeled data in LiDAR semantic segmentation.

\noindent\textbf{Compare with SoTA}. We compare LaserMix with GPC\footnote{Note that GPC uses private backbone / split. See Appendix for details.} \cite{GPC}, the SoTA 3D SSL method tested on SemanticKITTI \cite{SemanticKITTI}. The results in \cref{tab:benchmark-sota} show that our approach exhibits much better results than GPC \cite{GPC}, especially in scenarios where very few annotations are available. We also reimplemented popular SSL algorithms from the image segmentation domain and show their results in \cref{tab:benchmark}. We find that these methods, albeit competitive in 2D, only yield sub-par performance in the LiDAR SSL benchmark, highlighting the importance of exploiting the LiDAR data structure. 

\noindent\textbf{Compare with Full Labels}. As shown in \cref{fig:LaserMix}~(right), the comparisons between the prevailing LiDAR segmentation methods attests that our approach can achieve more competitive scores over the fully-supervised counterparts~\cite{RangeNet++,PolarNet,SalsaNext,PolarStream} while with $2\times$ to $5\times$ fewer annotations. The strong augmentation and regularization ability of LaserMix have yielded better results in the high-data regime and extreme low-data regime (\ie, $0.8\%$ labels on~\cite{ScribbleKITTI}), which validates the generalizability of our approach.

\begin{figure*}[t]
    \begin{center}
    \includegraphics[width=0.9999\textwidth]{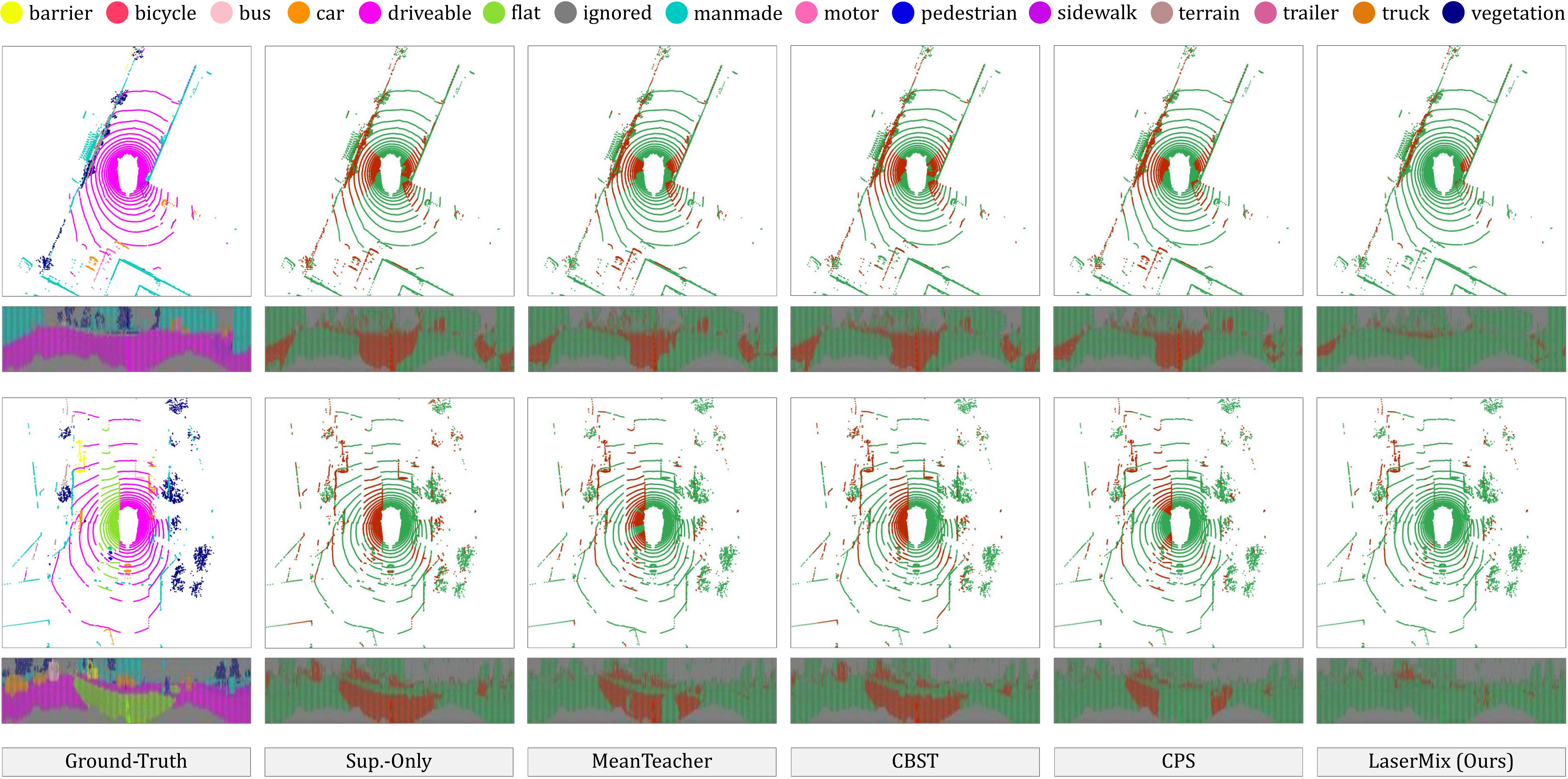}
    \end{center}
    \vspace{-0.45cm}
    \caption{Error maps visualized from LiDAR \textit{bird's eye view} and \textit{range view} on nuScenes~\cite{Panoptic-nuScenes}. Each example shows a street scene of size $50$m by $50$m by $8$m. The \textcolor{forestgreen}{\textbf{correct}} and \textcolor{red}{\textbf{incorrect}} predictions are painted in \textcolor{forestgreen}{\textbf{green}} and \textcolor{red}{\textbf{red}} to highlight the differences. Best viewed in color.}
    \label{fig:qualitative}
\end{figure*}

\noindent\textbf{Qualitative Examination}. \cref{fig:qualitative} visualizes the scene segmentation results for different SSL algorithms on the \textit{val} set of nuScenes~\cite{Panoptic-nuScenes}, where each example covers a driving scene centered by the ego-vehicle. We find that previous arts \cite{MeanTeacher,CBST,CPS} can only improve predictions in limited regions, while our approach holistically eliminates false predictions in almost every direction around the ego-vehicle. The consistency enlightened by LaserMix has yielded better segmentation accuracy under annotation scarcity. 

\subsection{Ablation Study}

Without loss of generalizability, we stick with the $10\%$ budget setting and range view backbones in our ablations.

\noindent\textbf{Framework Setups}. The component analysis in \cref{tab:ablation-supervision} shows that $\mathcal{L}_\text{mix}$ contributes significantly to the overall improvement. Meanwhile, using the Teacher net instead of the Student net to generate pseudo-labels leads to better results, as the formal is temporally ensembled and encourages consistency in-between mixed and original data, which is crucial besides enforcing confident predictions. It is worth noting that all of our model configurations have achieved superior results than the baseline MeanTeacher~\cite{MeanTeacher}, which further emphasizes the effectiveness of our framework in tackling  LiDAR segmentation with semi-supervisions.

\begin{table}
\caption{Ablation studies on laser beam partitions (horizontal: inclination direction $\phi$; vertical: azimuth direction $\alpha$). ($i$-$\alpha$, $j$-$\phi$) denotes that there are $i$ azimuth and $j$ inclination partitions.}
\vspace{-0.2cm}
\centering
\scalebox{0.77}{
\begin{tabular}{p{38.5pt}<{\centering}|p{38.5pt}<{\centering}|p{38.5pt}<{\centering}|p{38.5pt}<{\centering}|p{38.5pt}<{\centering}|p{38.5pt}<{\centering}}
\toprule
\rowcolor[gray]{.95} Baseline & ($1\alpha$, $2\phi$) & ($1\alpha$, $3\phi$) & ($1\alpha$, $4\phi$) & ($1\alpha$, $5\phi$) & ($1\alpha$, $6\phi$)
\\\midrule
\begin{minipage}[b]{0.16\columnwidth}\centering\raisebox{-.4\height}{\includegraphics[width=\linewidth]{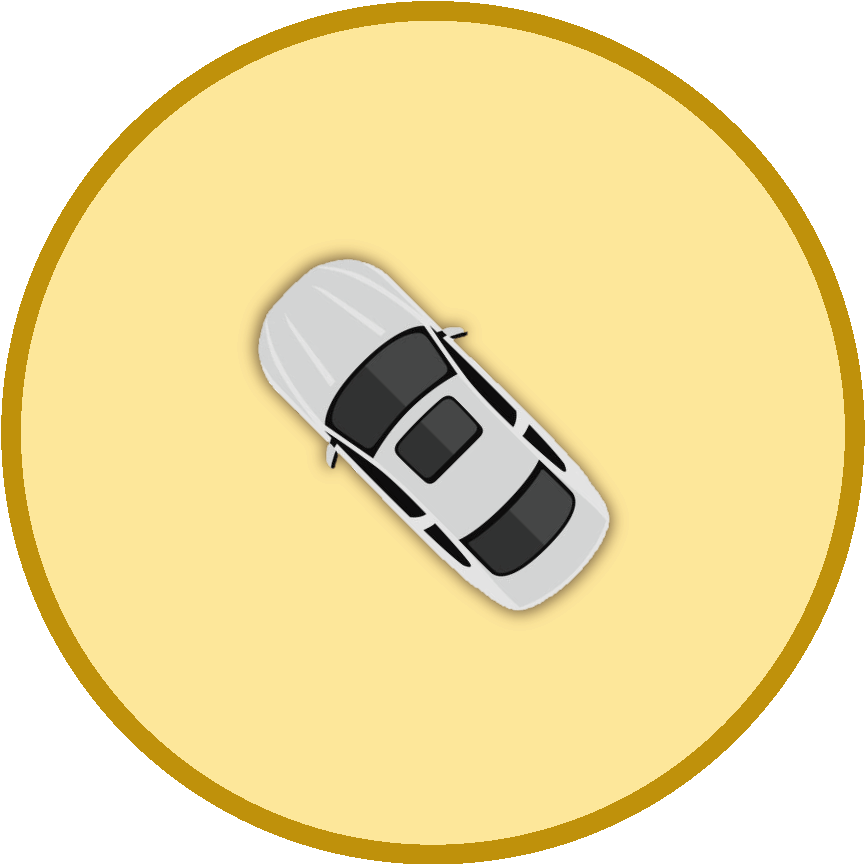}}\end{minipage} &
\begin{minipage}[b]{0.16\columnwidth}\centering\raisebox{-.4\height}{\includegraphics[width=\linewidth]{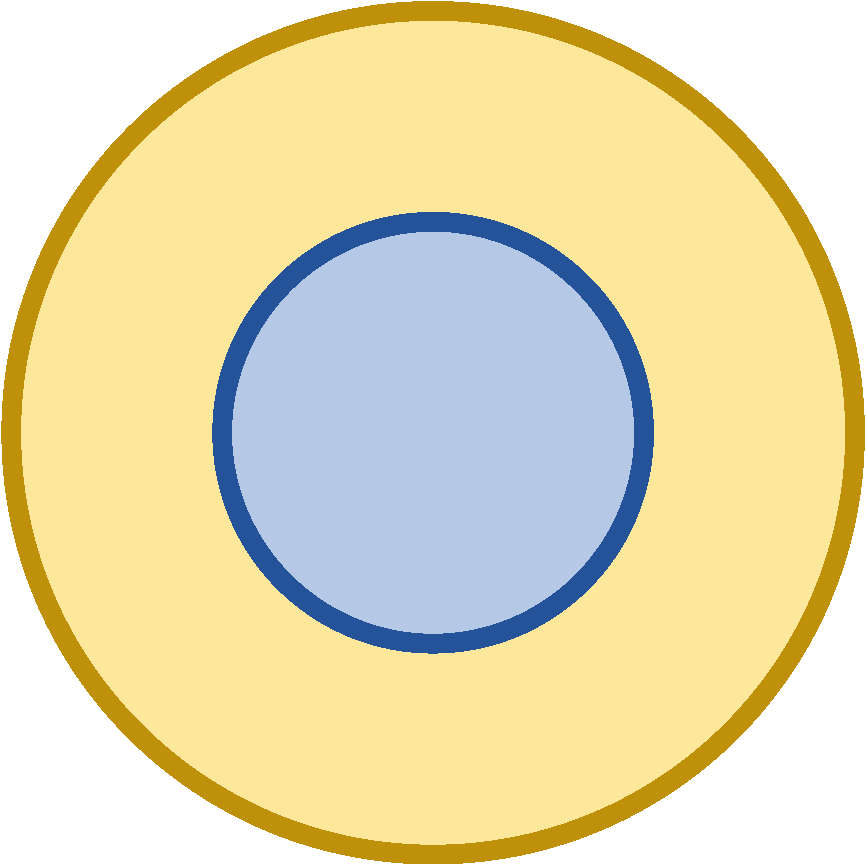}}\end{minipage} &
\begin{minipage}[b]{0.16\columnwidth}\centering\raisebox{-.4\height}{\includegraphics[width=\linewidth]{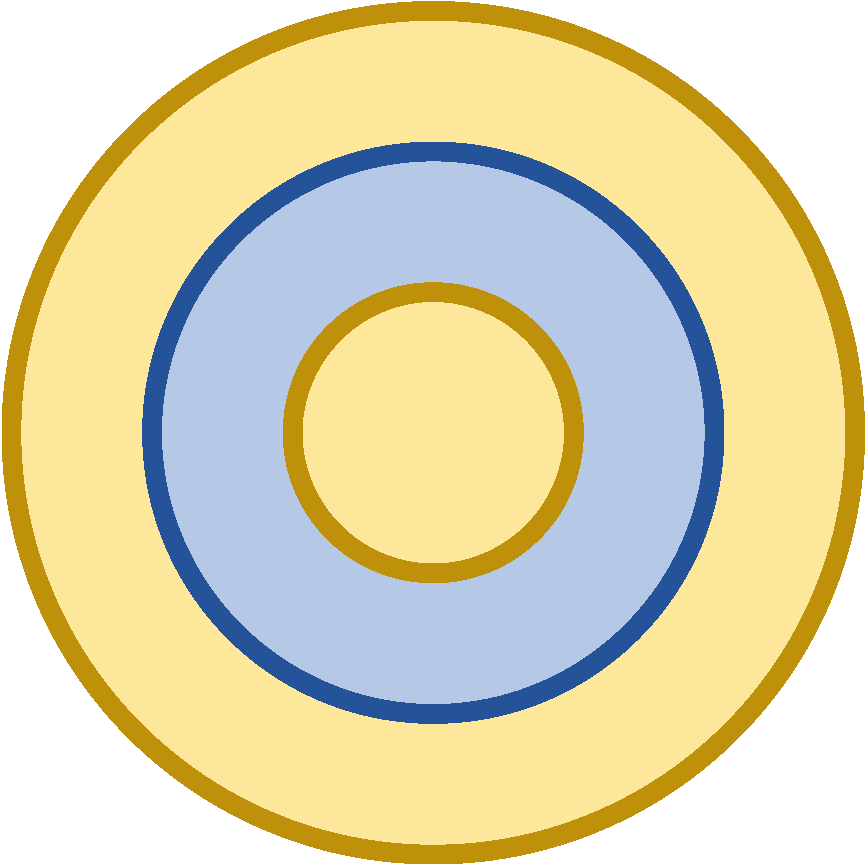}}\end{minipage} &
\begin{minipage}[b]{0.16\columnwidth}\centering\raisebox{-.4\height}{\includegraphics[width=\linewidth]{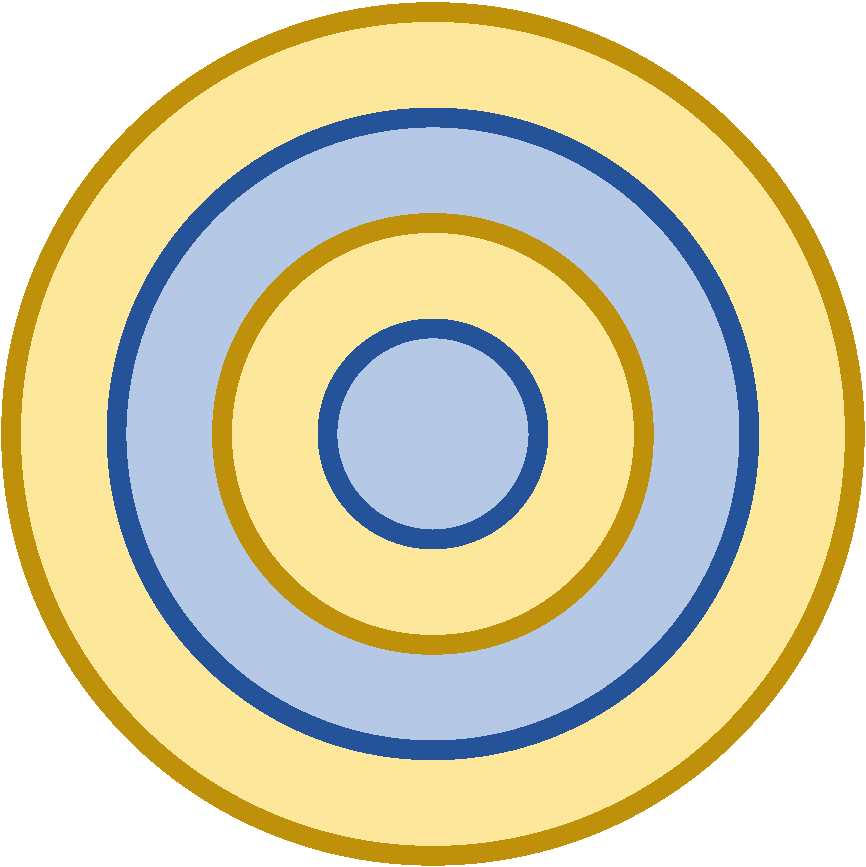}}\end{minipage} &
\begin{minipage}[b]{0.16\columnwidth}\centering\raisebox{-.4\height}{\includegraphics[width=\linewidth]{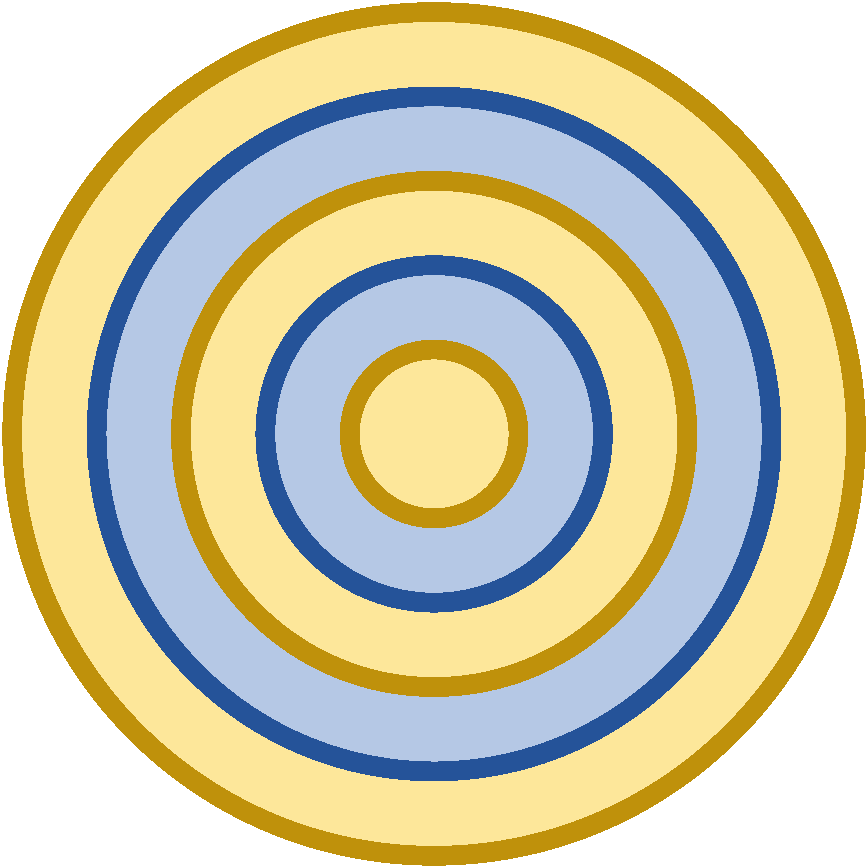}}\end{minipage} &
\begin{minipage}[b]{0.16\columnwidth}\centering\raisebox{-.4\height}{\includegraphics[width=\linewidth]{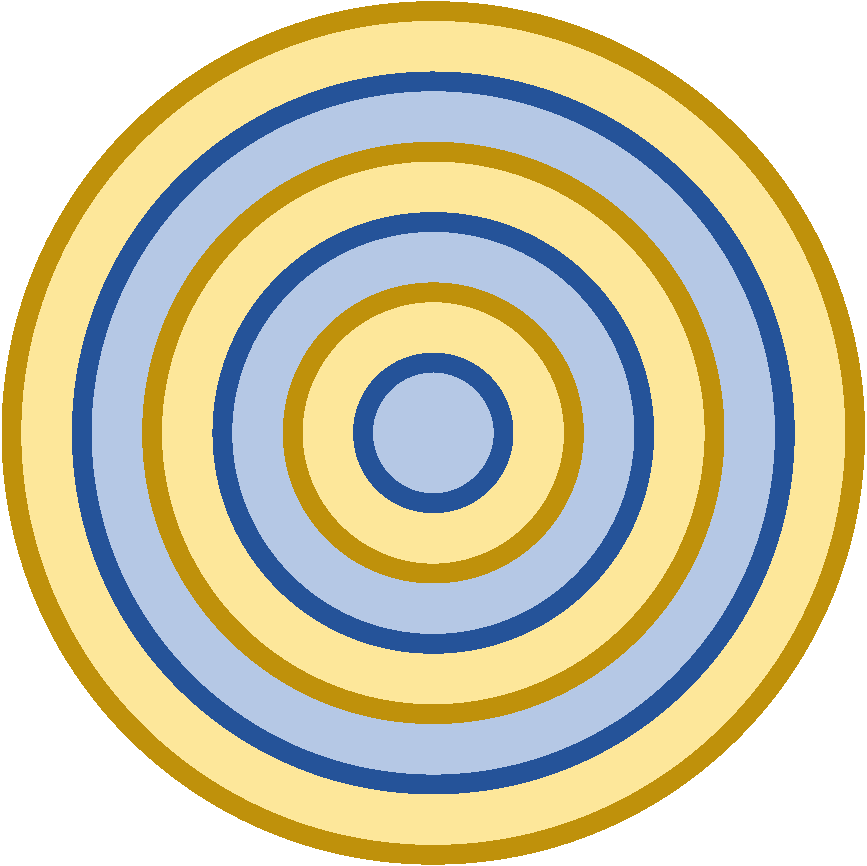}}\end{minipage} 
\\\midrule
$60.4$ & $63.5_\mathbf{{\textcolor{brown}{(+3.1)}}}$ & $65.2_\mathbf{{\textcolor{brown}{(+4.8)}}}$ & $66.5_\mathbf{{\textcolor{brown}{(+6.1)}}}$ & $66.2_\mathbf{{\textcolor{brown}{(+5.8)}}}$ & $65.4_\mathbf{{\textcolor{brown}{(+5.0)}}}$ 
\\\midrule
\rowcolor[gray]{.95} ($2\alpha$, $1\phi$) & ($2\alpha$, $2\phi$) & ($2\alpha$, $3\phi$) & ($2\alpha$, $4\phi$) & ($2\alpha$, $5\phi$) & ($2\alpha$, $6\phi$)
\\\midrule
\begin{minipage}[b]{0.16\columnwidth}\centering\raisebox{-.4\height}{\includegraphics[width=\linewidth]{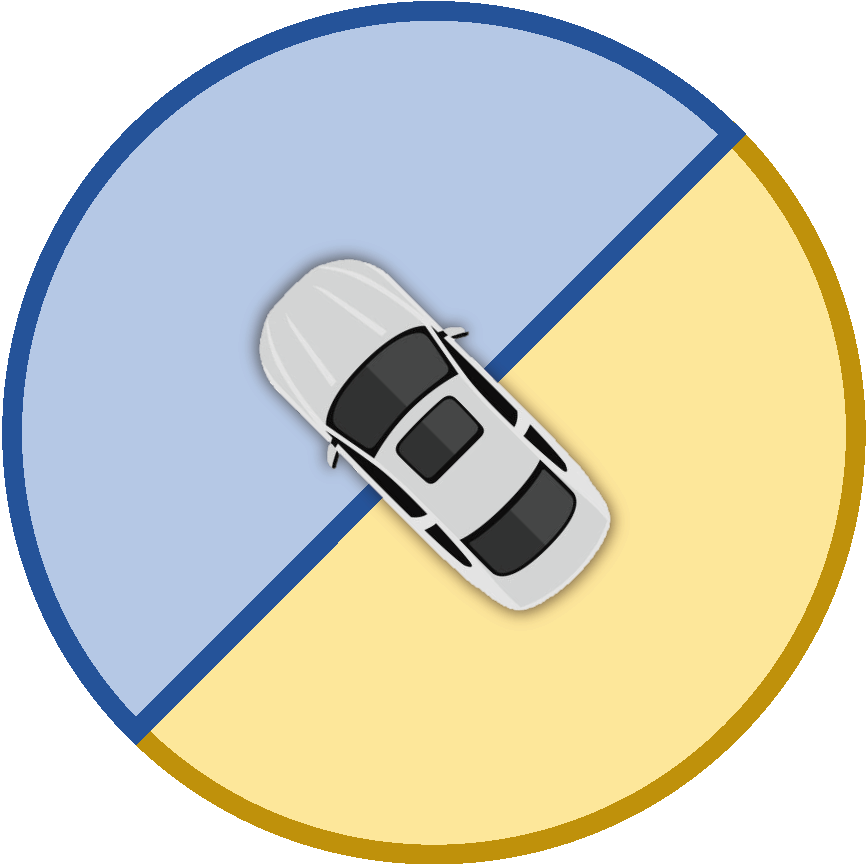}}\end{minipage} & \begin{minipage}[b]{0.16\columnwidth}\centering\raisebox{-.4\height}{\includegraphics[width=\linewidth]{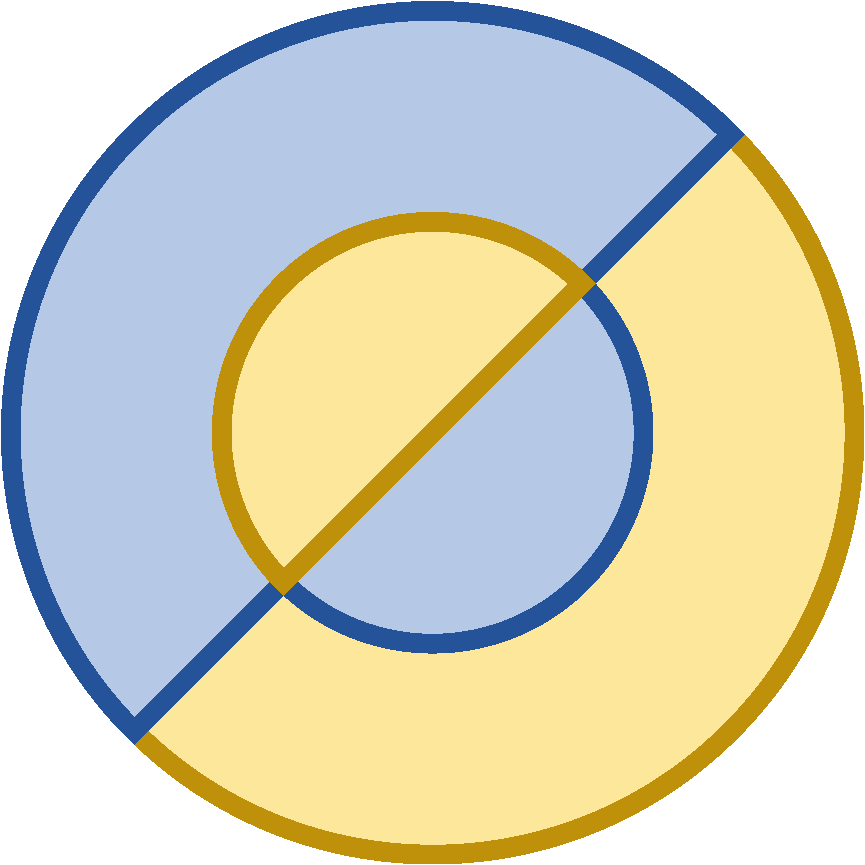}}\end{minipage} &
\begin{minipage}[b]{0.16\columnwidth}\centering\raisebox{-.4\height}{\includegraphics[width=\linewidth]{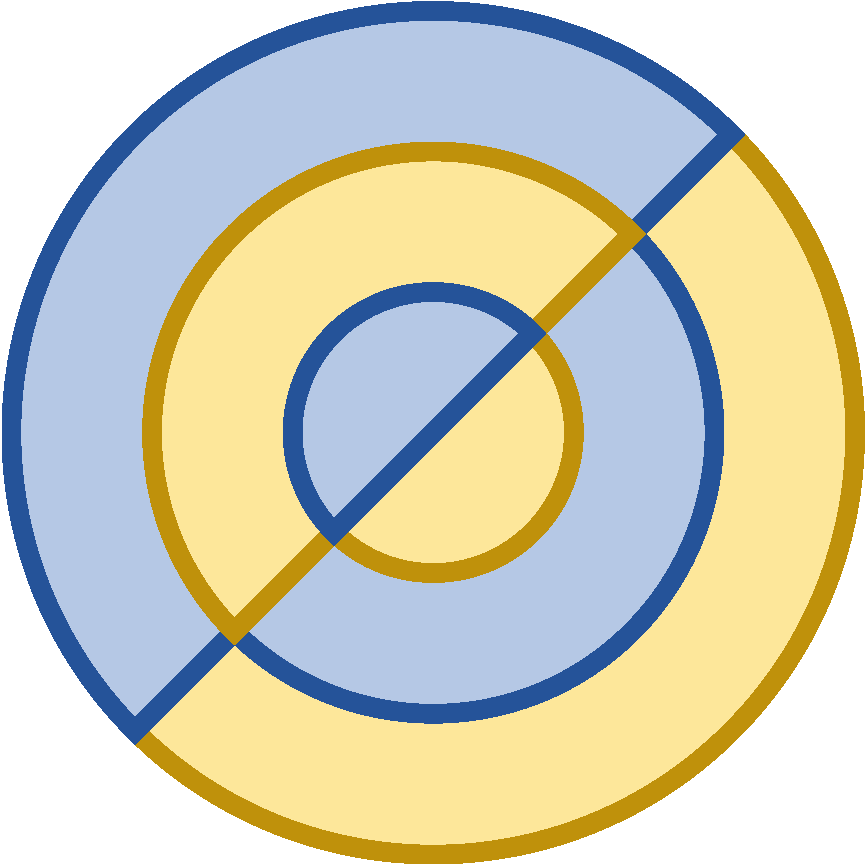}}\end{minipage} &
\begin{minipage}[b]{0.16\columnwidth}\centering\raisebox{-.4\height}{\includegraphics[width=\linewidth]{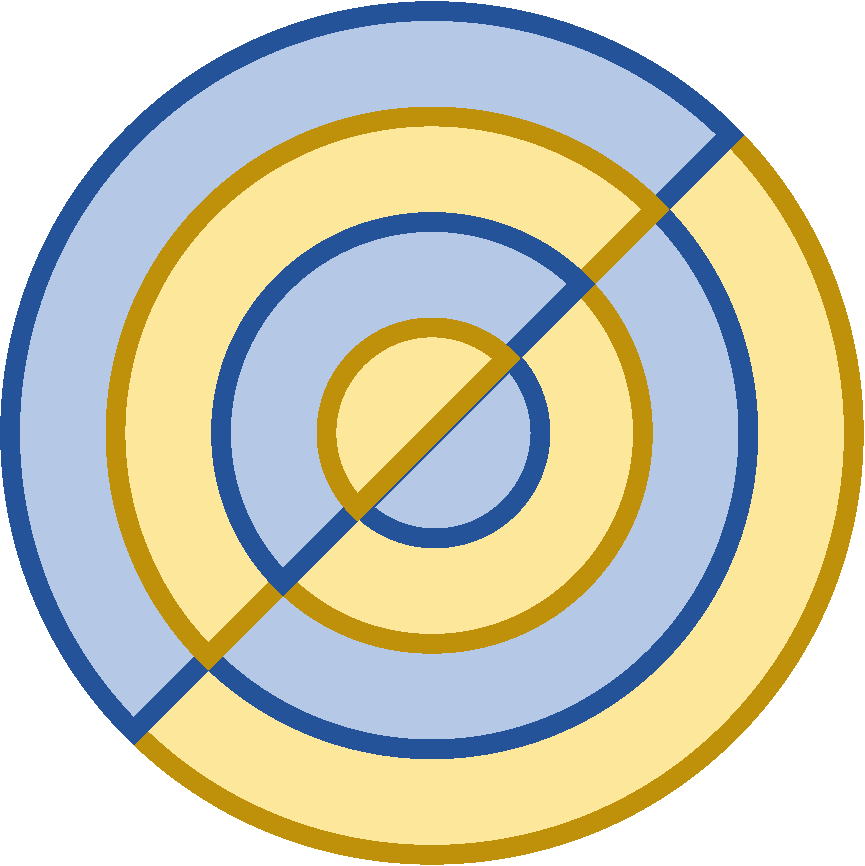}}\end{minipage} &
\begin{minipage}[b]{0.16\columnwidth}\centering\raisebox{-.4\height}{\includegraphics[width=\linewidth]{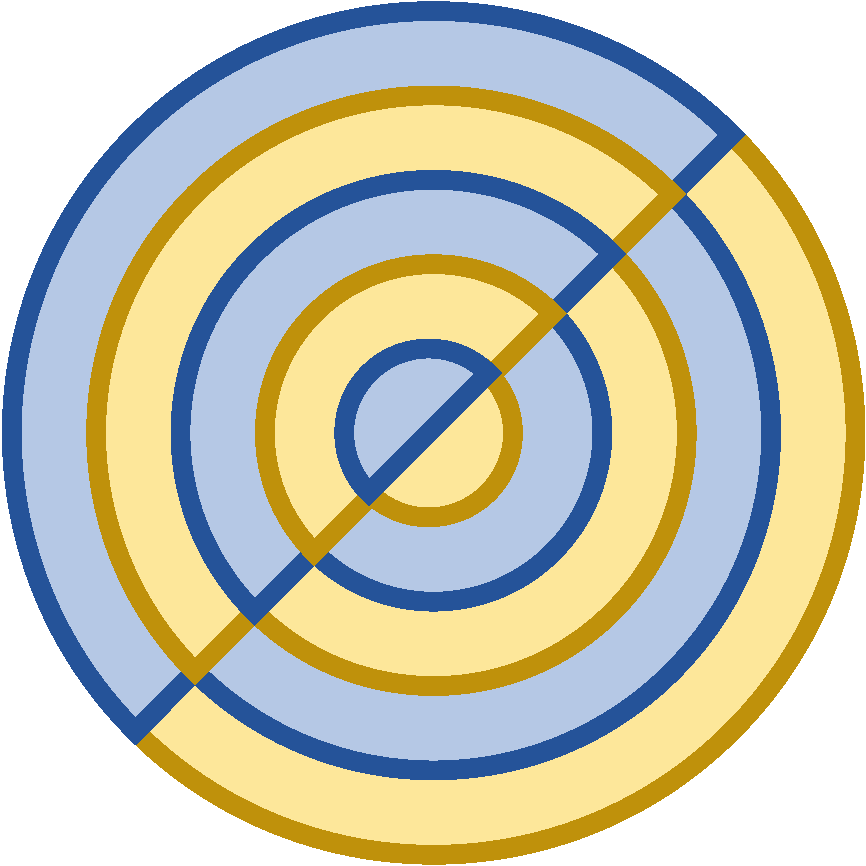}}\end{minipage} &
\begin{minipage}[b]{0.16\columnwidth}\centering\raisebox{-.4\height}{\includegraphics[width=\linewidth]{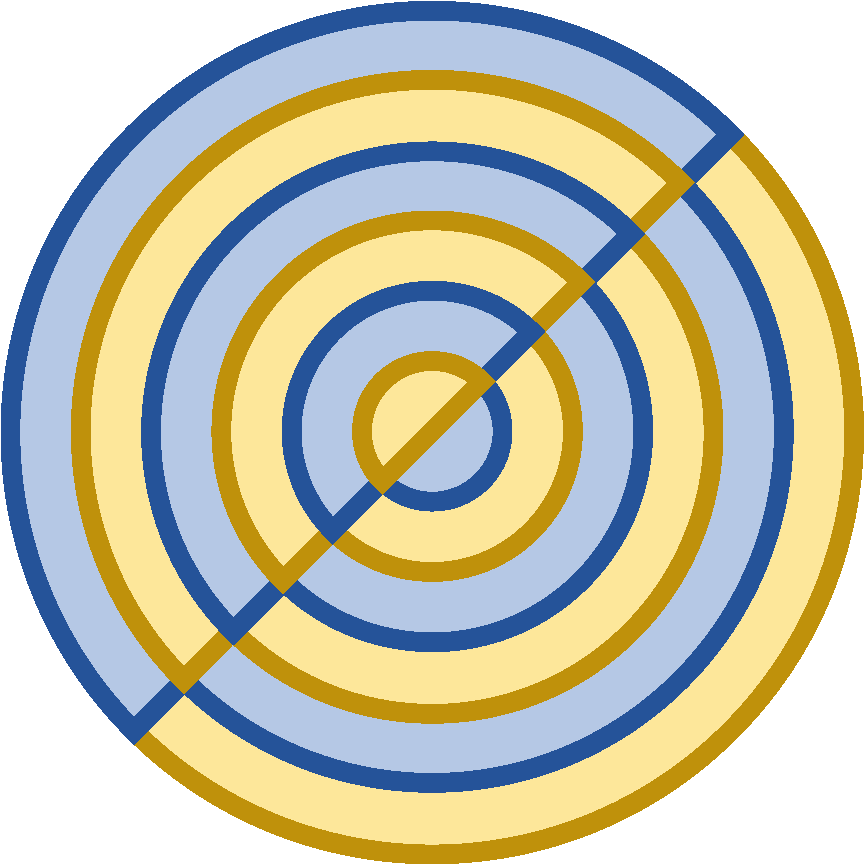}}\end{minipage} 
\\\midrule
$61.5_\mathbf{{\textcolor{brown}{(+1.1)}}}$ & $63.3_\mathbf{{\textcolor{brown}{(+2.9)}}}$ & $65.9_\mathbf{{\textcolor{brown}{(+5.5)}}}$ & $66.1_\mathbf{{\textcolor{brown}{(+5.7)}}}$ & $66.7_\mathbf{{\textcolor{brown}{(+6.3)}}}$ & $65.3_\mathbf{{\textcolor{brown}{(+4.9)}}}$ 
\\\midrule
\rowcolor[gray]{.95} ($3\alpha$, $1\phi$) & ($3\alpha$, $2\phi$) & ($3\alpha$, $3\phi$) & ($3\alpha$, $4\phi$) & ($3\alpha$, $5\phi$) & ($3\alpha$, $6\phi$)
\\\midrule
\begin{minipage}[b]{0.16\columnwidth}\centering\raisebox{-.4\height}{\includegraphics[width=\linewidth]{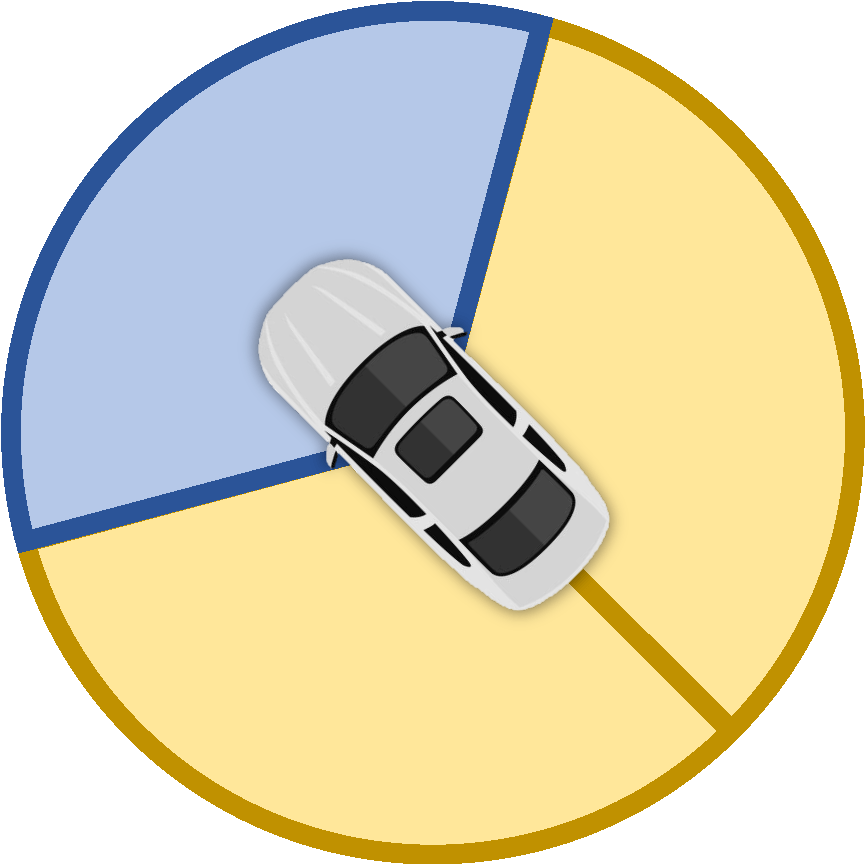}}\end{minipage} & \begin{minipage}[b]{0.16\columnwidth}\centering\raisebox{-.4\height}{\includegraphics[width=\linewidth]{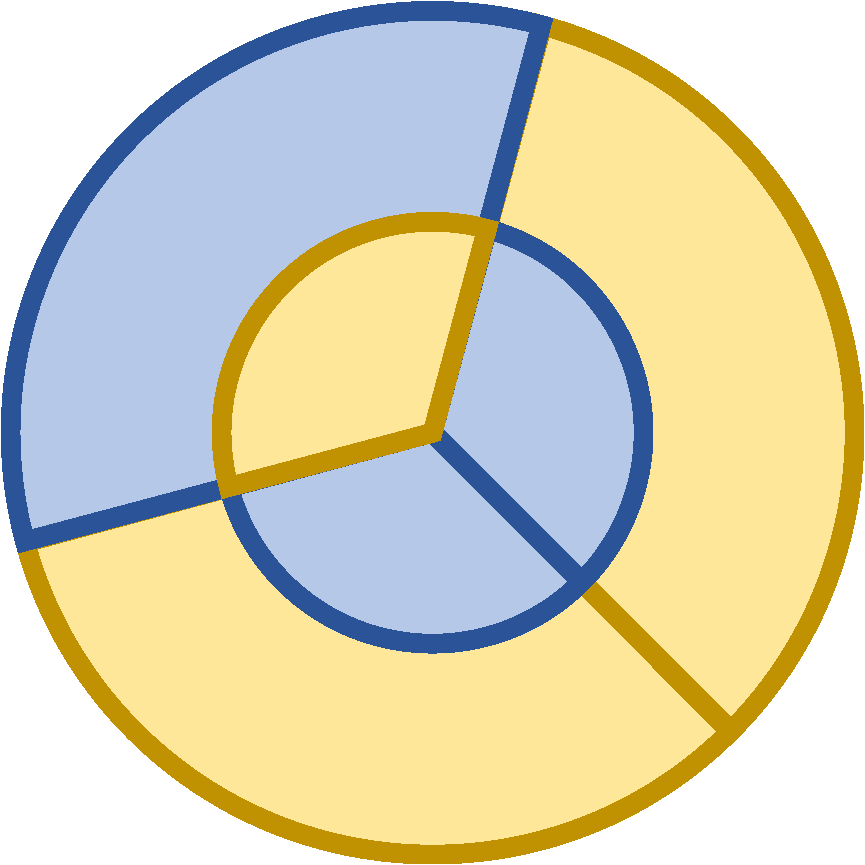}}\end{minipage} & \begin{minipage}[b]{0.16\columnwidth}\centering\raisebox{-.4\height}{\includegraphics[width=\linewidth]{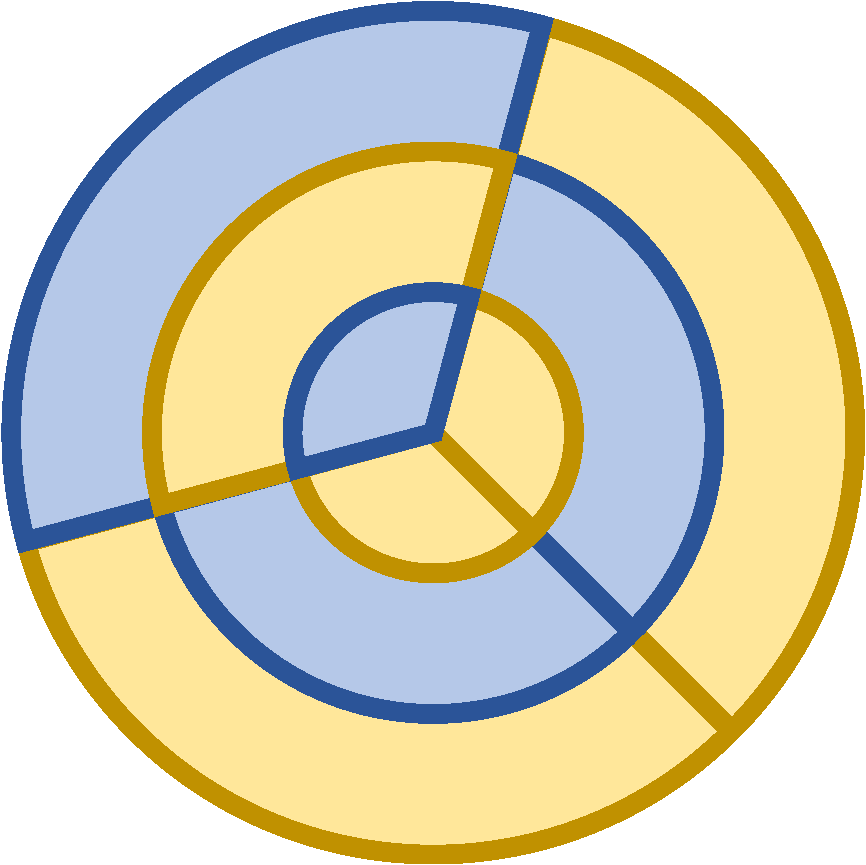}}\end{minipage} & \begin{minipage}[b]{0.16\columnwidth}\centering\raisebox{-.4\height}{\includegraphics[width=\linewidth]{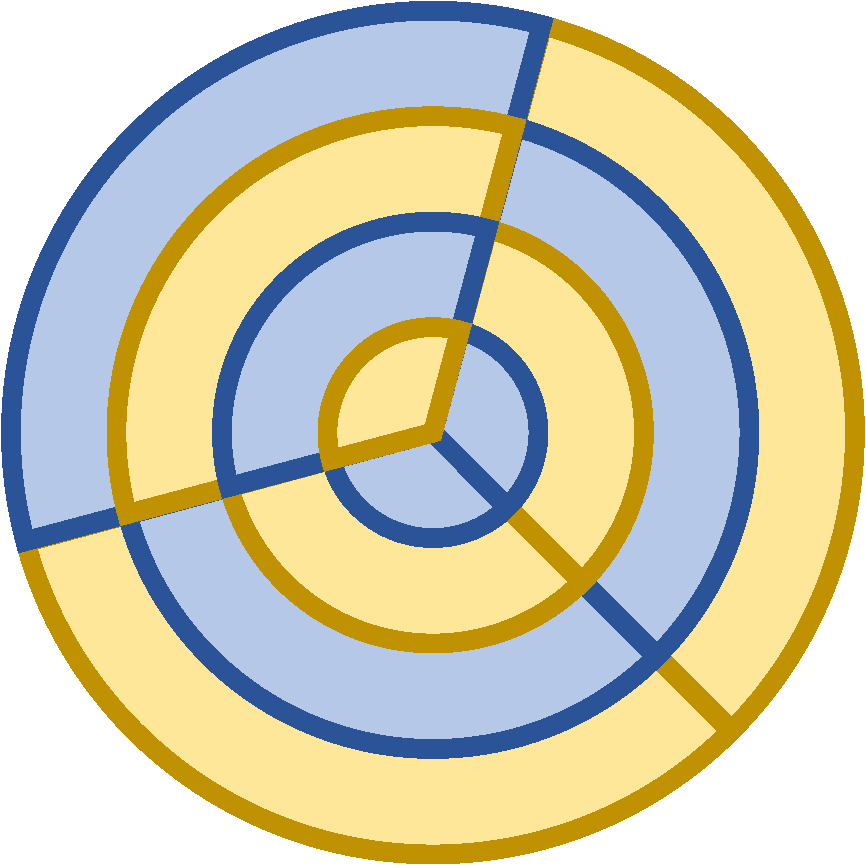}}\end{minipage} & \begin{minipage}[b]{0.16\columnwidth}\centering\raisebox{-.4\height}{\includegraphics[width=\linewidth]{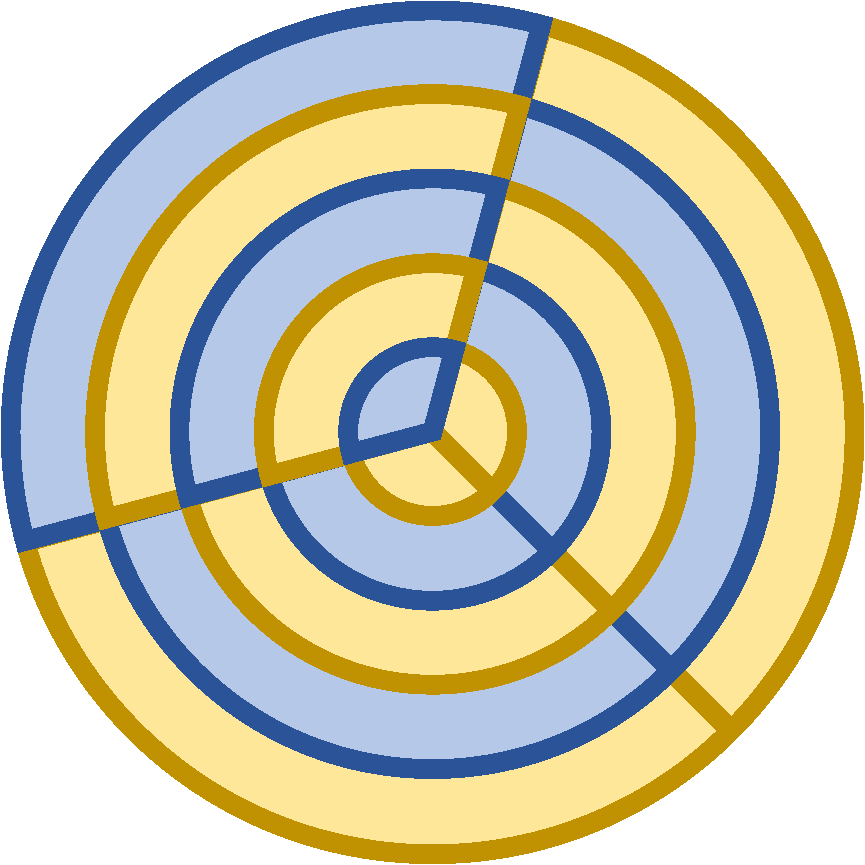}}\end{minipage} & \begin{minipage}[b]{0.16\columnwidth}\centering\raisebox{-.4\height}{\includegraphics[width=\linewidth]{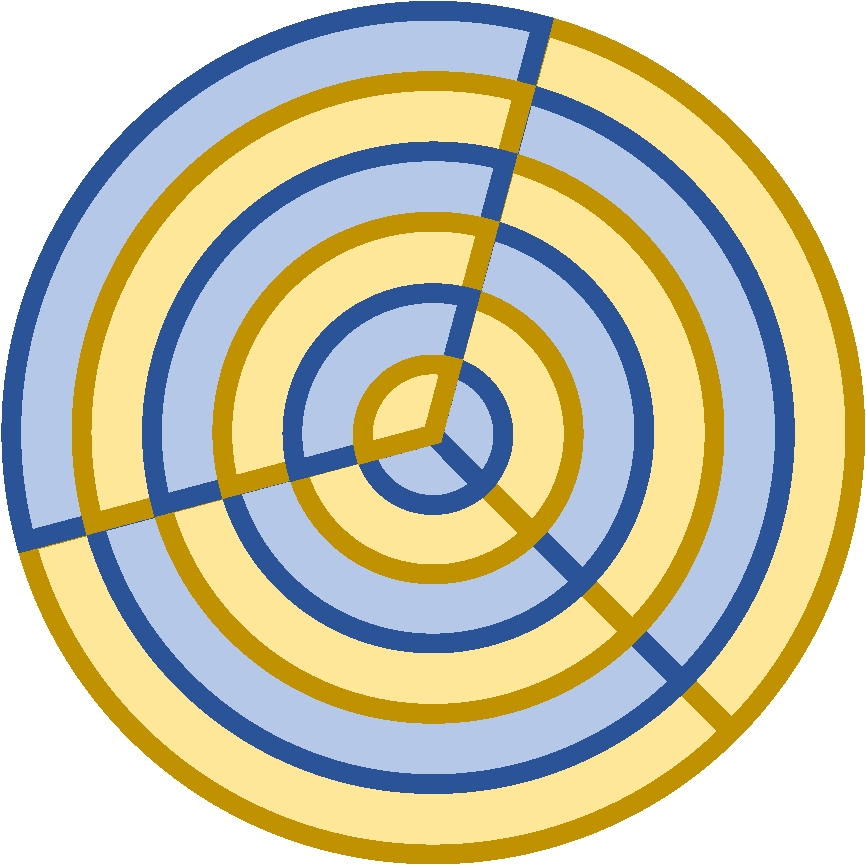}}\end{minipage} 
\\\midrule
$60.9_\mathbf{{\textcolor{brown}{(+0.6)}}}$ & $64.2_\mathbf{{\textcolor{brown}{(+3.8)}}}$ & $65.9_\mathbf{{\textcolor{brown}{(+5.5)}}}$ & $66.3_\mathbf{{\textcolor{brown}{(+5.9)}}}$ & $66.0_\mathbf{{\textcolor{brown}{(+5.6)}}}$ & $65.2_\mathbf{{\textcolor{brown}{(+4.8)}}}$ 
\\\midrule
\rowcolor[gray]{.95} ($4\alpha$, $1\phi$) & ($4\alpha$, $2\phi$) & ($4\alpha$, $3\phi$) & ($4\alpha$, $4\phi$) & ($4\alpha$, $5\phi$) & ($4\alpha$, $6\phi$)
\\\midrule
\begin{minipage}[b]{0.16\columnwidth}\centering\raisebox{-.4\height}{\includegraphics[width=\linewidth]{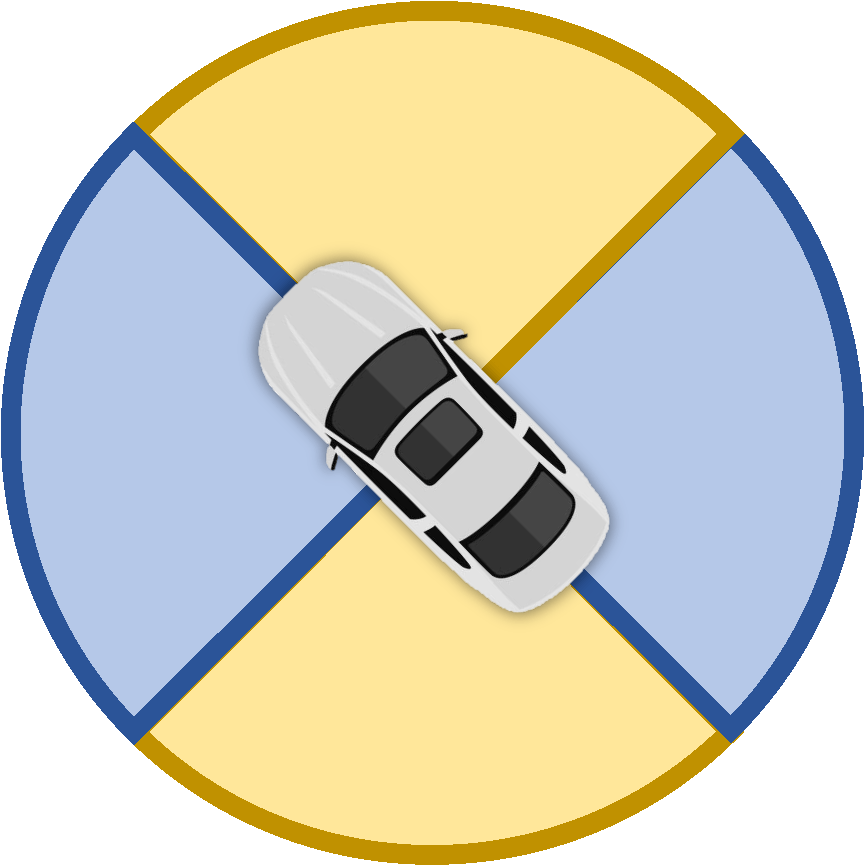}}\end{minipage} & \begin{minipage}[b]{0.16\columnwidth}\centering\raisebox{-.4\height}{\includegraphics[width=\linewidth]{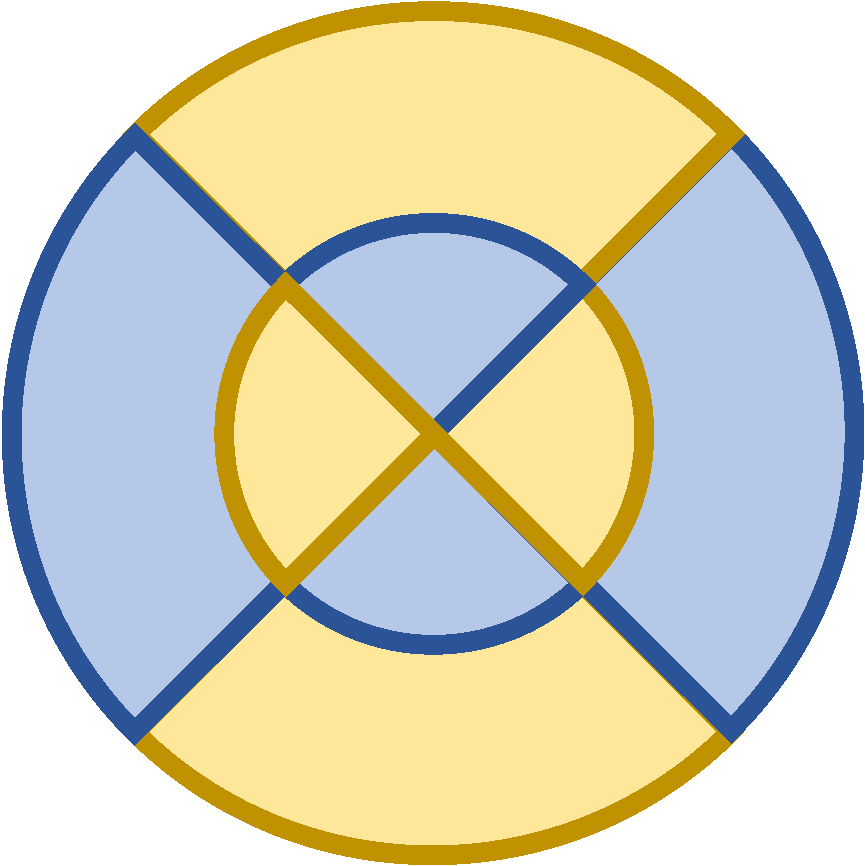}}\end{minipage} & \begin{minipage}[b]{0.16\columnwidth}\centering\raisebox{-.4\height}{\includegraphics[width=\linewidth]{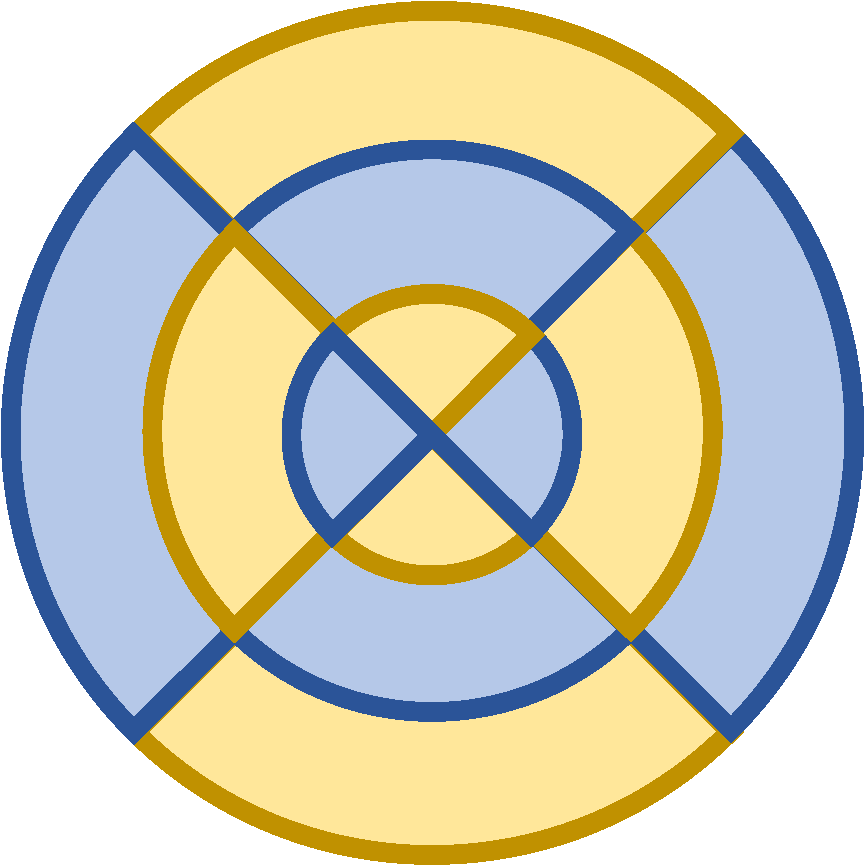}}\end{minipage} & \begin{minipage}[b]{0.16\columnwidth}\centering\raisebox{-.4\height}{\includegraphics[width=\linewidth]{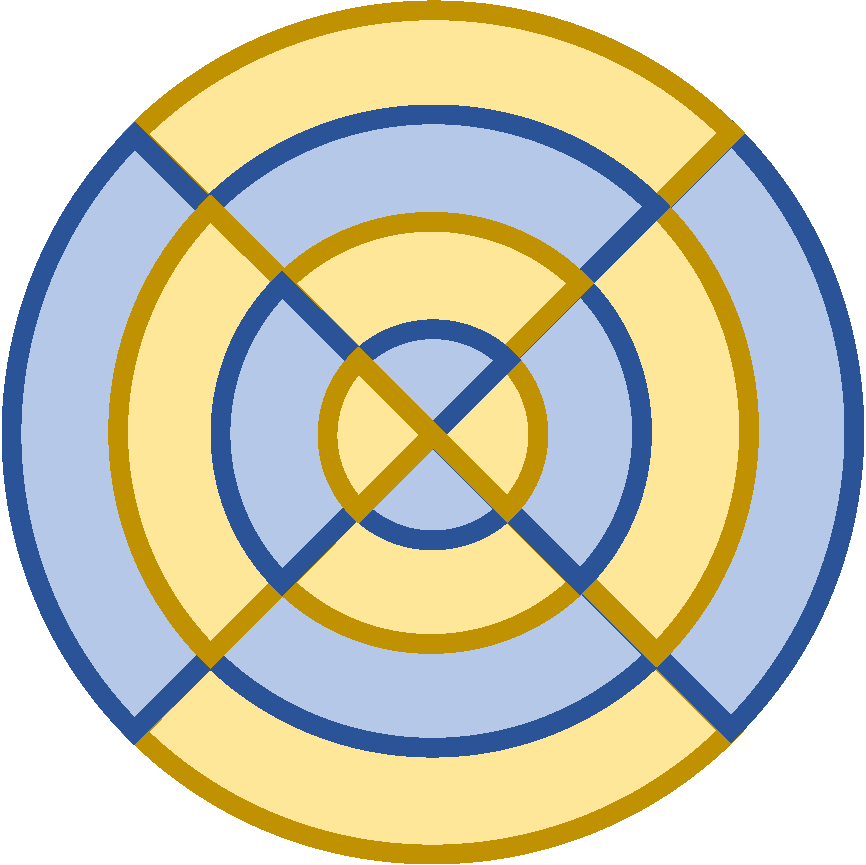}}\end{minipage} & \begin{minipage}[b]{0.16\columnwidth}\centering\raisebox{-.4\height}{\includegraphics[width=\linewidth]{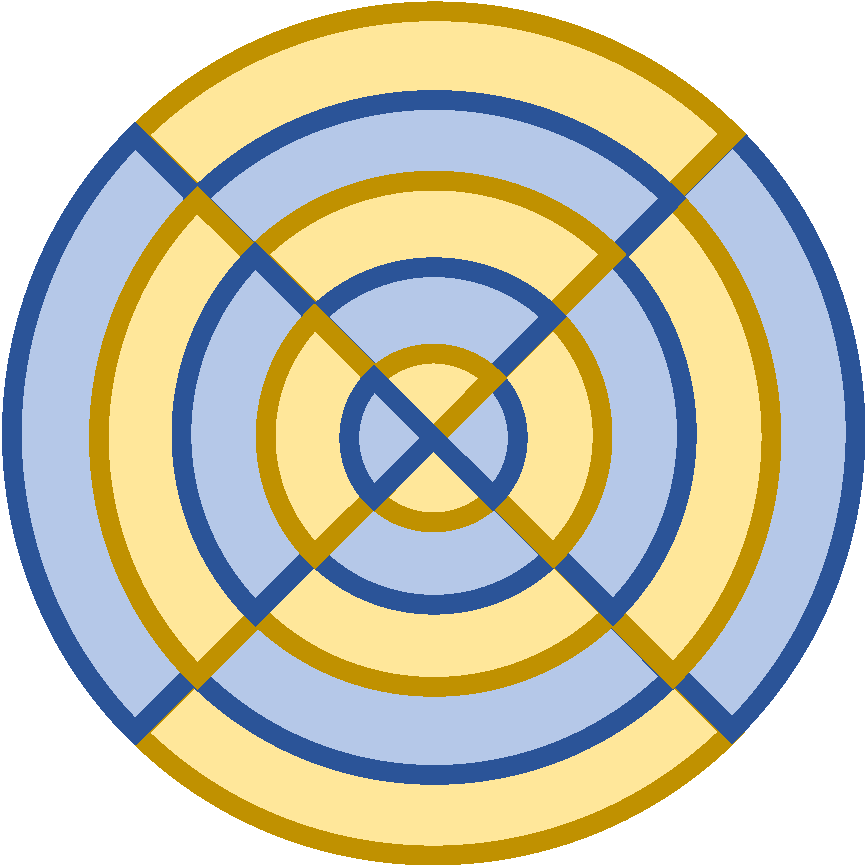}}\end{minipage} & \begin{minipage}[b]{0.16\columnwidth}\centering\raisebox{-.4\height}{\includegraphics[width=\linewidth]{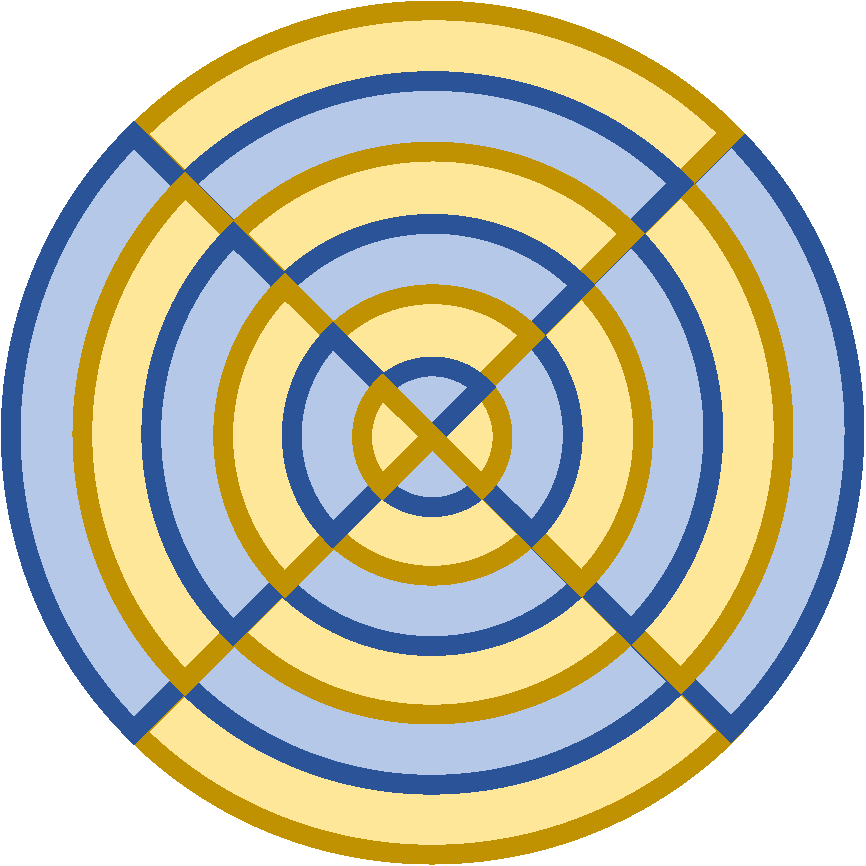}}\end{minipage} 
\\\midrule
$60.9_\mathbf{{\textcolor{brown}{(+0.6)}}}$ & $64.7_\mathbf{{\textcolor{brown}{(+4.3)}}}$ & $65.3_\mathbf{{\textcolor{brown}{(+4.9)}}}$ & $65.6_\mathbf{{\textcolor{brown}{(+5.2)}}}$ & $65.7_\mathbf{{\textcolor{brown}{(+5.3)}}}$ & $65.2_\mathbf{{\textcolor{brown}{(+4.8)}}}$ 
\\
\bottomrule
\end{tabular}}
\label{tab:granularity}
\vspace{-0.1cm}
\end{table}

\noindent\textbf{Mixing Strategies}. \cref{fig:ablation-stats} (left) compares LaserMix with other mixing methods~\cite{MixUp,CutOut,CutMix,Mix3D}. MixUp and CutMix can be considered as setting $A$ to random points and random areas, respectively. We observe that MixUp has no improvements over the baseline on average since there is no distribution pattern in random points. CutMix has a considerable improvement over the baseline, as there is always a structure prior in scene segmentation, \ie, the same semantic class points tend to cluster, which reduces the entropy in any continuous area. This prior is often used in image semantic segmentation SSL~\cite{CutMix-Seg}. However, our spatial prior is much stronger, where not only the area structure but also the area's spatial position has been considered. LaserMix outperforms CutMix by a large margin (up to $3.3\%$ mIoU) on all sets. CutOut can be considered as setting $\Xcomp$ to a dummy filling instead of sampling from datasets, and it leads to a considerable performance drop from CutMix.

\noindent\textbf{Orderless Mix}. We revert the area ordering (\ie, put the topmost laser beam at the bottom, and vice versa) in LaserMix, and the performance drops from $68.2\%$ to $64.4\%$ ($-3.8\%$ mIoU). When we shuffle the ordering, the performance drops to $63.8\%$ ($-4.4\%$ mIoU), which becomes comparable with CutMix. The results once again confirm the superiority of using spatial prior in LiDAR SSL.

\noindent\textbf{Other Heuristics}. Besides our proposed inclination partition, the LiDAR scans can also be split based on azimuth (sensor horizontal direction). Results in \cref{tab:granularity} reveal that in contrast to laser partitioning, pure azimuth splitting (the first column) does not improve the performance, which is attributed to the fact that the semantic distribution has a weak correlation in the azimuth direction. We also observe that the results tend to improve as the mixing granularity increases (row direction in \cref{tab:granularity}). The scores start to drop when the granularity is beyond a certain limit (\eg, the last two columns). We conjecture that over-fine-grained partition and mixing tend to hurt semantic coherence.

\noindent\textbf{Mix Unlabeled Data Only}. To verify that our method is more than trivially augmenting seen data and label pairs, we apply LaserMix only on unlabeled data. Instead of mixing an unlabeled scan with a labeled scan as described in \cref{sec:pipeline}, we mix two unlabeled scans with their pseudo-labels. The score drops from $68.2\%$ to $66.9\%$ ($-1.3\%$ mIoU) but still outperforms all existing methods by large margins. This verifies that LaserMix indeed brings a strong consistency regularization effect during SSL.

\noindent\textbf{EMA}. \cref{fig:ablation-stats}~(middle) provides results with different EMA decay rates, Typically, a rate between $0.9$ and $0.99$ yields the best possible results. Large rates like $0.999$ tend to hurt the consistency between the two networks. The results also verify that the teacher-student pipeline~\cite{MeanTeacher} has good synergy with our proposed SSL framework. Thanks to this simplicity, more modern SSL techniques can be easily incorporated into the current framework in future works. 

\noindent\textbf{Confidence Threshold}. As pseudo-labels play an important role in our framework, we further analyze the impact of the threshold parameter $T$ used in pseudo-label generation and show results in \cref{fig:ablation-stats}~(right). When $T$ is too low, a forced consistency to low-quality pseudo-labels tends to deteriorate the performance. When $T$ is too high, the benefits from mixing might diminish. Generally, $T$ is a dataset-dependent parameter and we find that a value around $0.9$ leads to the best possible results on the three tested LiDAR segmentation datasets~\cite{Panoptic-nuScenes,SemanticKITTI,ScribbleKITTI} in our benchmark.
\section{Conclusion}
\label{sec:conclusion}
In this work, we exploit the unique spatial prior in \lidar scenes for semi-supervised \lidar semantic segmentation. We proposed a statistically-principled and effective SSL pipeline, including LaserMix, a novel \lidar mixing technique that intertwines laser beams from different scans. Through comprehensive empirical analysis, we show the importance of spatial prior and the superiority of our approach on three popular benchmarks. The effectiveness and simplicity of our framework have shed light on the scalable deployment of the LiDAR semantic mapping system. Our future work seeks to enhance more fine-grained spatial partitions and introduce more modern SSL techniques via our proposed framework and further extend them to other related tasks, \textit{e.g.}, 3D object detection and tracking.

\noindent\textbf{Potential Negative Impacts}. Although we improve the \lidar segmentation performance in general, the label bias and out-of-domain data are not addressed or discussed in this method, which could be safety-critical issues when deploying in real-world autonomous driving applications.

\noindent\textbf{Acknowledgments}.
This study is supported by the Ministry of Education, Singapore, under its MOE AcRF Tier 2 (MOE-T2EP20221-0012), the National Research Foundation, Singapore under its AI Singapore Programme (AISG Award No: AISG2-PhD-2021-08-018), NTU NAP, and under the RIE2020 Industry Alignment Fund – Industry Collaboration Projects (IAF-ICP) Funding Initiative, as well as cash and in-kind contribution from the industry partner(s). This research is part of the programme DesCartes and is supported by the National Research Foundation, Prime Minister’s Office, Singapore under its Campus for Research Excellence and Technological Enterprise (CREATE) programme. We also thank Fangzhou Hong for the insightful discussions and feedback.

\section*{Appendix}
In this appendix, we supplement more content from the following aspects to support the findings and experimental results in the main body of this paper:
\begin{itemize}
    \item \cref{sec:lidar-representation} provides more technical details of the LiDAR range view and voxel representations.
    \item \cref{sec:case-study} gives a concrete case study on the strong spatial prior in the outdoor LiDAR data.
    \item \cref{sec:implementation-details} elaborates on additional implementation details for different SSL algorithms in our experiments.
    \item \cref{sec:additional-experimental-results} provides additional experimental results, including class-wise IoU scores (quantitative results) and visual comparisons (qualitative results).
    \item \cref{sec:public-resources} acknowledges the public resources used during the course of this work.
\end{itemize}

\section{LiDAR Representation}
\label{sec:lidar-representation}
The LiDAR data has a unique and structural format. Various representations have been proposed to better capture the internal information in LiDAR data, including raw points \cite{PointNet,RandLa-Net,KPConv}, range view (RV) \cite{RangeNet++,FIDNet,SqueezeSegV3,SalsaNext}, bird's eye view \cite{PolarNet,PolarStream}, and voxel \cite{Cylinder3D,SPVNAS} representations. This section reviews the technical details for RV projection and cylindrical voxel partition, which are currently the most efficient and the best-performing LiDAR representations, respectively.

\subsection{Range View Projection}
Given a LiDAR sensor with a fixed number (typically $32$, $64$, and $128$) of laser beams and $T$ times measurement in one scan cycle, we project LiDAR point $(p^x,p^y,p^z)$ within this scan into a matrix $x_{\text{rv}}(u,v)$ (\textit{i.e.}, range image) of size $h\times w$ via a mapping $\Pi: \mathbb{R}^{3}\mapsto\mathbb{R}^{2}$, where $h$ and $w$ are the height and width, respectively. More concretely, this can be formulated as follows:
\begin{equation}
\begin{pmatrix}
\mathit{u}  \\
\mathit{v}
\end{pmatrix}
=
\begin{pmatrix}
\frac{1}{2}~[1-\arctan(p^y,p^x)\pi^{-1}]~\mathit{w}  \\
~[1-(\arcsin(p^z,r^{-1})+\phi_{\text{down}})\xi^{-1}]~\mathit{h}~
\end{pmatrix},
\end{equation} 
where $(u,v)$ denotes the matrix grid coordinates of $x_{\text{rv}}$; $r=\sqrt{(p^x)^2+(p^y)^2+(p^z)^2}$ is the range between the point and the LiDAR sensor;  $\xi=|\phi_{\text{up}}|+|\phi_{\text{down}}|$ denotes the inclination range (also known as field-of-view or FOV) of the sensor; $\phi_{\text{up}}$ and $\phi_{\text{down}}$ are the inclinations at the upward direction and the downward direction, respectively.

Note that $h$ is set based on the number of laser beams of the LiDAR sensor, and $w$ is determined by its horizontal angular resolution. The projected range image $x_{\text{rv}}(u,v)$ serves as the input for RV-based LiDAR segmentation networks \cite{RangeNet++,SalsaNext,FIDNet}. The semantic labels are projected in the same way as $x_{\text{rv}}(u,v)$.

For range view representation, training losses are calculated on the range view predictions of size $[k, h, w]$, where $k$ denotes the number of semantic classes.


\subsection{Cylindrical Partition}
The cylinder voxels used in \cite{Cylinder3D} exhibit better segmentation performance than the conventional cubic voxels on the LiDAR data. This is because the outdoor LiDAR point clouds have varying density, which decreases as the range increases. More formally, the cylindrical partition transforms points in the Cartesian coordinate $(p^x,p^y,p^z)$ into cylinder coordinate $(\rho,\alpha,p^z)$, where $\rho$ is the distance to the origin in $X$-$Y$ plane and $\alpha$ is the azimuth in the sensor horizontal direction. The transformation can be formulated as follows:
\begin{equation}
\rho = \sqrt{(p^x)^2 + (p^y)^2}, ~~~~~~~~ \alpha = \arctan(\frac{p^y}{p^x}).
\end{equation} 

Given a predefined voxel resolution $[n_\rho, n_\alpha, n_z]$, points in the cylinder coordinate can be partitioned into the corresponding voxel cells. The semantic labels are split into partitioned cylinder voxels, where all points within the same voxel are assigned a unified label via majority voting.

For cylindrical representation, training losses are calculated on the voxel predictions of size $[k, n_\rho, n_\alpha, n_z]$, where $k$ denotes the number of semantic classes.

\section{Case Study: Spatial Prior in LiDAR Data}
\label{sec:case-study}
As mentioned in the main body of this paper, the LiDAR point clouds collected by the LiDAR sensor on top of the autonomous vehicle contain inherent spatial cues, which lead to strong patterns in laser beam partition. In this section, we conduct a case study on SemanticKITTI~\cite{SemanticKITTI} to verify our findings (see \cref{tab:spatial-prior}).

\subsection{Laser Partition}
The LiDAR scans in the SemanticKITTI~\cite{SemanticKITTI} dataset are collected by the Velodyne-HDLE64 sensor, which contains $64$ laser beams emitted isotropically around the ego-vehicle with predefined inclination angles. In this study, we split each LiDAR point cloud into eight non-overlapping areas, \textit{i.e.}, $A=\{a_1, a_2, ..., a_8\}$. Each area $a_i$ contains points captured from the consecutive $8$ laser beams.

\subsection{Spatial Prior}
As can be seen from the fourth column in \cref{tab:spatial-prior}, different semantic classes have their own behaviors in these predefined areas. Specifically, the \textit{road} class occupies mostly the first four areas (close to the ego-vehicle) while hardly appearing in the last two areas (far from the ego-vehicle). The \textit{vegetation} class and the \textit{building} class behavior conversely to \textit{road} and appear at the long-distance areas (\textit{e.g.}, $a_6$, $a_7$, $a_8$). The dynamic classes, including \textit{car}, \textit{bicyclist}, \textit{motorcyclist}, and \textit{person}, tend to appear in the middle-distance areas (\textit{e.g.}, $a_4$, $a_5$, $a_6$). Similarly, from the heatmaps shown in the fifth column in \cref{tab:spatial-prior}, we can see that these semantic classes tend to appear (lighter colors) in only certain areas. For example, the \textit{traffic-sign} class has a high likelihood to appear in the long-distance regions from the ego-vehicle (upper areas in the corresponding heatmap).

These unique distributions reflect the spatial layout of street scenes in the real world. In this work, we propose to leverage these strong spatial cues to construct our SSL framework. The experimental results verify that the spatial prior can better encourage consistency regularization in LiDAR segmentation under annotation scarcity.

\begin{table*}[t]
\caption{A case study on the \textbf{strong spatial prior} in the LiDAR data (statistics calculated from the SemanticKITTI~\cite{SemanticKITTI} dataset in this example). For each semantic class, we show its type (static or dynamic), occupation (valid $\#$ of points in percentage), distribution among eight areas ($A=\{a_1, a_2, ..., a_8\}$, \textit{i.e.}, eight laser beam groups), and the heatmap in range view (lighter colors correspond to areas that have a higher likelihood to appear and vice versa).}
\vspace{-0.1cm}
\centering\scalebox{0.78}{
\begin{tabular}{c|c|c|c|c}
\toprule
Class & Type & Proportion & Distribution & Heatmap
\\\midrule\midrule
vegetation & static & $24.825\%$ & \begin{minipage}[b]{0.66\columnwidth}\centering\raisebox{-.4\height}{\includegraphics[width=\linewidth]{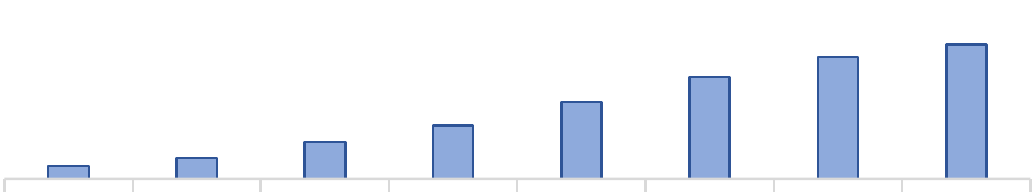}}\end{minipage} & \begin{minipage}[b]{0.95\columnwidth}\centering\raisebox{-.4\height}{\includegraphics[width=\linewidth]{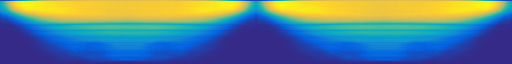}}\end{minipage}
\\\midrule
road & static & $22.545\%$ & \begin{minipage}[b]{0.66\columnwidth}\centering\raisebox{-.4\height}{\includegraphics[width=\linewidth]{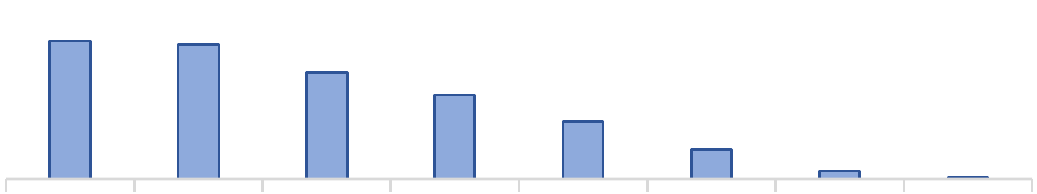}}\end{minipage} & \begin{minipage}[b]{0.95\columnwidth}\centering\raisebox{-.4\height}{\includegraphics[width=\linewidth]{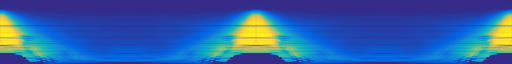}}\end{minipage}
\\\midrule
sidewalk & static & $16.353\%$ & \begin{minipage}[b]{0.66\columnwidth}\centering\raisebox{-.4\height}{\includegraphics[width=\linewidth]{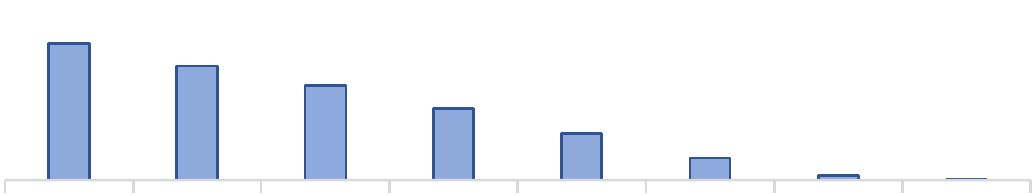}}\end{minipage} & \begin{minipage}[b]{0.95\columnwidth}\centering\raisebox{-.4\height}{\includegraphics[width=\linewidth]{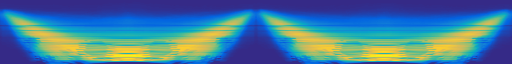}}\end{minipage}
\\\midrule
building & static & $12.118\%$ & \begin{minipage}[b]{0.66\columnwidth}\centering\raisebox{-.4\height}{\includegraphics[width=\linewidth]{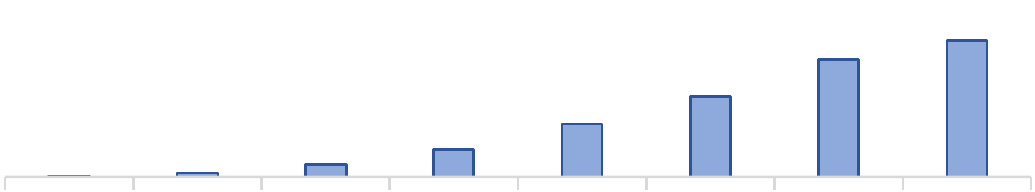}}\end{minipage} & \begin{minipage}[b]{0.95\columnwidth}\centering\raisebox{-.4\height}{\includegraphics[width=\linewidth]{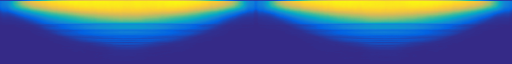}}\end{minipage}
\\\midrule
terrain & static & $8.122\%$ & \begin{minipage}[b]{0.66\columnwidth}\centering\raisebox{-.4\height}{\includegraphics[width=\linewidth]{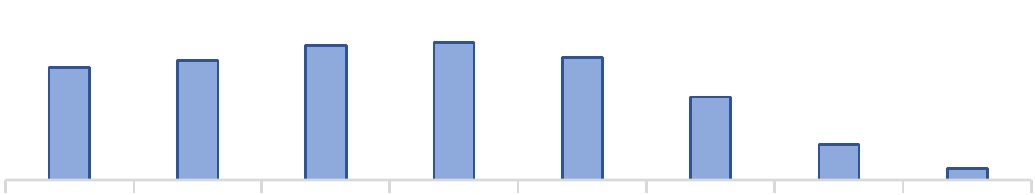}}\end{minipage} & \begin{minipage}[b]{0.95\columnwidth}\centering\raisebox{-.4\height}{\includegraphics[width=\linewidth]{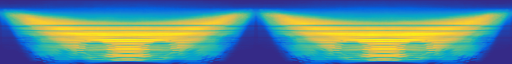}}\end{minipage}
\\\midrule
fence & static & $7.827\%$ & \begin{minipage}[b]{0.66\columnwidth}\centering\raisebox{-.4\height}{\includegraphics[width=\linewidth]{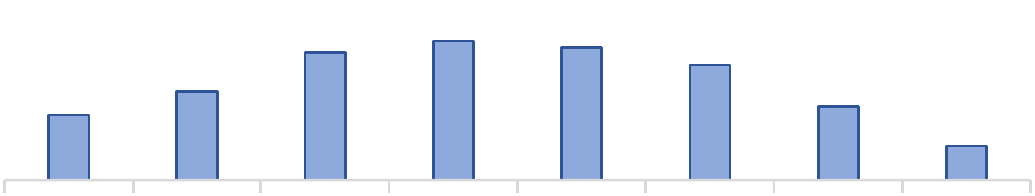}}\end{minipage} & \begin{minipage}[b]{0.95\columnwidth}\centering\raisebox{-.4\height}{\includegraphics[width=\linewidth]{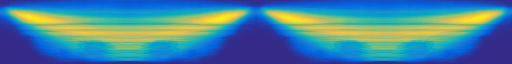}}\end{minipage}
\\\midrule
car & dynamic & $4.657\%$ & \begin{minipage}[b]{0.66\columnwidth}\centering\raisebox{-.4\height}{\includegraphics[width=\linewidth]{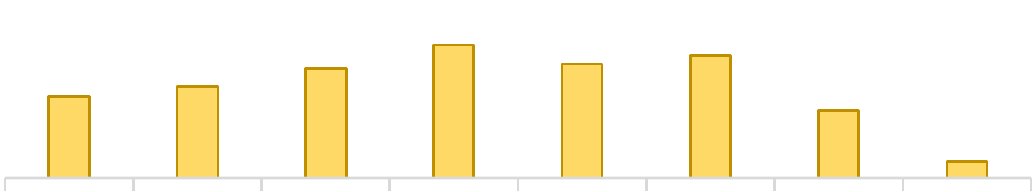}}\end{minipage} &
\begin{minipage}[b]{0.95\columnwidth}\centering\raisebox{-.4\height}{\includegraphics[width=\linewidth]{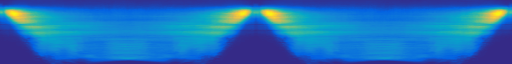}}\end{minipage}
\\\midrule
parking & static & $1.681\%$ & \begin{minipage}[b]{0.66\columnwidth}\centering\raisebox{-.4\height}{\includegraphics[width=\linewidth]{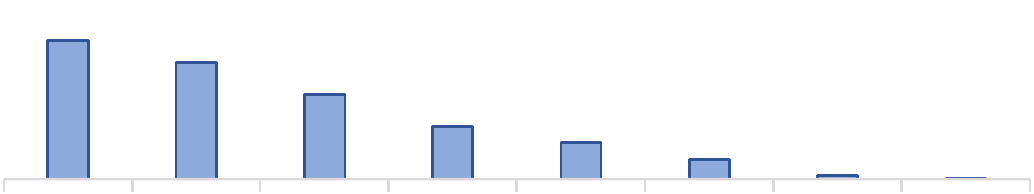}}\end{minipage} &
\begin{minipage}[b]{0.95\columnwidth}\centering\raisebox{-.4\height}{\includegraphics[width=\linewidth]{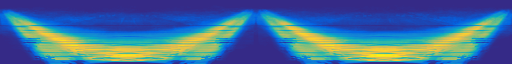}}\end{minipage}
\\\midrule
trunk & static & $0.580\%$ & \begin{minipage}[b]{0.66\columnwidth}\centering\raisebox{-.4\height}{\includegraphics[width=\linewidth]{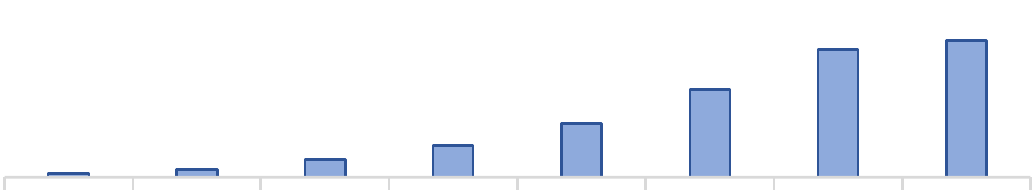}}\end{minipage} & \begin{minipage}[b]{0.95\columnwidth}\centering\raisebox{-.4\height}{\includegraphics[width=\linewidth]{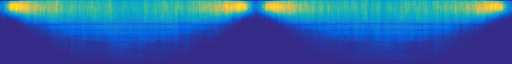}}\end{minipage}
\\\midrule
other-ground & static & $0.396\%$ & \begin{minipage}[b]{0.66\columnwidth}\centering\raisebox{-.4\height}{\includegraphics[width=\linewidth]{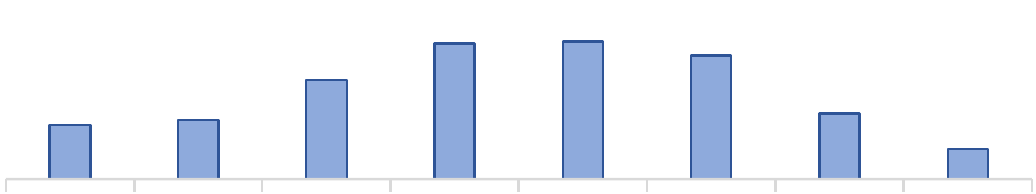}}\end{minipage} & \begin{minipage}[b]{0.95\columnwidth}\centering\raisebox{-.4\height}{\includegraphics[width=\linewidth]{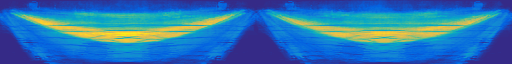}}\end{minipage}
\\\midrule
pole & static & $0.296\%$ & \begin{minipage}[b]{0.66\columnwidth}\centering\raisebox{-.4\height}{\includegraphics[width=\linewidth]{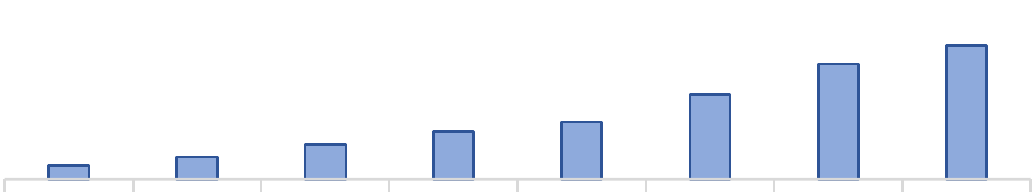}}\end{minipage} & \begin{minipage}[b]{0.95\columnwidth}\centering\raisebox{-.4\height}{\includegraphics[width=\linewidth]{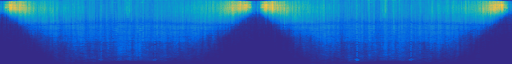}}\end{minipage}
\\\midrule
other-vehicle & dynamic & $0.229\%$ & \begin{minipage}[b]{0.66\columnwidth}\centering\raisebox{-.4\height}{\includegraphics[width=\linewidth]{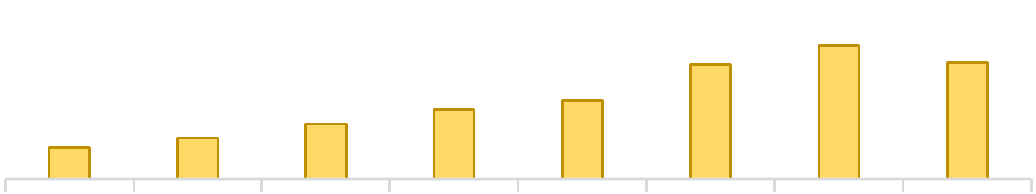}}\end{minipage} &  \begin{minipage}[b]{0.95\columnwidth}\centering\raisebox{-.4\height}{\includegraphics[width=\linewidth]{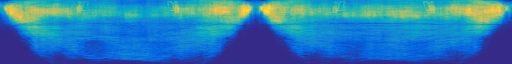}}\end{minipage}
\\\midrule
truck & dynamic & $0.193\%$ & \begin{minipage}[b]{0.66\columnwidth}\centering\raisebox{-.4\height}{\includegraphics[width=\linewidth]{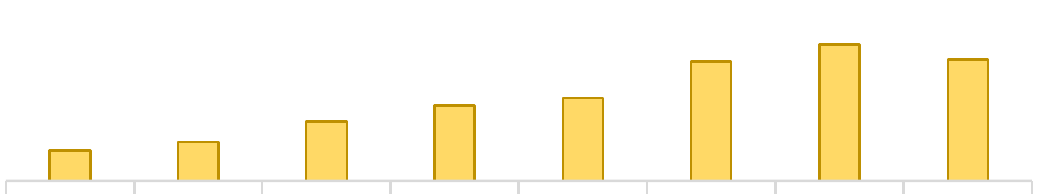}}\end{minipage} & \begin{minipage}[b]{0.95\columnwidth}\centering\raisebox{-.4\height}{\includegraphics[width=\linewidth]{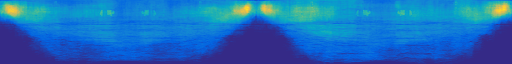}}\end{minipage}
\\\midrule
traffic-sign & static & $0.061\%$ & \begin{minipage}[b]{0.66\columnwidth}\centering\raisebox{-.4\height}{\includegraphics[width=\linewidth]{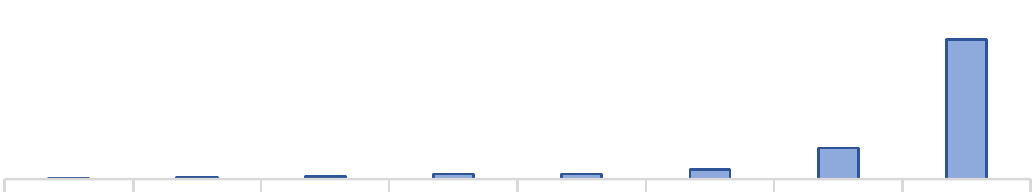}}\end{minipage} & \begin{minipage}[b]{0.95\columnwidth}\centering\raisebox{-.4\height}{\includegraphics[width=\linewidth]{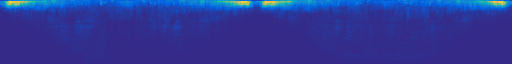}}\end{minipage}
\\\midrule
motorcycle & dynamic & $0.045\%$ & \begin{minipage}[b]{0.66\columnwidth}\centering\raisebox{-.4\height}{\includegraphics[width=\linewidth]{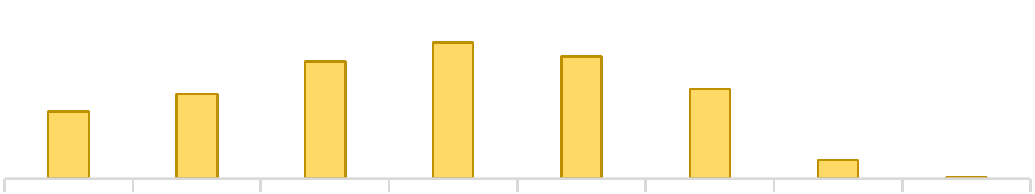}}\end{minipage} & \begin{minipage}[b]{0.95\columnwidth}\centering\raisebox{-.4\height}{\includegraphics[width=\linewidth]{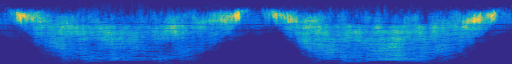}}\end{minipage}
\\\midrule
person & dynamic & $0.036\%$ & \begin{minipage}[b]{0.66\columnwidth}\centering\raisebox{-.4\height}{\includegraphics[width=\linewidth]{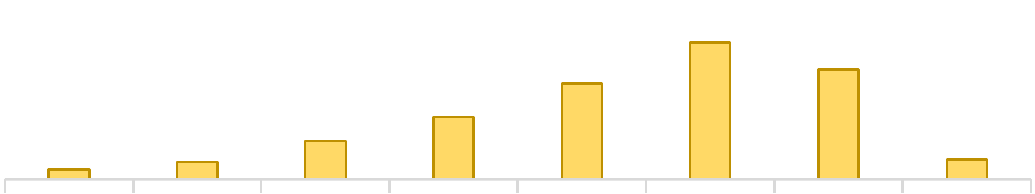}}\end{minipage} & \begin{minipage}[b]{0.95\columnwidth}\centering\raisebox{-.4\height}{\includegraphics[width=\linewidth]{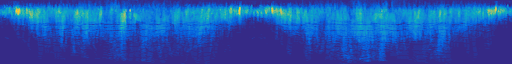}}\end{minipage}
\\\midrule
bicycle & dynamic & $0.018\%$ & \begin{minipage}[b]{0.66\columnwidth}\centering\raisebox{-.4\height}{\includegraphics[width=\linewidth]{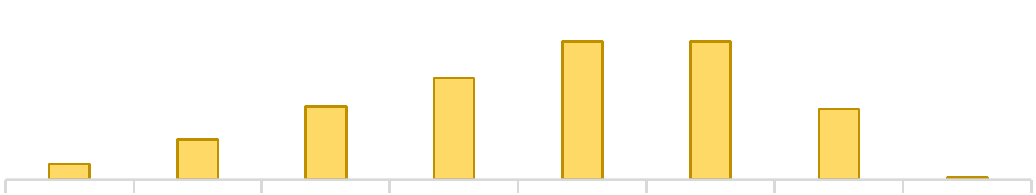}}\end{minipage} & \begin{minipage}[b]{0.95\columnwidth}\centering\raisebox{-.4\height}{\includegraphics[width=\linewidth]{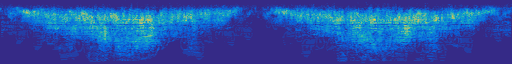}}\end{minipage}
\\\midrule
bicyclist & dynamic & $0.014\%$ & \begin{minipage}[b]{0.66\columnwidth}\centering\raisebox{-.4\height}{\includegraphics[width=\linewidth]{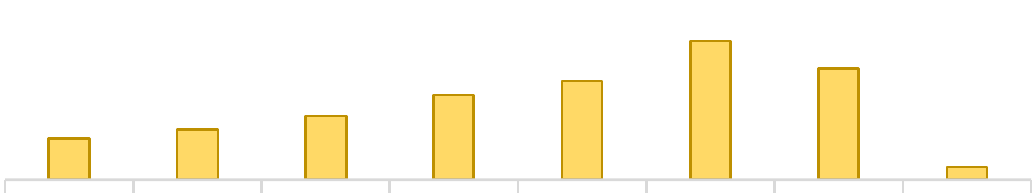}}\end{minipage} & \begin{minipage}[b]{0.95\columnwidth}\centering\raisebox{-.4\height}{\includegraphics[width=\linewidth]{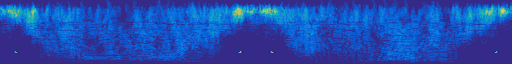}}\end{minipage}
\\\midrule
motorcyclist & dynamic & $0.004\%$ & \begin{minipage}[b]{0.66\columnwidth}\centering\raisebox{-.4\height}{\includegraphics[width=\linewidth]{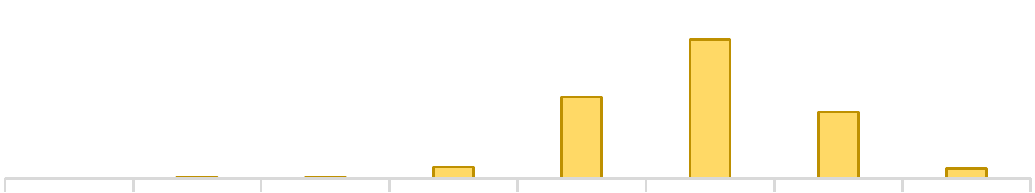}}\end{minipage} & \begin{minipage}[b]{0.95\columnwidth}\centering\raisebox{-.4\height}{\includegraphics[width=\linewidth]{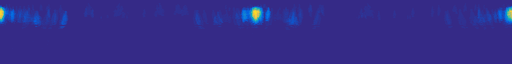}}\end{minipage}
\\
\bottomrule
\end{tabular}}
\label{tab:spatial-prior}
\end{table*}

\section{Additional Implementation Detail}
\label{sec:implementation-details}

In this section, we first compare the configuration details for the three LiDAR segmentation datasets (nuScenes~\cite{Panoptic-nuScenes}, SemanticKITTI~\cite{SemanticKITTI}, and ScribbleKITTI~\cite{ScribbleKITTI}) used in this work (see \cref{tab:datasets}). We then provide more detailed information on different SSL algorithms implemented in our semi-supervised LiDAR segmentation benchmark.

\subsection{Dataset}
\noindent\textbf{nuScenes}. As a comprehensive autonomous driving dataset, nuScenes\footnote{Refer to the \textit{lidarseg} set in nuScenes, details at \url{https://www.nuscenes.org/lidar-segmentation}.}~\cite{Panoptic-nuScenes} provides $1000$ driving scenes of $20$s duration each collected by a $32$-beam LiDAR sensor from Boston and Singapore. We follow the official \textit{train} and \textit{val} sample splittings. The total number of LiDAR scans is $40000$. The training and validation sets contain $28130$ and $6019$ scans, respectively. The semantic labels are annotated within the ranges: $p^x\in[50m, -50m]$, $p^y\in[50m, -50m]$, and $p^z\in[3m, -5m]$. Points outside the range are labeled as \textit{ignored}. The inclination range is $[10^{\circ}, -30^{\circ}]$. We use the official label mapping which contains $16$ semantic classes.

\noindent\textbf{SemanticKITTI}. Derived from the famous KITTI Vision Odometry Benchmark, SemanticKITTI~\cite{SemanticKITTI} is another large-scale LiDAR segmentation dataset widely adopted in academia. It consists of $22$ driving sequences, which are split into a \textit{train} set (Seq. $00$ to $10$, where $08$ is used for validation) and a \textit{test} set (Seq. $11$ to $21$). The LiDAR point clouds are captured from Karlsruhe, Germany, by a 64-beam LiDAR sensor. The inclination range is $[3^{\circ}, -25^{\circ}]$. We follow the official label mapping and use $19$ semantic classes in our experiments.

\noindent\textbf{ScribbleKITTI}. Efficiently annotating LiDAR point clouds is a viable solution for scaling up LiDAR segmentation. ScribbleKITTI~\cite{ScribbleKITTI} adopts scribbles to annotate SemanticKITTI~\cite{SemanticKITTI}, resulting in around $8.06\%$ semantic labels compared to the dense annotations. The other configurations are the same as SemanticKITTI~\cite{SemanticKITTI}. We use the densely annotated set (Seq. $08$ in SemanticKITTI~\cite{SemanticKITTI}) as the validation set.

In summary, we choose datasets with different numbers of laser beams (\textit{i.e.}, $32$ for nuScenes~\cite{Panoptic-nuScenes} and $64$ for SemanticKITTI~\cite{SemanticKITTI} and ScribbleKITTI~\cite{ScribbleKITTI}), different inclination ranges (\textit{i.e.}, $[10^{\circ}, -30^{\circ}]$ for nuScenes~\cite{Panoptic-nuScenes} and $[3^{\circ}, -25^{\circ}]$ for SemanticKITTI~\cite{SemanticKITTI} and ScribbleKITTI~\cite{ScribbleKITTI}), and different annotation proportions (\textit{i.e.}, $100\%$ for nuScenes~\cite{Panoptic-nuScenes} and SemanticKITTI~\cite{SemanticKITTI} and $8.06\%$ for ScribbleKITTI~\cite{ScribbleKITTI}). Our proposed SSL framework exhibit constant and evident improvements on all three datasets, which further verifies the scalability of our approaches.

\begin{table*}[t]
\caption{Configuration details for the three LiDAR segmentation datasets (nuScenes~\cite{nuScenes}, SemanticKITTI~\cite{SemanticKITTI}, and ScribbleKITTI~\cite{ScribbleKITTI}) used in this work. Rows from top to bottom: visualization examples, number of semantic classes, number of training scans, number of validation scans, resolution for range view inputs, resolution for voxel inputs, number of laser beams, inclination angle range, $x$-axis range, $y$-axis range, $z$-axis range, the proportion of semantic labels, sensor intensity examples, range examples, and semantic label examples. Images in the second row are adopted from \cite{Panoptic-nuScenes} and \cite{ScribbleKITTI}. Images in the last three rows are generated from the corresponding datasets.}
\vspace{-0.1cm}
\centering\scalebox{0.77}{
\begin{tabular}{c|c|c|c}
\toprule
& nuScenes~\cite{nuScenes} & SemanticKITTI~\cite{SemanticKITTI} & ScribbleKITTI~\cite{ScribbleKITTI}
\\\midrule
Vis. & \begin{minipage}[b]{0.762\columnwidth}\centering\raisebox{-.4\height}{\includegraphics[width=\linewidth]{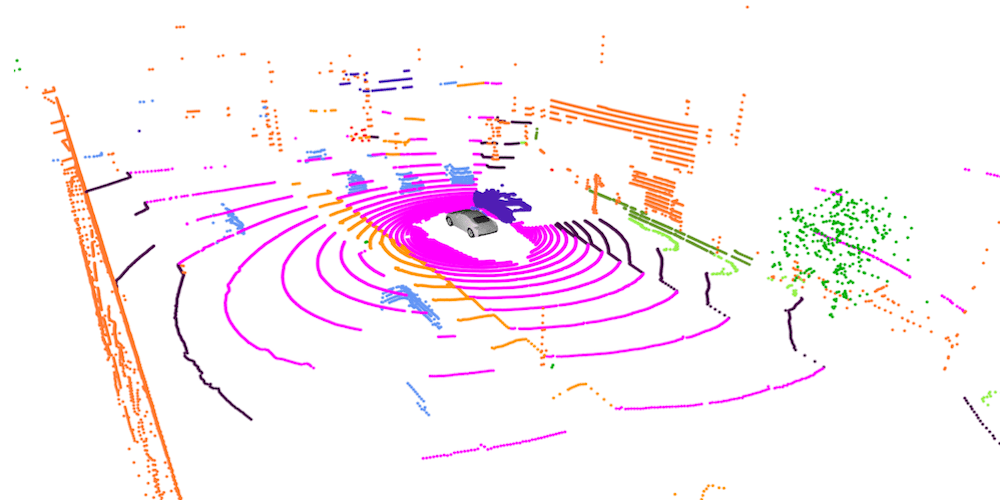}}\end{minipage} & \begin{minipage}[b]{0.762\columnwidth}\centering\raisebox{-.4\height}{\includegraphics[width=\linewidth]{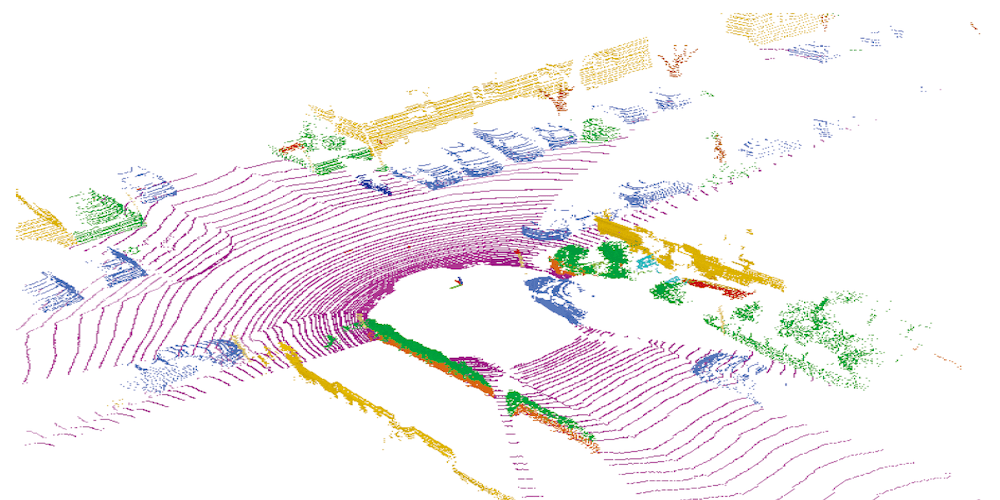}}\end{minipage} & \begin{minipage}[b]{0.762\columnwidth}\centering\raisebox{-.4\height}{\includegraphics[width=\linewidth]{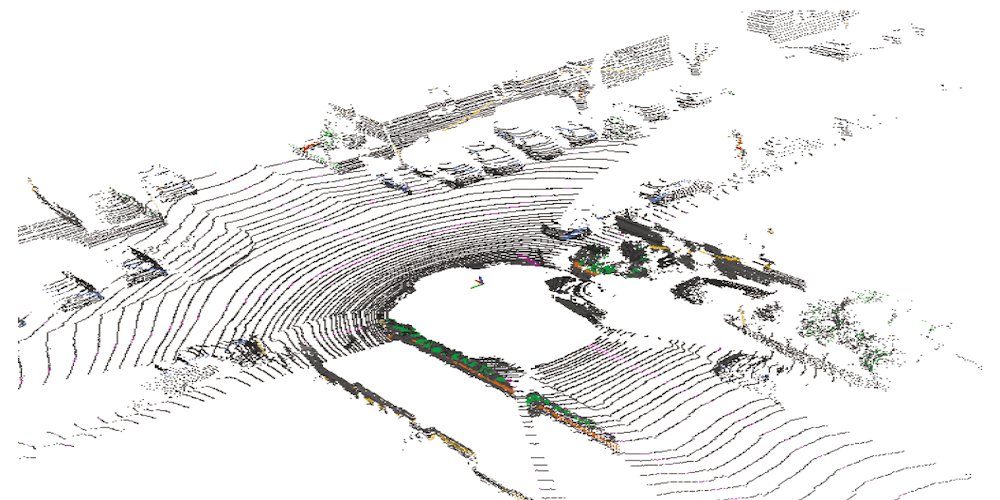}}\end{minipage}
\\\midrule
\#Class & $16$ & $19$ & $19$
\\\midrule
\#Train & $29130$ & $19130$ & $19130$
\\\midrule
\#Val & $6019$ & $4071$ & $4071$
\\\midrule
Res.~(RV) & $32\times1920$ & $64\times2048$ & $64\times2048$
\\\midrule
Res.~(voxel) & $[240, 180, 20]$ & $[240, 180, 20]$ & $[240, 180, 20]$
\\\midrule
\#Beam & $32$ & $64$ & $64$
\\\midrule
$[\phi_{\text{up}}, \phi_{\text{low}}]$ & $[10^{\circ}, -30^{\circ}]$ & $[3^{\circ}, -25^{\circ}]$ & $[3^{\circ}, -25^{\circ}]$ 
\\\midrule
$[p^x_{\text{max}}, p^x_{\text{min}}]$ & $[50m, -50m]$ & $[50m, -50m]$ & $[50m, -50m]$
\\\midrule
$[p^y_{\text{max}}, p^y_{\text{min}}]$ & $[50m, -50m]$ & $[50m, -50m]$ & $[50m, -50m]$
\\\midrule
$[p^z_{\text{max}}, p^z_{\text{min}}]$ & $[3m, -5m]$ & $[2m, -4m]$ & $[2m, -4m]$
\\\midrule
\#Label & $100\%$ & $100\%$ & $8.06\%$
\\\midrule
Intensity & \begin{minipage}[b]{0.762\columnwidth}\centering\raisebox{-.4\height}{\includegraphics[width=\linewidth]{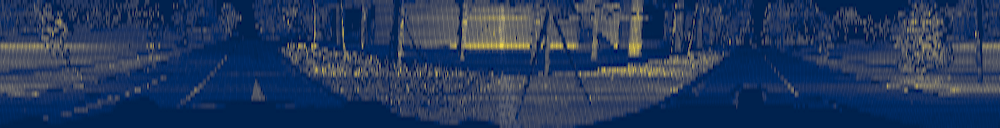}}\end{minipage} & \begin{minipage}[b]{0.762\columnwidth}\centering\raisebox{-.4\height}{\includegraphics[width=\linewidth]{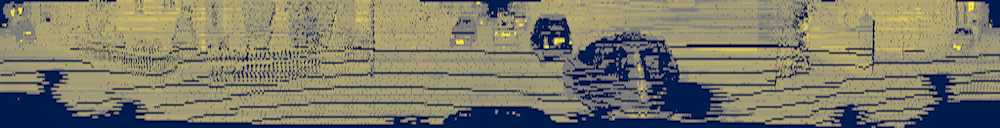}}\end{minipage} & \begin{minipage}[b]{0.762\columnwidth}\centering\raisebox{-.4\height}{\includegraphics[width=\linewidth]{figs_supp/semkitti-intensity.png}}\end{minipage}
\\\midrule
Range & \begin{minipage}[b]{0.762\columnwidth}\centering\raisebox{-.4\height}{\includegraphics[width=\linewidth]{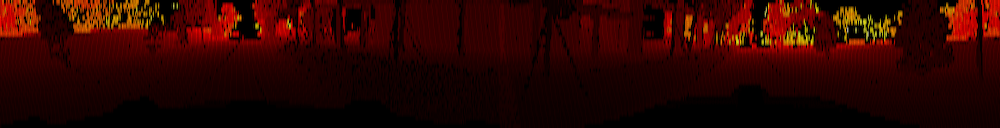}}\end{minipage} & \begin{minipage}[b]{0.762\columnwidth}\centering\raisebox{-.4\height}{\includegraphics[width=\linewidth]{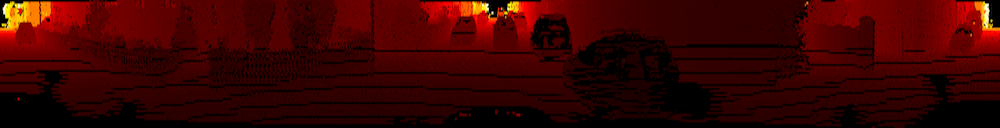}}\end{minipage} & \begin{minipage}[b]{0.762\columnwidth}\centering\raisebox{-.4\height}{\includegraphics[width=\linewidth]{figs_supp/semkitti-range.png}}\end{minipage}
\\\midrule
Semantics & \begin{minipage}[b]{0.762\columnwidth}\centering\raisebox{-.4\height}{\includegraphics[width=\linewidth]{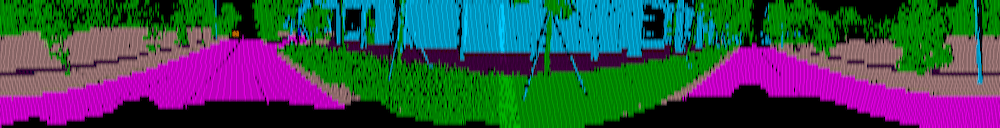}}\end{minipage} & \begin{minipage}[b]{0.762\columnwidth}\centering\raisebox{-.4\height}{\includegraphics[width=\linewidth]{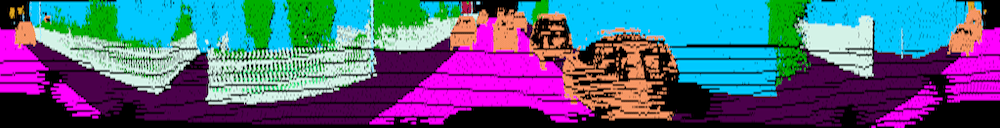}}\end{minipage} & \begin{minipage}[b]{0.762\columnwidth}\centering\raisebox{-.4\height}{\includegraphics[width=\linewidth]{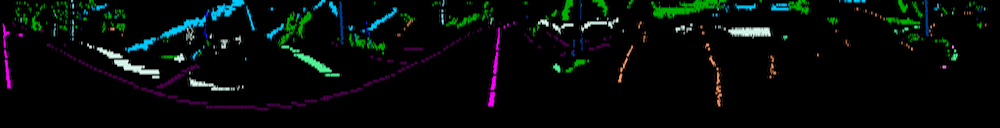}}\end{minipage}
\\
\bottomrule
\end{tabular}}
\vspace{0.1cm}
\label{tab:datasets}
\end{table*}

\subsection{Model Configuration}
\noindent\textbf{FIDNet}. We use the \textit{ResNet34-point} variant in FIDNet~\cite{FIDNet} as our \textit{range view} segmentation backbone. It contains fewer parameters ($6.05$M) than the one used in the original paper ($19.64$M) while still maintaining good segmentation performance: $58.8\%$ mIoU (compared to $59.5\%$ mIoU) on the \textit{val} set of SemanticKITTI~\cite{SemanticKITTI}, and $71.6\%$ mIoU (compared to $72.3\%$ mIoU) on the \textit{val} set of nuScenes~\cite{Panoptic-nuScenes}. We refer to the FIDNet~\cite{FIDNet} paper for more details on the model architecture and other related configurations.

\noindent\textbf{Cylinder3D}. We use a more compact version of Cylinder3D~\cite{Cylinder3D} as the \textit{voxel} segmentation backbone in our experiments, which has $28.13$M parameters (compared to $56.26$M for the one used in the original paper). We also use a smaller voxel resolution ($[240, 180, 20]$) compared to the original configuration ($[480, 360, 32]$). This saves around $4\times$ memory consumption and further helps to speed up training. We found that with the smaller resolution (larger voxel size), the performance drops from $76.1\%$ mIoU to $74.1\%$ mIoU on the \textit{val} set of nuScenes~\cite{Panoptic-nuScenes}. We refer to the Cylinder3D~\cite{Cylinder3D} paper for more details on the model architecture and other related configurations.

\noindent\textbf{Training Configurations}. All SSL algorithms implemented in this work share the same LiDAR segmentation backbones, \textit{i.e.}, FIDNet~\cite{FIDNet} for the LiDAR \textit{range view} representation and Cylinder3D~\cite{Cylinder3D} for the LiDAR \textit{voxel} representation. For both FIDNet and Cylinder3D, we adopt AdamW~\cite{AdamW} as the optimizer and use the OneCycle learning rate scheduler~\cite{OneCycle}. The maximum learning rate is $0.0025$ for FIDNet and $0.001$ for Cylinder3D. The batch size for the LiDAR \textit{range view} representation is $10$ for nuScenes and $4$ for SemanticKITTI and ScribbleKITTI. The batch size for the LiDAR \textit{voxel} representation is $8$ for nuScenes and $4$ for SemanticKITTI and ScribbleKITTI.

\noindent\textbf{Data Augmentation}. The data augmentations used for the \textit{range view} inputs for all SSL algorithms include random jittering, scaling, flipping (for nuScenes), and shifting (for SemanticKITTI and ScribbleKITTI). The data augmentations used for the \textit{voxel} inputs for all SSL algorithms include random rotation and flipping (for nuScenes, SemanticKITTI, and ScribbleKITTI).

\noindent\textbf{Other Configurations}. For LaserMix, the number of spatial areas is uniformly sampled from $1$ to $6$. The weight $\lambda_{\text{mix}}$ is set as $1$ for all three datasets. The weight $\lambda_{\text{mt}}$ is set as $1$e$3$ for nuScenes and $2$e$3$ for SemanticKITTI and ScribbleKITTI. For CPS~\cite{CPS}, the weight $\lambda_{\text{cps}}$ is set as $1$ for all three datasets. We tried $2$ and $6$ and found $1$ yielded the best results. For MeanTeacher~\cite{MeanTeacher}, the weight $\lambda_{\text{mt}}$ is set as $1$e$3$ for nuScenes and $2$e$3$ for SemanticKITTI and ScribbleKITTI. For CutMix-Seg~\cite{CutMix-Seg}, the weight $\lambda_{\text{cons}}$ is set as $1$ which is the same as the original paper. For CBST~\cite{CBST}, we use the \textit{sup.-only} checkpoints to generate the pseudo-labels and then train the segmentation network from scratch with the pseudo-labels. We refer to the original papers for the aforementioned algorithms \cite{CPS,MeanTeacher,CutMix-Seg,CBST} for additional technical or implementation details.

\noindent\textbf{GPC Split}. 
In the main body, we compared our approach with GPC \cite{GPC}, a 3D SSL method using contrastive learning on point clouds. Since this model is not open-sourced, we directly use the scores reported in their paper for comparison, which might involve factors that are not aligned, \textit{e.g.}, different backbones and data splits. To better align the benchmark settings, we form a sequential track in our codebase\footnote{\url{https://github.com/ldkong1205/LaserMix}.} taking into account the LiDAR data collection nature. Kindly refer to our benchmark for more details on this track.

\begin{table}[t]
\caption{Benchmarking results on the \textit{val} set of Cityscapes~\cite{Cityscapes}.}
\vspace{-0.2cm}
\label{tab:benchmark-cityscapes}
\setlength{\tabcolsep}{4pt}
\centering\scalebox{0.82}{
\begin{tabular}{c|cccc}
\toprule
\textbf{Method} & $1/16$ & $1/8$ & $1/4$ & $1/2$ 
\\\midrule
\cellcolor{yellow!12.5}MeanTeacher~\cite{MeanTeacher} & \cellcolor{yellow!12.5}$66.1$ & \cellcolor{yellow!12.5}$71.2$ & \cellcolor{yellow!12.5}$74.4$ & \cellcolor{yellow!12.5}$76.3$
\\\midrule
\textbf{\textit{w/} Ours} & $68.7$ & $72.3$ & $75.7$ & $76.8$
\\
 $\Delta$ $\uparrow$ & \textcolor{brown}{\small{$\mathbf{+2.6}$}} & \textcolor{brown}{\small{$\mathbf{+1.1}$}} & \textcolor{brown}{\small{$\mathbf{+1.3}$}} & \textcolor{brown}{\small{$\mathbf{+0.5}$}}
\\\midrule
CCT~\cite{CCT} & $66.4$ & $72.5$ & $75.7$ & $76.8$
\\
GCT~\cite{GCT} & $65.8$ & $71.3$ & $75.3$ & $77.1$
\\
CPS~\cite{CPS} & $69.8$ & $74.4$ & $76.9$ & $78.6$
\\\midrule
\cellcolor{yellow!12.5}CPS-CutMix~\cite{CPS} & \cellcolor{yellow!12.5}$74.5$ & \cellcolor{yellow!12.5}$76.6$ & \cellcolor{yellow!12.5}$77.8$ & \cellcolor{yellow!12.5}$78.8$
\\\midrule
\textbf{\textit{w/} Ours} & $75.5$ & $77.1$ & $78.3$ & $79.1$
\\
$\Delta$ $\uparrow$ & \textcolor{brown}{\small{$\mathbf{+1.0}$}} & \textcolor{brown}{\small{$\mathbf{+0.5}$}} & \textcolor{brown}{\small{$\mathbf{+0.5}$}} & \textcolor{brown}{\small{$\mathbf{+0.3}$}}
\\\bottomrule
\end{tabular}}
\label{tab:cityscapes}
\end{table}

\section{Additional Experimental Result}
\label{sec:additional-experimental-results}

In this section, we provide the class-wise IoU results for our comparative studies and ablation studies in the main body of this paper. Since our proposed SSL framework is a generic design, we also include the benchmarking results on Cityscapes~\cite{Cityscapes} to further verify our generalizability on structural RGB data. To provide more qualitative comparisons, we attach a video demo containing visualizations from the \textit{val} set of SemanticKITTI~\cite{SemanticKITTI}.

\noindent\textbf{Comparative Study}. \cref{table:class-nuScenes-val}, \cref{table:class-SemKitti-val}, and \cref{table:class-ScribbleKitti-val} provide the class-wise IoU scores for different SSL algorithms on the \textit{val} set of nuScenes~\cite{Panoptic-nuScenes}, SemanticKITTI~\cite{SemanticKITTI}, and ScribbleKITTI~\cite{ScribbleKITTI}, respectively. For almost all semantic classes, we observe overt improvements from LaserMix. This can be credited to the strong consistency regularization encouraged by our SSL framework.

\noindent\textbf{Ablation Study}. \cref{table:class-ablation-inclination-rv} and \cref{table:class-ablation-inclination-voxel} provide the class-wise IoU scores for the granularity studies of the LiDAR \textit{range view} and \textit{voxel} representations, respectively. Among different LaserMix strategies, we find that increasing the granularity along inclination tends to yield better segmentation performance. In our benchmarking experiments, we combine different strategies together by uniformly sampling the number of spatial areas. This simple ensembling further increases diversity and provides higher segmentation scores.

\noindent\textbf{Extension to RGB Data}. To further attest to the scalability of our proposed spatial-prior SSL framework, we conduct experiments on Cityscapes~\cite{Cityscapes}, which contains structural RGB images collected from street scenes. We follow the data split from recent work \cite{CPS} and show the results in \cref{tab:cityscapes}. Since the images from this dataset also contain strong spatial cues, the mixing strategy used here is similar to that for the LiDAR \textit{range view} representation, \textit{i.e.}, partitioning areas along the image vertical direction. We combine our proposed $\mathcal{L}_{\text{mix}}$ with MeanTeacher~\cite{MeanTeacher} ($\mathcal{L}_{\text{mt}}$) and CPS~\cite{CPS} ($\mathcal{L}_{\text{cps}}$). The results verify that our SSL framework can also encourage consistency for image data. For all four splits, our approaches constantly improve the segmentation performance on top of the SoTA methods~\cite{MeanTeacher,CPS}.

\begin{figure*}[t]
    \begin{center}
    \includegraphics[width=0.985\linewidth]{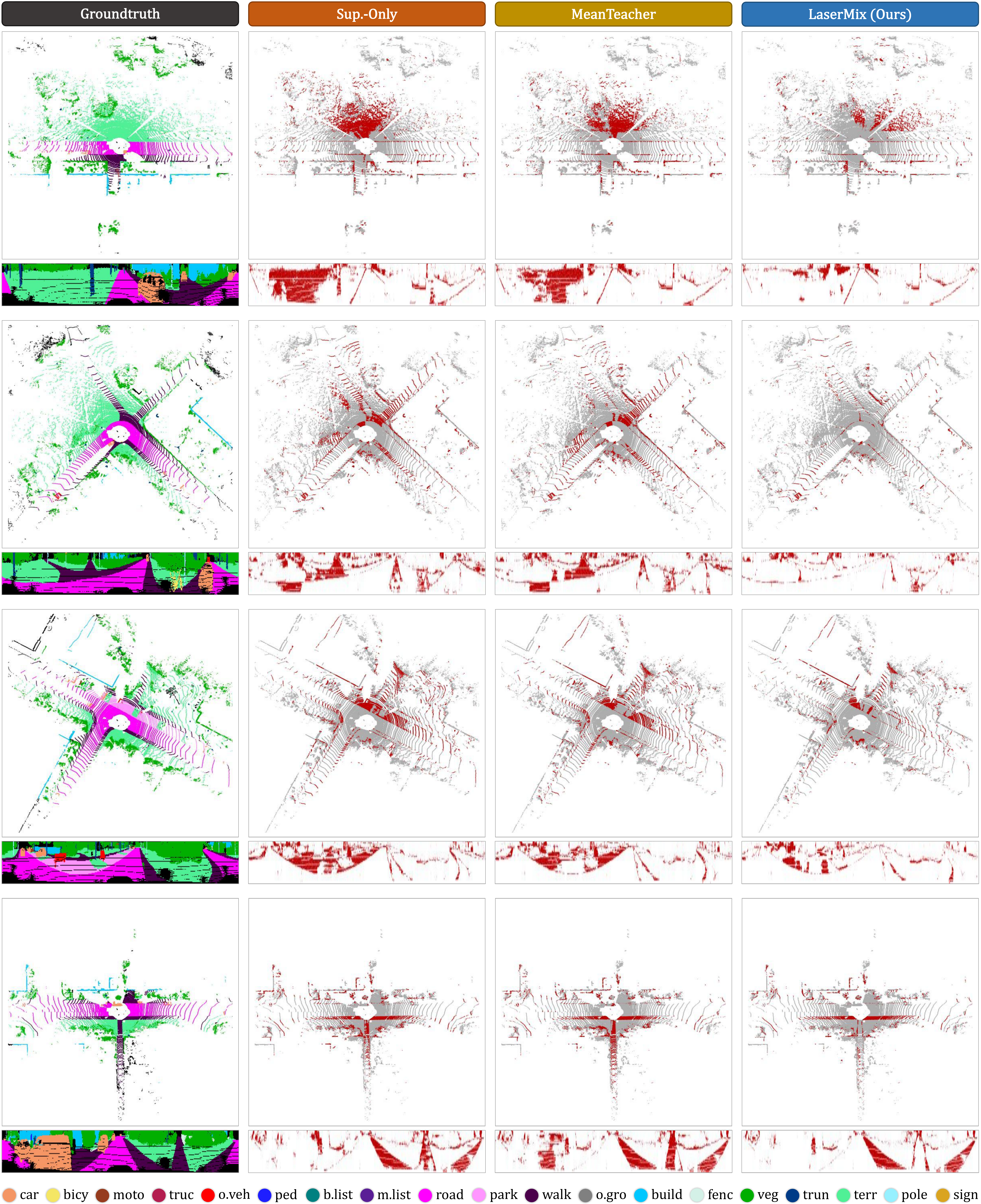}
    \end{center}
    \vspace{-0.45cm}
    \caption{\textbf{Additional qualitative results (error maps)}. Models are trained with $1\%$ labeled data on SemanticKITTI~\cite{SemanticKITTI}. To highlight the differences, the \textbf{\textcolor{correct}{correct}} and \textbf{\textcolor{incorrect}{incorrect}} predictions are painted in \textbf{\textcolor{correct}{gray}} and \textbf{\textcolor{incorrect}{red}}, respectively. Each scene is visualized from the bird's eye view (top) and range view (bottom) and covers a region of size $50$m by $50$m, centered around the ego-vehicle. Best viewed in colors.}
    \label{figure:qualitative_supp_01}
    \vspace{-0.cm}
\end{figure*}

\begin{figure*}[t]
    \begin{center}
    \includegraphics[width=0.985\linewidth]{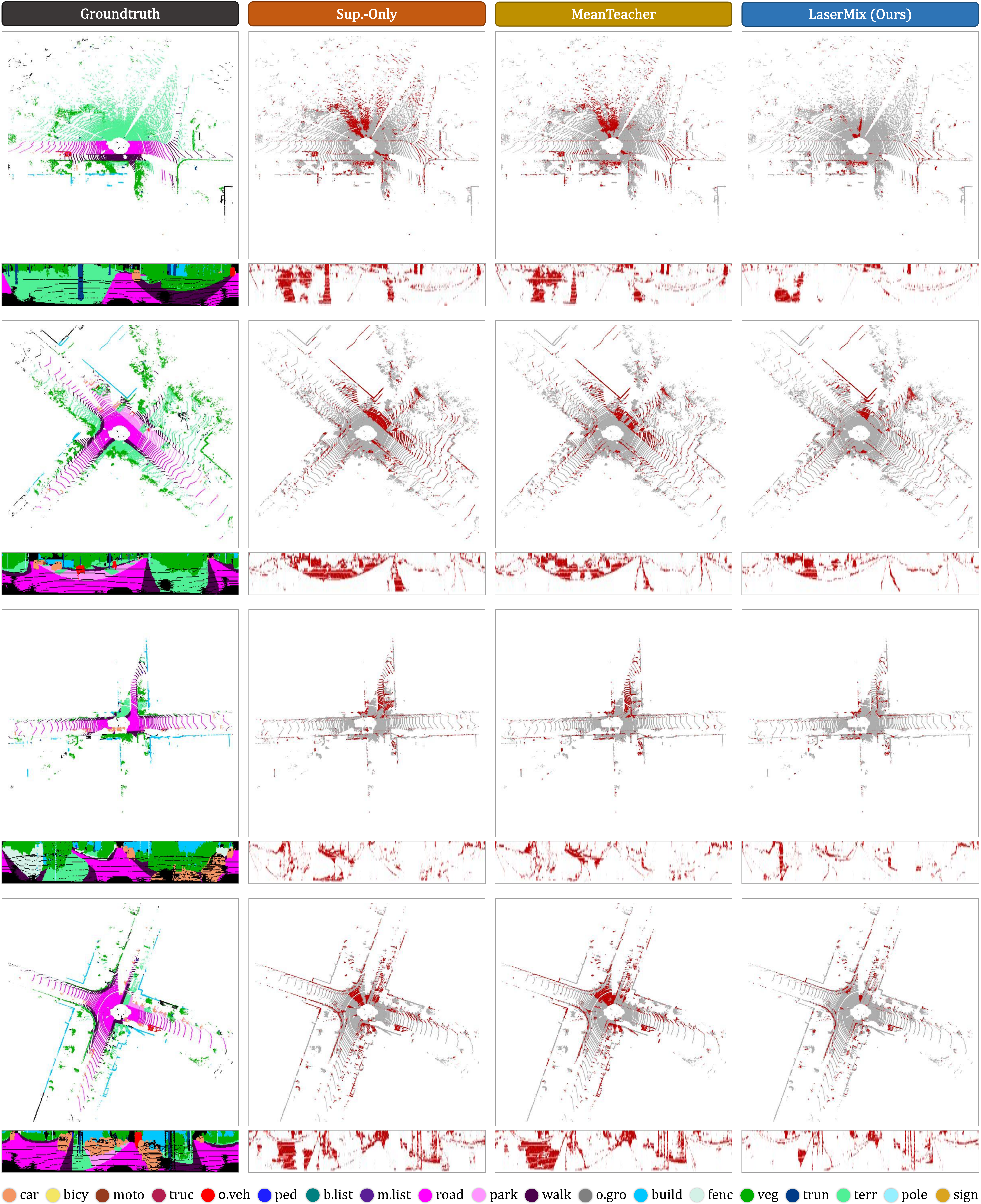}
    \end{center}
    \vspace{-0.45cm}
    \caption{\textbf{Additional qualitative results (error maps)}. Models are trained with $1\%$ labeled data on SemanticKITTI~\cite{SemanticKITTI}. To highlight the differences, the \textbf{\textcolor{correct}{correct}} and \textbf{\textcolor{incorrect}{incorrect}} predictions are painted in \textbf{\textcolor{correct}{gray}} and \textbf{\textcolor{incorrect}{red}}, respectively. Each scene is visualized from the bird's eye view (top) and range view (bottom) and covers a region of size $50$m by $50$m, centered around the ego-vehicle. Best viewed in colors.}
    \label{figure:qualitative_supp_02}
    \vspace{-0.cm}
\end{figure*}

\begin{figure*}[t]
    \begin{center}
    \includegraphics[width=0.984\linewidth]{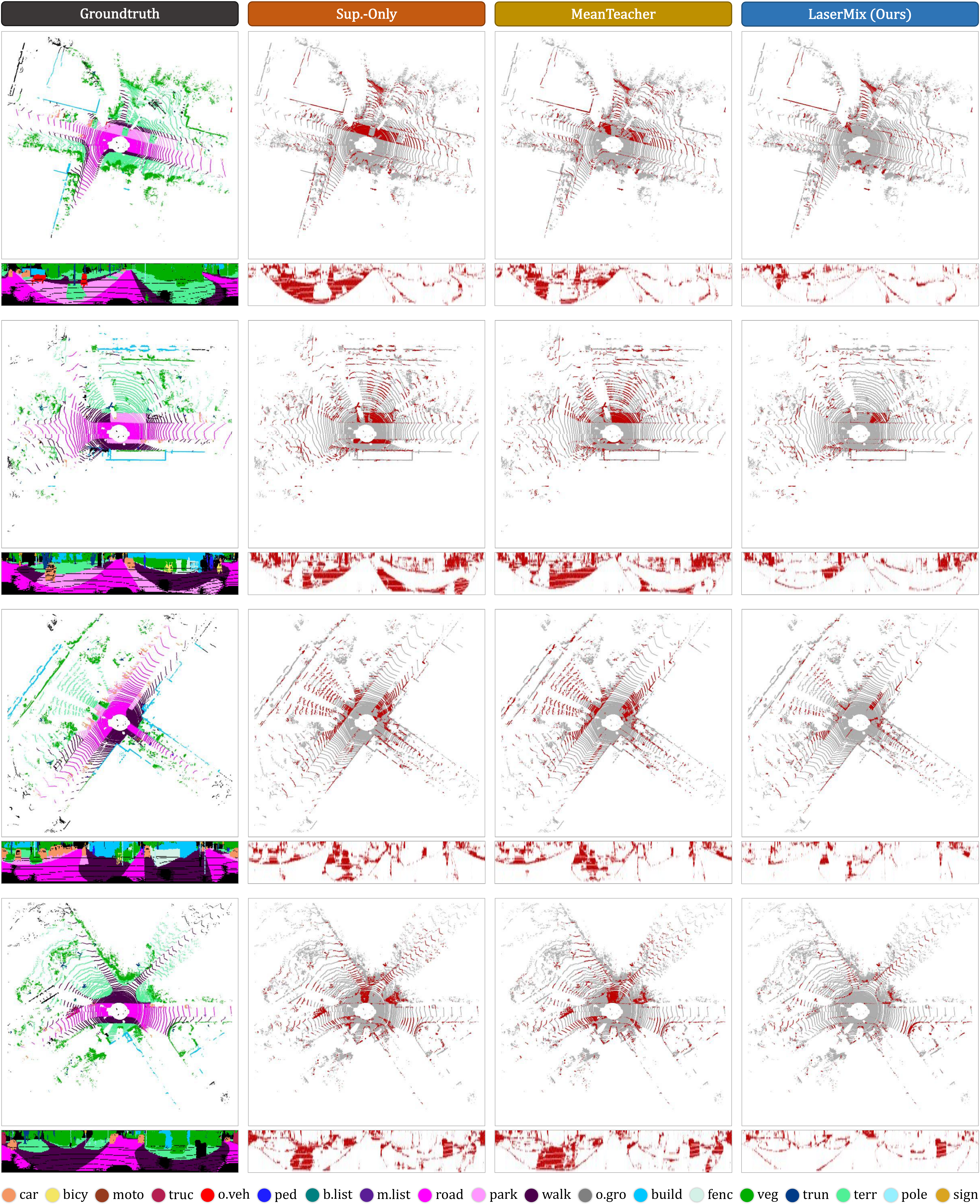}
    \end{center}
    \vspace{-0.45cm}
    \caption{\textbf{Additional qualitative results (error maps)}. Models are trained with $1\%$ labeled data on SemanticKITTI~\cite{SemanticKITTI}. To highlight the differences, the \textbf{\textcolor{correct}{correct}} and \textbf{\textcolor{incorrect}{incorrect}} predictions are painted in \textbf{\textcolor{correct}{gray}} and \textbf{\textcolor{incorrect}{red}}, respectively. Each scene is visualized from the bird's eye view (top) and range view (bottom) and covers a region of size $50$m by $50$m, centered around the ego-vehicle. Best viewed in colors.}
    \label{figure:qualitative_supp_03}
    \vspace{-0.cm}
\end{figure*}

\begin{figure*}[t]
    \begin{center}
    \includegraphics[width=0.985\linewidth]{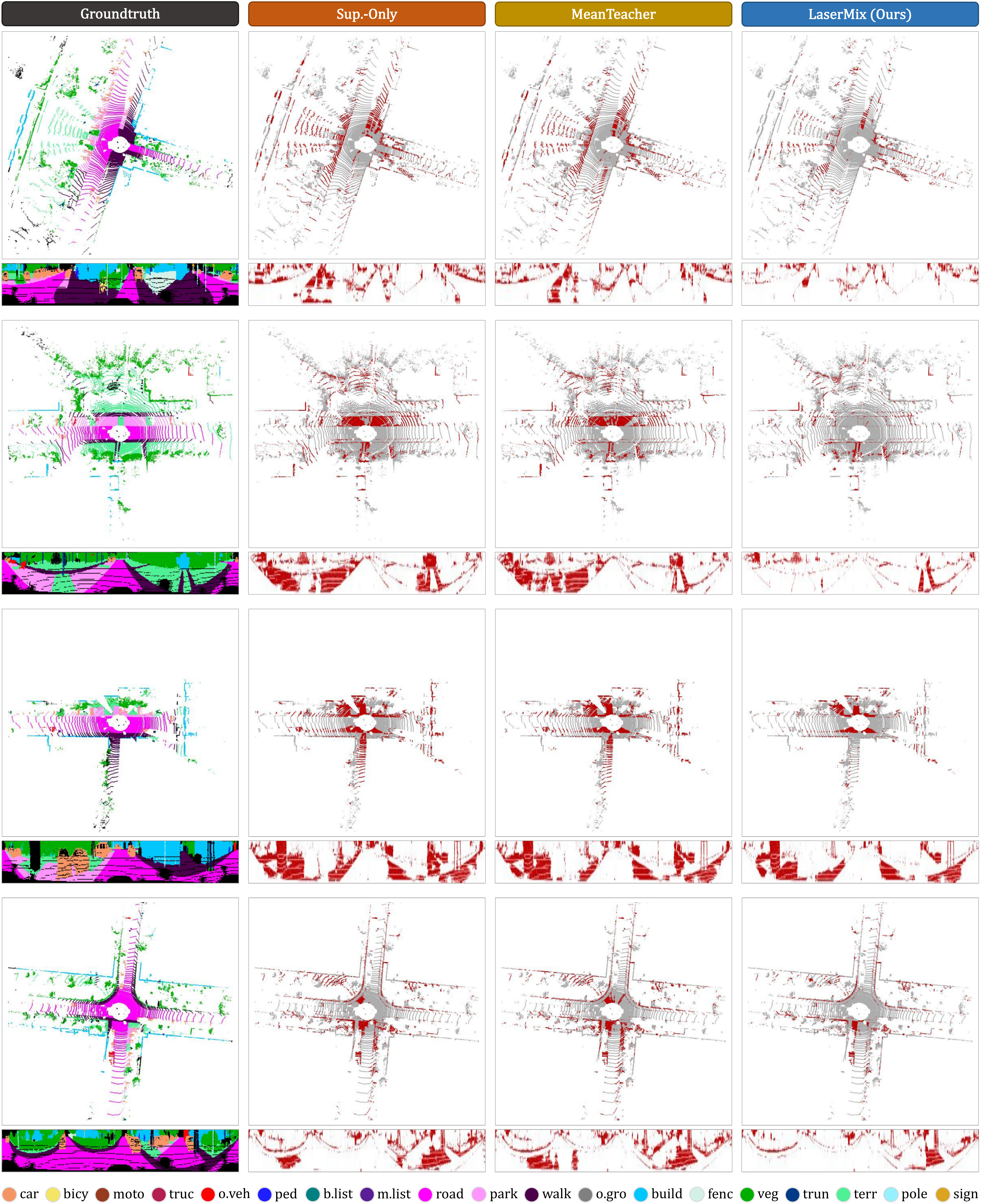}
    \end{center}
    \vspace{-0.45cm}
    \caption{\textbf{Additional qualitative results (error maps)}. Models are trained with $1\%$ labeled data on SemanticKITTI~\cite{SemanticKITTI}. To highlight the differences, the \textbf{\textcolor{correct}{correct}} and \textbf{\textcolor{incorrect}{incorrect}} predictions are painted in \textbf{\textcolor{correct}{gray}} and \textbf{\textcolor{incorrect}{red}}, respectively. Each scene is visualized from the bird's eye view (top) and range view (bottom) and covers a region of size $50$m by $50$m, centered around the ego-vehicle. Best viewed in colors.}
    \label{figure:qualitative_supp_04}
    \vspace{-0.cm}
\end{figure*}

\begin{figure*}[t]
    \begin{center}
    \includegraphics[width=0.985\linewidth]{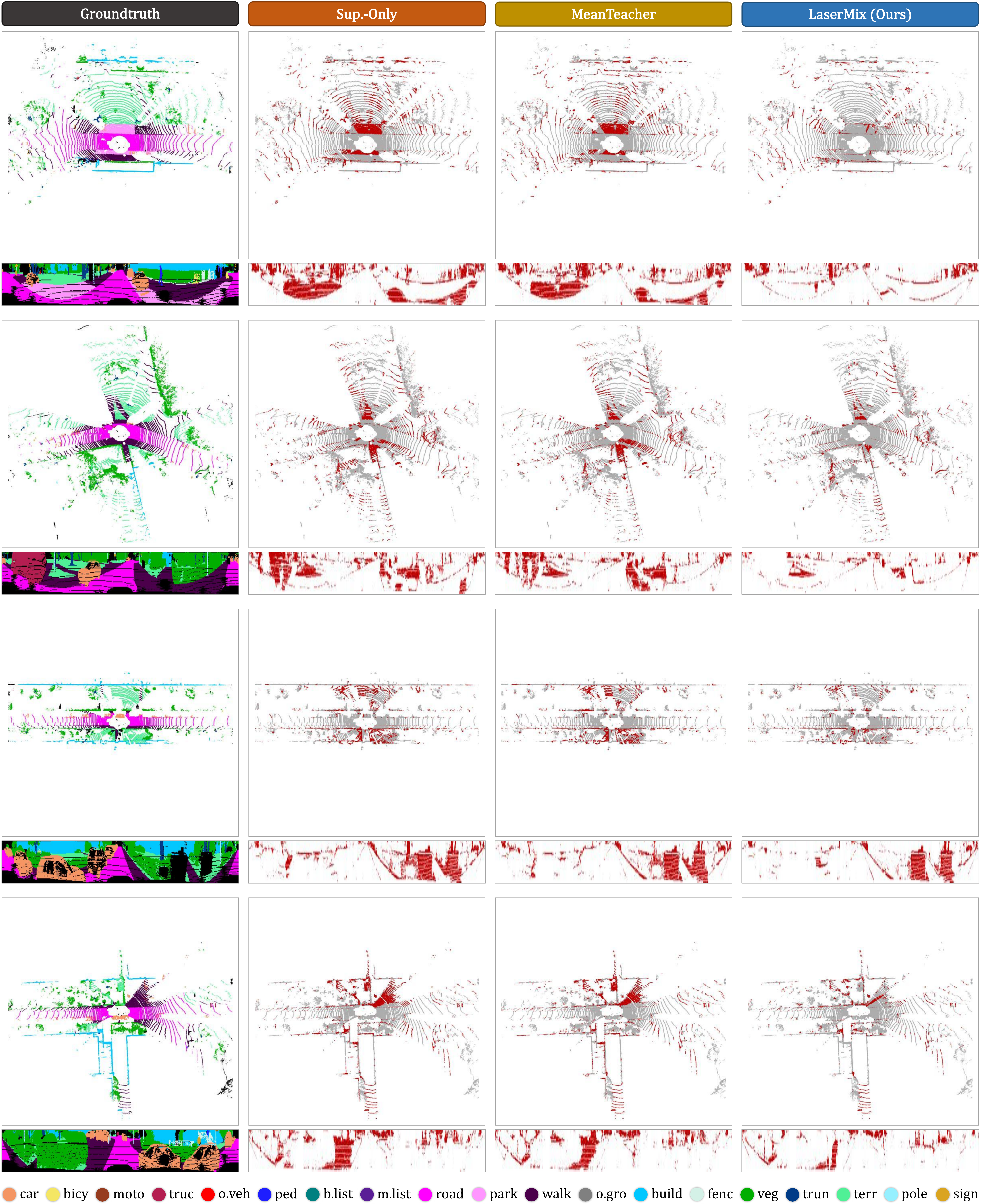}
    \end{center}
    \vspace{-0.45cm}
    \caption{\textbf{Additional qualitative results (error maps)}. Models are trained with $1\%$ labeled data on SemanticKITTI~\cite{SemanticKITTI}. To highlight the differences, the \textbf{\textcolor{correct}{correct}} and \textbf{\textcolor{incorrect}{incorrect}} predictions are painted in \textbf{\textcolor{correct}{gray}} and \textbf{\textcolor{incorrect}{red}}, respectively. Each scene is visualized from the bird's eye view (top) and range view (bottom) and covers a region of size $50$m by $50$m, centered around the ego-vehicle. Best viewed in colors.}
    \label{figure:qualitative_supp_05}
    \vspace{-0.cm}
\end{figure*}

\begin{figure*}[t]
    \begin{center}
    \includegraphics[width=0.985\linewidth]{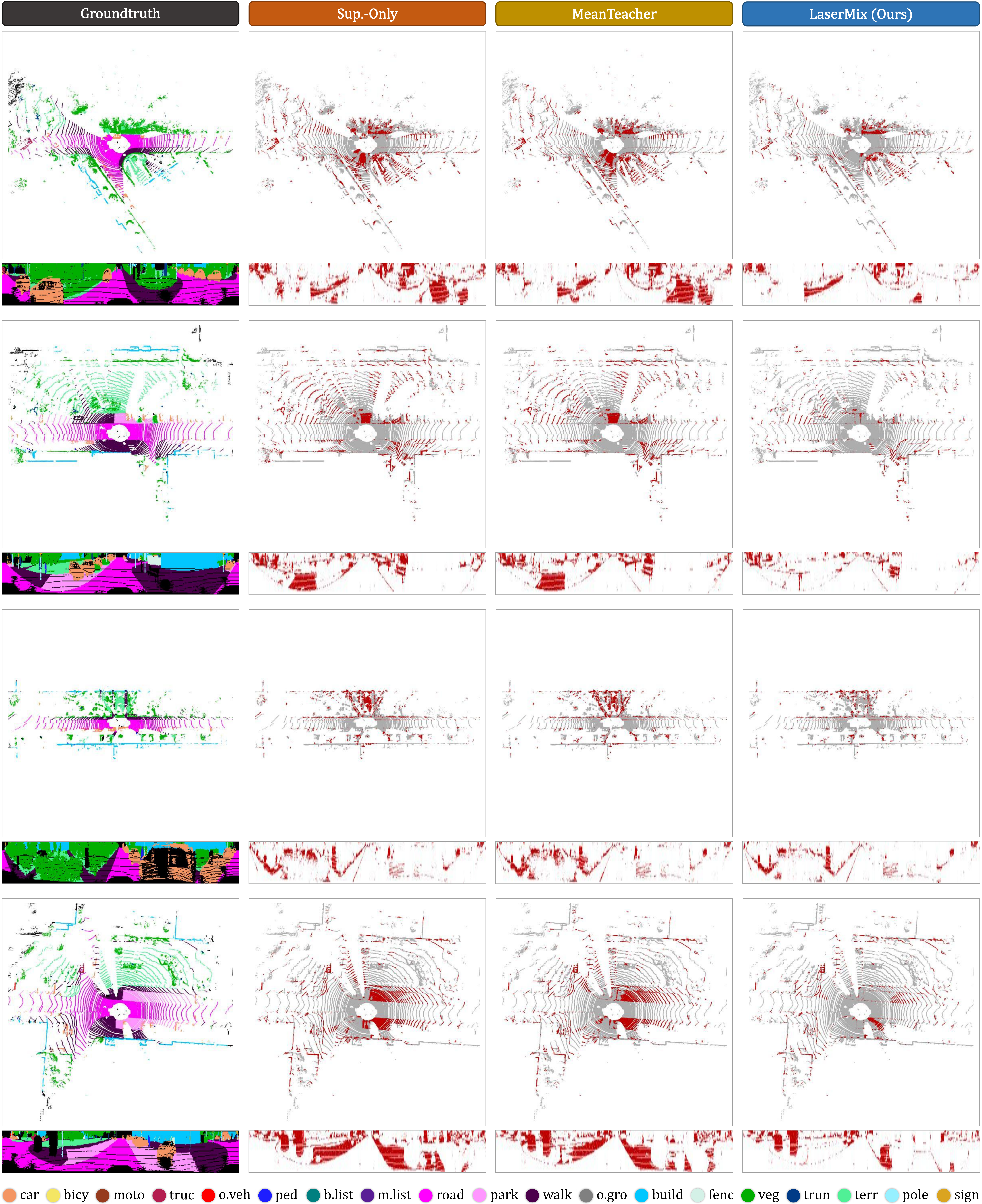}
    \end{center}
    \vspace{-0.45cm}
    \caption{\textbf{Additional qualitative results (error maps)}. Models are trained with $1\%$ labeled data on SemanticKITTI~\cite{SemanticKITTI}. To highlight the differences, the \textbf{\textcolor{correct}{correct}} and \textbf{\textcolor{incorrect}{incorrect}} predictions are painted in \textbf{\textcolor{correct}{gray}} and \textbf{\textcolor{incorrect}{red}}, respectively. Each scene is visualized from the bird's eye view (top) and range view (bottom) and covers a region of size $50$m by $50$m, centered around the ego-vehicle. Best viewed in colors.}
    \label{figure:qualitative_supp_06}
    \vspace{-0.cm}
\end{figure*}

\noindent\textbf{Video Demos}.
We have attached three demos to show more qualitative results of our approach (see our project page\footnote{\url{https://github.com/ldkong1205/LaserMix}.}). Specifically, we show the \textit{error maps}, \textit{i.e.}, the \textit{differences} between the model predictions and the ground-truth, on the \textit{val} sets of SemanticKITTI~\cite{SemanticKITTI}. The models are trained with $1\%$ labeled data, as discussed in our experiment section. We compare the \textit{sup.-only} model and MeanTeacher~\cite{MeanTeacher}. As usual, the error maps are visualized from the LiDAR \textit{bird's eye view} and \textit{range view}. Each sub-figure in the video frame shows a LiDAR point cloud of a street scene of size $50$m ($|p^x_\text{max}|$) by $50$m ($|p^y_\text{max}|$) by $6$m ($|p^z_\text{max}|-|p^z_\text{min}|$). Additionally, we have included several examples from the demos in this file, \textit{i.e.}, \cref{figure:qualitative_supp_01}, \cref{figure:qualitative_supp_02}, \cref{figure:qualitative_supp_03}, \cref{figure:qualitative_supp_04}, \cref{figure:qualitative_supp_05}, and 
\cref{figure:qualitative_supp_06}.

\section{Public Resources Used}
\label{sec:public-resources}

We acknowledge the use of the following public resources, during the course of this work:

\begin{itemize}
    \item nuScenes\footnote{\url{https://www.nuscenes.org/nuscenes}.} \dotfill CC BY-NC-SA 4.0
    \item nuScenes-devkit\footnote{\url{https://github.com/nutonomy/nuscenes-devkit}.} \dotfill Apache License 2.0
    \item SemanticKITTI\footnote{\url{http://semantic-kitti.org}.} \dotfill CC BY-NC-SA 4.0
    \item SemanticKITTI-API\footnote{\url{https://github.com/PRBonn/semantic-kitti-api}.} \dotfill MIT License
    \item ScribbleKITTI\footnote{\url{https://github.com/ouenal/scribblekitti}.} \dotfill Unknown
    \item FIDNet\footnote{\url{https://github.com/placeforyiming/IROS21-FIDNet-SemanticKITTI}.} \dotfill Unknown
    \item Cylinder3D\footnote{\url{https://github.com/xinge008/Cylinder3D}.} \dotfill Apache License 2.0
    \item TorchSemiSeg\footnote{\url{https://github.com/charlesCXK/TorchSemiSeg}.} \dotfill MIT License
    \item Mix3D\footnote{\url{https://github.com/kumuji/mix3d}.} \dotfill Unknown
    \item MixUp\footnote{\url{https://github.com/facebookresearch/mixup-cifar10}.} \dotfill Attribution-NonCommercial 4.0
    \item CutMix\footnote{\url{https://github.com/clovaai/CutMix-PyTorch}.} \dotfill MIT License
    \item CutMix-Seg\footnote{\url{https://github.com/Britefury/cutmix-semisup-seg}.} \dotfill MIT License
    \item CBST\footnote{\url{https://github.com/yzou2/CBST}.} \dotfill Attribution-NonCommercial 4.0
    \item MeanTeacher\footnote{\url{https://github.com/CuriousAI/mean-teacher}.} \dotfill Attribution-NonCommercial 4.0
\end{itemize}

\newpage
\begin{table*}[t]
\caption{\textbf{Class-wise IoU scores} of different SSL algorithms on the \textit{val} set of \textbf{nuScenes} \cite{nuScenes}. All IoU scores are given in percentage ($\%$). The \textcolor{yellow!95}{\textit{\textbf{sup.-only}}} and the \textcolor{blue!50}{\textbf{best}} scores for each semantic class within each split are highlighted in \textcolor{yellow!95}{\textbf{yellow}} and \textcolor{blue!50}{\textbf{blue}}, respectively.}
\vspace{-0.1cm}
\centering\scalebox{0.629}{
\begin{tabular}{c|c|c|c|cccccccccccccccc}
\toprule
Split & Repr. & Method & \rotatebox{0}{mIoU} & \rotatebox{0}{barr} & \rotatebox{0}{bicy} & \rotatebox{0}{bus} & \rotatebox{0}{car} & \rotatebox{0}{const} & \rotatebox{0}{moto} & \rotatebox{0}{ped} & \rotatebox{0}{cone} & \rotatebox{0}{trail} & \rotatebox{0}{truck} & \rotatebox{0}{driv} & \rotatebox{0}{othe} & \rotatebox{0}{walk} & \rotatebox{0}{terr} & \rotatebox{0}{manm} & \rotatebox{0}{veg}
\\\midrule\midrule
\multirow{14}{*}{$1\%$} & \multirow{7}{*}{\rotatebox{90}{Range View}} & \textit{Sup.-only} & \cellcolor{yellow!12.5}$38.3$ & \cellcolor{yellow!12.5}$23.7$ & \cellcolor{yellow!12.5}$0.5$ & \cellcolor{yellow!12.5}$34.2$ & \cellcolor{yellow!12.5}$68.5$ & \cellcolor{yellow!12.5}$0.9$ & \cellcolor{yellow!12.5}$2.6$ & \cellcolor{yellow!12.5}$25.1$ & \cellcolor{yellow!12.5}$26.6$ & \cellcolor{yellow!12.5}$13.1$ & \cellcolor{yellow!12.5}$28.6$ & \cellcolor{yellow!12.5}$89.8$ & \cellcolor{yellow!12.5}$41.3$ & \cellcolor{yellow!12.5}$49.7$ & \cellcolor{yellow!12.5}$61.2$ & \cellcolor{yellow!12.5}$73.4$ & \cellcolor{yellow!12.5}$74.2$
\\\cmidrule{3-20}
& & MeanTeacher~\cite{MeanTeacher} & $42.1$ & $30.5$ & $0.9$ & $35.6$ & $71.9$ & $0.5$ & $4.2$ & $39.6$ & $33.2$ & $15.2$ & $26.3$ & $92.2$ & \cellcolor{blue!9}$51.3$ & $55.2$ & $63.3$ & $77.4$ & $75.9$
\\
& & CBST~\cite{CBST} & $40.9$ & $28.1$ & $1.7$ & $39.3$ & $71.1$ & \cellcolor{blue!9}$2.0$ & $2.9$ & $27.6$ & $32.6$ & $14.6$ & $32.8$ & $90.8$ & $44.1$ & $51.9$ & $63.6$ & $75.3$ & $75.9$
\\
& & CutMix-Seg~\cite{CutMix-Seg} & $43.8$ & $44.6$ & $0.7$ & $30.8$ & $75.7$ & $0.4$ & $2.7$ & $36.9$ & $38.1$ & $18.1$ & $29.8$ & $92.5$ & $45.9$ & $56.9$ & $67.6$ & $80.1$ & $80.0$
\\
& & CPS~\cite{CPS} & $40.7$ & $28.4$ & $0.4$ & $38.3$ & $73.5$ & $0.3$ & $0.5$ & $36.1$ & $25.4$ & $13.6$ & $22.4$ & $91.5$ & $44.1$ & $54.4$ & $66.1$ & $77.8$ & $78.8$
\\\cmidrule{3-20}
& & \textbf{LaserMix} & \cellcolor{blue!9}$49.5$ & \cellcolor{blue!9}$50.7$ & \cellcolor{blue!9}$1.8$ & \cellcolor{blue!9}$39.6$ & \cellcolor{blue!9}$80.7$ & $0.6$ & \cellcolor{blue!9}$17.9$ & \cellcolor{blue!9}$53.4$ & \cellcolor{blue!9}$47.6$ & \cellcolor{blue!9}$23.2$ & \cellcolor{blue!9}$41.9$ & \cellcolor{blue!9}$93.5$ & $45.5$ & \cellcolor{blue!9}$60.9$ & \cellcolor{blue!9}$69.4$ & \cellcolor{blue!9}$82.1$ & \cellcolor{blue!9}$82.4$
\\\cmidrule{2-20}
& \multirow{6}{*}{\rotatebox{90}{Voxel}} & \textit{Sup.-only} & \cellcolor{yellow!12.5}$50.9$ & \cellcolor{yellow!12.5}$41.1$ & \cellcolor{yellow!12.5}$1.9$ & \cellcolor{yellow!12.5}$60.0$ & \cellcolor{yellow!12.5}$77.2$ & \cellcolor{yellow!12.5}$7.4$ & \cellcolor{yellow!12.5}$33.7$ & \cellcolor{yellow!12.5}$47.6$ & \cellcolor{yellow!12.5}$39.6$ & \cellcolor{yellow!12.5}$21.3$ & \cellcolor{yellow!12.5}$51.1$ & \cellcolor{yellow!12.5}$93.4$ & \cellcolor{yellow!12.5}$51.9$ & \cellcolor{yellow!12.5}$60.2$ & \cellcolor{yellow!12.5}$65.8$ & \cellcolor{yellow!12.5}$82.1$ & \cellcolor{yellow!12.5}$80.8$
\\\cmidrule{3-20}
& & MeanTeacher~\cite{MeanTeacher} & $51.6$ & $48.9$ & $0.8$ & \cellcolor{blue!9}$70.4$ & \cellcolor{blue!9}$79.2$ & $1.5$ & \cellcolor{blue!9}$33.8$ & $50.6$ & $13.9$ & $26.9$ & \cellcolor{blue!9}$58.0$ & \cellcolor{blue!9}$93.8$ & $54.5$ & $62.1$ & $66.5$ & $82.9$ & $82.6$
\\
& & CBST~\cite{CBST} & $53.0$ & \cellcolor{blue!9}$59.3$ & \cellcolor{blue!9}$2.6$ & $68.2$ & $77.8$ & $14.2$ & $8.3$ & $54.6$ & $42.2$ & $24.9$ & $51.3$ & $93.1$ & \cellcolor{blue!9}$57.5$ & $60.5$ & $67.0$ & $83.5$ & $82.6$
\\
& & CPS~\cite{CPS} & $52.9$ & $44.1$ & $1.8$ & $63.3$ & $79.1$ & $4.4$ & $28.1$ & $54.2$ & $42.7$ & $22.7$ & $57.6$ & $92.9$ & $56.4$ & \cellcolor{blue!9}$63.5$ & \cellcolor{blue!9}$68.4$ & $83.8$ & \cellcolor{blue!9}$84.0$
\\\cmidrule{3-20}
& & \textbf{LaserMix} & \cellcolor{blue!9}$55.3$ & $53.6$ & $1.9$ & $67.2$ & \cellcolor{blue!9}$79.2$ & \cellcolor{blue!9}$21.1$ & $29.7$ & \cellcolor{blue!9}$57.3$ & \cellcolor{blue!9}$46.8$ & \cellcolor{blue!9}$28.0$ & $55.5$ & $93.6$ & $54.8$ & $62.1$ & $66.5$ & \cellcolor{blue!9}$83.9$ & $83.3$

\\\midrule\midrule

\multirow{14}{*}{$10\%$} & \multirow{7}{*}{\rotatebox{90}{Range View}} & \textit{Sup.-only} & \cellcolor{yellow!12.5}$57.5$ & \cellcolor{yellow!12.5}$65.4$ & \cellcolor{yellow!12.5}$14.7$ & \cellcolor{yellow!12.5}$58.9$ & \cellcolor{yellow!12.5}$82.0$ & \cellcolor{yellow!12.5}$20.7$ & \cellcolor{yellow!12.5}$17.1$ & \cellcolor{yellow!12.5}$60.4$ & \cellcolor{yellow!12.5}$54.0$ & \cellcolor{yellow!12.5}$35.1$ & \cellcolor{yellow!12.5}$54.2$ & \cellcolor{yellow!12.5}$94.6$ & \cellcolor{yellow!12.5}$60.5$ & \cellcolor{yellow!12.5}$65.3$ & \cellcolor{yellow!12.5}$70.0$ & \cellcolor{yellow!12.5}$84.1$ & \cellcolor{yellow!12.5}$82.9$
\\\cmidrule{3-20}
& & MeanTeacher~\cite{MeanTeacher} & $60.4$ & $69.0$ & $12.5$ & $67.0$ & $83.6$ & $27.2$ & $22.0$ & $63.7$ & $55.0$ & $40.4$ & $58.8$ & $95.0$ & $63.8$ & $67.2$ & $71.3$ & $85.6$ & $84.6$
\\
& & CBST~\cite{CBST} & $60.5$ & $68.0$ & $16.8$ & $61.6$ & $83.2$ & $28.4$ & $40.6$ & $62.1$ & $56.4$ & $34.5$ & $54.1$ & $94.9$ & $62.7$ & $66.3$ & $71.2$ & $84.5$ & $83.3$
\\
& & CutMix-Seg~\cite{CutMix-Seg} & $63.9$ & $70.5$ & $22.7$ & $69.2$ & $83.4$ & $26.6$ & \cellcolor{blue!9}$69.5$ & $65.2$ & $54.7$ & $39.4$ & $59.2$ & $95.2$ & $59.2$ & $68.3$ & $72.3$ & $84.3$ & $82.8$
\\
& & CPS~\cite{CPS} & $60.8$ & $66.6$ & $9.8$ & $66.6$ & $85.7$ & $22.4$ & $19.7$ & $63.5$ & $57.2$ & $43.4$ & $62.2$ & $95.4$ & $65.3$ & $69.8$ & $73.8$ & $85.9$ & $85.3$
\\\cmidrule{3-20}
& & \textbf{LaserMix} & \cellcolor{blue!9}$68.2$ & \cellcolor{blue!9}$73.9$ & \cellcolor{blue!9}$27.0$ & \cellcolor{blue!9}$74.8$ & \cellcolor{blue!9}$86.9$ & \cellcolor{blue!9}$34.2$ & $69.1$ & \cellcolor{blue!9}$68.7$ & \cellcolor{blue!9}$60.9$ & \cellcolor{blue!9}$48.0$ & \cellcolor{blue!9}$65.5$ & \cellcolor{blue!9}$95.8$ & \cellcolor{blue!9}$68.1$ & \cellcolor{blue!9}$71.0$ & \cellcolor{blue!9}$74.3$ & \cellcolor{blue!9}$87.3$ & \cellcolor{blue!9}$86.0$
\\\cmidrule{2-20}
& \multirow{6}{*}{\rotatebox{90}{Voxel}} & \textit{Sup.-only} & \cellcolor{yellow!12.5}$65.9$ & \cellcolor{yellow!12.5}$69.5$ & \cellcolor{yellow!12.5}$14.0$ & \cellcolor{yellow!12.5}$87.2$ & \cellcolor{yellow!12.5}$83.5$ & \cellcolor{yellow!12.5}$30.8$ & \cellcolor{yellow!12.5}$61.9$ & \cellcolor{yellow!12.5}$63.5$ & \cellcolor{yellow!12.5}$55.7$ & \cellcolor{yellow!12.5}$47.6$ & \cellcolor{yellow!12.5}$75.2$ & \cellcolor{yellow!12.5}$95.1$ & \cellcolor{yellow!12.5}$63.5$ & \cellcolor{yellow!12.5}$68.1$ & \cellcolor{yellow!12.5}$69.6$ & \cellcolor{yellow!12.5}$85.8$ & \cellcolor{yellow!12.5}$84.1$
\\\cmidrule{3-20}
& & MeanTeacher~\cite{MeanTeacher} & $66.0$ & $71.1$ & $19.7$ & $85.1$ & $83.3$ & \cellcolor{blue!9}$42.0$ & $43.5$ & $64.0$ & $54.9$ & $45.6$ & $73.7$ & $95.3$ & $66.8$ & $69.8$ & $69.6$ & $86.7$ & $84.9$
\\
& & CBST~\cite{CBST} & $66.5$ & $70.1$ & $13.6$ & $85.9$ & $82.2$ & $35.1$ & $59.1$ & $61.9$ & $52.1$ & \cellcolor{blue!9}$57.5$ & $74.0$ & $94.5$ & $65.0$ & $70.1$ & \cellcolor{blue!9}$71.8$ & $86.6$ & $85.0$
\\
& & CPS~\cite{CPS} & $66.3$ & \cellcolor{blue!9}$72.3$ & $16.4$ & $84.5$ & $81.8$ & $38.5$ & $60.3$ & $62.7$ & $53.4$ & $47.1$ & $70.1$ & $94.7$ & $65.4$ & $70.1$ & $71.7$ & \cellcolor{blue!9}$87.1$ & $85.5$
\\\cmidrule{3-20}
& & \textbf{LaserMix} & \cellcolor{blue!9}$69.9$ & $72.1$ & \cellcolor{blue!9}$23.3$ & \cellcolor{blue!9}$87.7$ & \cellcolor{blue!9}$84.6$ & $41.3$ & \cellcolor{blue!9}$72.4$ & \cellcolor{blue!9}$67.9$ & \cellcolor{blue!9}$57.2$ & $56.7$ & \cellcolor{blue!9}$77.2$ & \cellcolor{blue!9}$95.5$ & \cellcolor{blue!9}$67.4$ & \cellcolor{blue!9}$70.8$ & $71.2$ & $87.0$ & \cellcolor{blue!9}$85.6$

\\\midrule\midrule

\multirow{14}{*}{$20\%$} & \multirow{7}{*}{\rotatebox{90}{Range View}} & \textit{Sup.-only} & \cellcolor{yellow!12.5}$62.7$ & \cellcolor{yellow!12.5}$69.4$ & \cellcolor{yellow!12.5}$19.1$ & \cellcolor{yellow!12.5}$69.7$ & \cellcolor{yellow!12.5}$84.9$ & \cellcolor{yellow!12.5}$29.1$ & \cellcolor{yellow!12.5}$36.4$ & \cellcolor{yellow!12.5}$64.0$ & \cellcolor{yellow!12.5}$58.5$ & \cellcolor{yellow!12.5}$44.6$ & \cellcolor{yellow!12.5}$61.0$ & \cellcolor{yellow!12.5}$95.2$ & \cellcolor{yellow!12.5}$63.6$ & \cellcolor{yellow!12.5}$67.2$ & \cellcolor{yellow!12.5}$70.9$ & \cellcolor{yellow!12.5}$85.7$ & \cellcolor{yellow!12.5}$84.6$
\\\cmidrule{3-20}
& & MeanTeacher~\cite{MeanTeacher} & $65.4$ & $70.7$ & $18.9$ & $75.3$ & $85.6$ & $32.4$ & $48.5$ & $72.2$ & $59.0$ & $46.1$ & $64.0$ & $95.2$ & $65.3$ & $68.3$ & $72.8$ & $86.9$ & $85.7$
\\
& & CBST~\cite{CBST} & $64.3$ & $71.6$ & $19.0$ & $70.4$ & $84.4$ & $29.9$ & $49.7$ & $66.2$ & $60.8$ & $46.3$ & $61.3$ & $95.4$ & $62.3$ & $68.4$ & $71.7$ & $86.1$ & $84.6$
\\
& & CutMix-Seg~\cite{CutMix-Seg} & $64.8$ & $72.7$ & $23.2$ & $71.8$ & $86.3$ & $34.3$ & $38.2$ & $69.4$ & $59.1$ & $46.9$ & $63.2$ & $95.5$ & $62.0$ & $69.1$ & $72.7$ & $86.9$ & $85.6$
\\
& & CPS~\cite{CPS} & $64.9$ & $69.6$ & $7.0$ & $75.1$ & $86.6$ & $23.5$ & $50.8$ & $68.5$ & $59.0$ & $50.2$ & $66.5$ & $95.8$ & \cellcolor{blue!9}$68.8$ & $71.1$ & $73.9$ & $85.9$ & $85.6$
\\\cmidrule{3-20}
& & \textbf{LaserMix} & \cellcolor{blue!9}$70.6$ & \cellcolor{blue!9}$74.1$ & \cellcolor{blue!9}$26.1$ & \cellcolor{blue!9}$80.3$ & \cellcolor{blue!9}$89.2$ & \cellcolor{blue!9}$36.2$ & \cellcolor{blue!9}$74.6$ & \cellcolor{blue!9}$73.1$ & \cellcolor{blue!9}$62.8$ & \cellcolor{blue!9}$55.0$ & \cellcolor{blue!9}$73.4$ & \cellcolor{blue!9}$96.0$ & $68.6$ & \cellcolor{blue!9}$71.3$ & \cellcolor{blue!9}$74.3$ & \cellcolor{blue!9}$88.1$ & \cellcolor{blue!9}$86.7$
\\\cmidrule{2-20}
& \multirow{6}{*}{\rotatebox{90}{Voxel}} & \textit{Sup.-only} & \cellcolor{yellow!12.5}$66.6$ & \cellcolor{yellow!12.5}$71.5$ & \cellcolor{yellow!12.5}$27.1$ & \cellcolor{yellow!12.5}$82.1$ & \cellcolor{yellow!12.5}$82.7$ & \cellcolor{yellow!12.5}$37.2$ & \cellcolor{yellow!12.5}$68.6$ & \cellcolor{yellow!12.5}$63.6$ & \cellcolor{yellow!12.5}$53.4$ & \cellcolor{yellow!12.5}$42.2$ & \cellcolor{yellow!12.5}$70.5$ & \cellcolor{yellow!12.5}$94.8$ & \cellcolor{yellow!12.5}$65.9$ & \cellcolor{yellow!12.5}$67.8$ & \cellcolor{yellow!12.5}$69.4$ & \cellcolor{yellow!12.5}$85.1$ & \cellcolor{yellow!12.5}$83.7$
\\\cmidrule{3-20}
& & MeanTeacher~\cite{MeanTeacher} & $67.1$ & $72.1$ & $26.0$ & $89.1$ & \cellcolor{blue!9}$84.4$ & $39.5$ & $18.4$ & \cellcolor{blue!9}$71.3$ & $57.6$ & $59.3$ & $77.5$ & $95.6$ & $66.9$ & $71.3$ & $71.9$ & $87.6$ & $85.8$
\\
& & CBST~\cite{CBST} & $69.6$ & $73.4$ & $29.5$ & $86.1$ & $83.7$ & $37.0$ & $75.7$ & $66.7$ & $56.6$ & $53.0$ & $73.7$ & $95.5$ & $68.5$ & $71.5$ & $70.8$ & $87.3$ & $85.6$
\\
& & CPS~\cite{CPS} & $70.0$ & $73.1$ & $29.3$ & $88.0$ & $83.4$ & $37.2$ & $76.0$ & $66.6$ & $57.8$ & $54.5$ & $75.7$ & $95.5$ & $67.8$ & $71.2$ & $70.5$ & $87.4$ & \cellcolor{blue!9}$85.9$
\\\cmidrule{3-20}
& & \textbf{LaserMix} & \cellcolor{blue!9}$71.8$ & \cellcolor{blue!9}$73.6$ & \cellcolor{blue!9}$32.1$ & \cellcolor{blue!9}$89.6$ & $84.1$ & \cellcolor{blue!9}$41.4$ & \cellcolor{blue!9}$77.0$ & $69.0$ & \cellcolor{blue!9}$60.0$ & \cellcolor{blue!9}$60.9$ & \cellcolor{blue!9}$78.7$ & \cellcolor{blue!9}$95.8$ & \cellcolor{blue!9}$69.6$ & \cellcolor{blue!9}$72.2$ & \cellcolor{blue!9}$72.9$ & \cellcolor{blue!9}$87.9$ & $84.5$

\\\midrule\midrule

\multirow{14}{*}{$50\%$} & \multirow{7}{*}{\rotatebox{90}{Range View}} & \textit{Sup.-only} & \cellcolor{yellow!12.5}$67.6$ & \cellcolor{yellow!12.5}$72.5$ & \cellcolor{yellow!12.5}$32.6$ & \cellcolor{yellow!12.5}$78.5$ & \cellcolor{yellow!12.5}$87.4$ & \cellcolor{yellow!12.5}$32.8$ & \cellcolor{yellow!12.5}$43.6$ & \cellcolor{yellow!12.5}$70.6$ & \cellcolor{yellow!12.5}$62.3$ & \cellcolor{yellow!12.5}$54.0$ & \cellcolor{yellow!12.5}$68.3$ & \cellcolor{yellow!12.5}$95.7$ & \cellcolor{yellow!12.5}$66.4$ & \cellcolor{yellow!12.5}$69.8$ & \cellcolor{yellow!12.5}$72.7$ & \cellcolor{yellow!12.5}$87.7$ & \cellcolor{yellow!12.5}$86.4$
\\\cmidrule{3-20}
& & MeanTeacher~\cite{MeanTeacher} & $69.4$ & $73.4$ & $33.0$ & $81.2$ & $87.6$ & $35.2$ & $61.0$ & $71.9$ & $62.3$ & $55.1$ & $69.4$ & $95.8$ & $66.5$ & $71.1$ & $73.1$ & $87.5$ & $86.1$ 
\\
& & CBST~\cite{CBST} & $69.3$ & $72.7$ & $35.2$ & $80.8$ & $88.0$ & $35.7$ & $53.7$ & $68.2$ & $62.9$ & $60.2$ & $72.0$ & $95.5$ & $67.4$ & $70.3$ & $73.0$ & $87.3$ & $85.8$
\\
& & CutMix-Seg~\cite{CutMix-Seg} & $69.8$ & $74.4$ & $33.5$ & $79.9$ & $88.7$ & $37.3$ & $60.8$ & $70.9$ & $62.0$ & $57.8$ & $70.6$ & $95.8$ & $67.3$ & $70.9$ & $73.3$ & $87.5$ & $85.8$
\\
& & CPS~\cite{CPS} & $68.0$ & $71.2$ & $31.8$ & $71.9$ & $87.1$ & $29.0$ & $57.4$ & $67.4$ & $62.3$ & $58.6$ & $69.0$ & $95.6$ & $68.7$ & $71.1$ & $74.1$ & $86.7$ & $85.4$
\\\cmidrule{3-20}
& & \textbf{LaserMix} & \cellcolor{blue!9}$73.0$ & \cellcolor{blue!9}$76.0$ & \cellcolor{blue!9}$35.6$ & \cellcolor{blue!9}$85.0$ & \cellcolor{blue!9}$89.9$ & \cellcolor{blue!9}$43.3$ & \cellcolor{blue!9}$76.6$ & \cellcolor{blue!9}$72.5$ & \cellcolor{blue!9}$63.9$ & \cellcolor{blue!9}$61.5$ & \cellcolor{blue!9}$75.1$ & \cellcolor{blue!9}$96.1$ & \cellcolor{blue!9}$69.6$ & \cellcolor{blue!9}$72.3$ & \cellcolor{blue!9}$74.8$ & \cellcolor{blue!9}$88.2$ & \cellcolor{blue!9}$86.9$
\\\cmidrule{2-20}
& \multirow{6}{*}{\rotatebox{90}{Voxel}} & \textit{Sup.-only} & \cellcolor{yellow!12.5}$71.2$ & \cellcolor{yellow!12.5}$73.1$ & \cellcolor{yellow!12.5}$35.6$ & \cellcolor{yellow!12.5}$89.0$ & \cellcolor{yellow!12.5}$85.2$ & \cellcolor{yellow!12.5}$41.2$ & \cellcolor{yellow!12.5}$73.3$ & \cellcolor{yellow!12.5}$67.9$ & \cellcolor{yellow!12.5}$59.2$ & \cellcolor{yellow!12.5}$50.9$ & \cellcolor{yellow!12.5}$78.4$ & \cellcolor{yellow!12.5}$95.6$ & \cellcolor{yellow!12.5}$71.5$ & \cellcolor{yellow!12.5}$72.0$ & \cellcolor{yellow!12.5}$73.0$ & \cellcolor{yellow!12.5}$87.3$ & \cellcolor{yellow!12.5}$85.9$
\\\cmidrule{3-20}
& & MeanTeacher~\cite{MeanTeacher} & $71.7$ & $73.7$ & $36.2$ & $90.6$ & \cellcolor{blue!9}$85.0$ & $42.3$ & $76.5$ & $68.3$ & $54.9$ & $61.4$ & $74.3$ & $95.7$ & $69.9$ & $72.2$ & $72.6$ & $87.2$ & $86.0$
\\
& & CBST~\cite{CBST} & $71.6$ & $73.3$ & $36.1$ & $90.2$ & $84.8$ & $42.2$ & $75.7$ & $67.8$ & $56.6$ & \cellcolor{blue!9}$61.5$ & $74.3$ & $95.7$ & $69.1$ & $72.2$ & $72.7$ & $87.1$ & $85.9$
\\
& & CPS~\cite{CPS} & $72.5$ & $73.9$ & $35.6$ & $91.0$ & $84.9$ & $42.9$ & \cellcolor{blue!9}$79.0$ & $68.6$ & \cellcolor{blue!9}$60.3$ & $60.1$ & $78.3$ & \cellcolor{blue!9}$95.8$ & \cellcolor{blue!9}$71.2$ & $72.3$ & $73.2$ & $87.6$ & $85.2$
\\\cmidrule{3-20}
& & \textbf{LaserMix} & \cellcolor{blue!9}$73.2$ & \cellcolor{blue!9}$74.5$ & \cellcolor{blue!9}$36.3$ & \cellcolor{blue!9}$91.1$ & $84.9$ & \cellcolor{blue!9}$48.2$ & $78.5$ & \cellcolor{blue!9}$70.5$ & $59.6$ & $59.8$ & \cellcolor{blue!9}$78.9$ & $95.1$ & $70.7$ & \cellcolor{blue!9}$73.5$ & \cellcolor{blue!9}$74.1$ & \cellcolor{blue!9}$88.6$ & \cellcolor{blue!9}$86.9$
\\\bottomrule
\end{tabular}}
\label{table:class-nuScenes-val}
\end{table*}

\newpage
\begin{table*}[t]
\caption{\textbf{Class-wise IoU scores} of different SSL algorithms on the \textit{val} set of \textbf{SemanticKITTI} \cite{SemanticKITTI}. All IoU scores are given in percentage ($\%$). The \textcolor{yellow!95}{\textit{\textbf{sup.-only}}} and the \textcolor{blue!50}{\textbf{best}} scores for each semantic class within each split are highlighted in \textcolor{yellow!95}{\textbf{yellow}} and \textcolor{blue!50}{\textbf{blue}}, respectively.}
\vspace{-0.1cm}
\centering\scalebox{0.629}{
\begin{tabular}{c|c|c|c|ccccccccccccccccccc}
\toprule
Split & Repr. & Method & \rotatebox{0}{mIoU} & \rotatebox{0}{car} & \rotatebox{0}{bicy} & \rotatebox{0}{moto} & \rotatebox{0}{truck} & \rotatebox{0}{bus} & \rotatebox{0}{ped} & \rotatebox{0}{b.cyc} & \rotatebox{0}{m.cyc} & \rotatebox{0}{road} & \rotatebox{0}{park} & \rotatebox{0}{walk} & \rotatebox{0}{o.gro} & \rotatebox{0}{build} & \rotatebox{0}{fence} & \rotatebox{0}{veg} & \rotatebox{0}{trunk} & \rotatebox{0}{terr} & \rotatebox{0}{pole} & \rotatebox{0}{sign}
\\\midrule\midrule
\multirow{14}{*}{$1\%$} & \multirow{7}{*}{\rotatebox{90}{Range View}} & \textit{Sup.-only} & \cellcolor{yellow!12.5}$36.2$ & \cellcolor{yellow!12.5}$86.8$ & \cellcolor{yellow!12.5}$0.6$ & \cellcolor{yellow!12.5}$0.0$ & \cellcolor{yellow!12.5}$13.0$ & \cellcolor{yellow!12.5}$5.7$ & \cellcolor{yellow!12.5}$12.1$ & \cellcolor{yellow!12.5}$6.6$ & \cellcolor{yellow!12.5}$0.0$ & \cellcolor{yellow!12.5}$87.9$ & \cellcolor{yellow!12.5}$13.4$ & \cellcolor{yellow!12.5}$71.3$ & \cellcolor{yellow!12.5}$0.1$ & \cellcolor{yellow!12.5}$80.4$ & \cellcolor{yellow!12.5}$42.3$ & \cellcolor{yellow!12.5}$78.7$ & \cellcolor{yellow!12.5}$38.1$ & \cellcolor{yellow!12.5}$62.8$ & \cellcolor{yellow!12.5}$52.5$ & \cellcolor{yellow!12.5}$35.7$
\\\cmidrule{3-23}
& & MeanTeacher~\cite{MeanTeacher} & $37.5$ & $88.0$ & $0.1$ & $0.1$ & \cellcolor{blue!9}$12.4$ & $3.6$ & $13.0$ & $12.6$ & $0.0$ & $89.2$ & $19.6$ & $73.0$ & $0.0$ & $81.6$ & $44.8$ & $80.2$ & $41.8$ & $64.4$ & $54.0$ & $33.3$
\\
& & CBST~\cite{CBST} & $39.9$ & \cellcolor{blue!9}$89.4$ & $1.9$ & $0.0$ & $4.6$ & \cellcolor{blue!9}$5.8$ & \cellcolor{blue!9}$27.3$ & $3.4$ & $0.0$ & $91.3$ & $25.9$ & $76.5$ & $0.0$ & \cellcolor{blue!9}$83.9$ & $49.1$ & \cellcolor{blue!9}$82.7$ & \cellcolor{blue!9}$56.4$ & $68.1$ & $57.5$ & $33.6$
\\
& & CutMix-Seg~\cite{CutMix-Seg} & $37.4$ & $86.6$ & $0.2$ & $0.0$ & $3.2$ & $1.5$ & $18.6$ & $6.4$ & $0.0$ & $90.8$ & $24.2$ & $74.9$ & $0.0$ & $81.5$ & $45.5$ & $81.3$ & $50.0$ & $65.7$ & $52.9$ & $34.6$
\\
& & CPS~\cite{CPS} & $36.5$ & $88.9$ & $0.0$ & $0.0$ & $3.1$ & $0.4$ & $5.7$ & $2.7$ & $0.0$ & $90.8$ & $13.7$ & $76.7$ & $0.0$ & $83.4$ & \cellcolor{blue!9}$52.2$ & $79.9$ & $40.8$ & $63.8$ & $55.9$ & $32.3$
\\\cmidrule{3-23}
& & \textbf{LaserMix} & \cellcolor{blue!9}$43.4$ & $88.8$ & \cellcolor{blue!9}$37.1$ & \cellcolor{blue!9}$0.2$ & $2.1$ & $4.1$ & $10.7$ & \cellcolor{blue!9}$40.7$ & \cellcolor{blue!9}$0.2$ & \cellcolor{blue!9}$91.9$ & \cellcolor{blue!9}$32.3$ & \cellcolor{blue!9}$77.0$ & $0.0$ & \cellcolor{blue!9}$83.9$ & $48.8$ & $81.4$ & $55.9$ & \cellcolor{blue!9}$69.4$ & \cellcolor{blue!9}$59.0$ & \cellcolor{blue!9}$41.7$
\\\cmidrule{2-23}
& \multirow{6}{*}{\rotatebox{90}{Voxel}} & \textit{Sup.-only} & \cellcolor{yellow!12.5}$45.4$ & \cellcolor{yellow!12.5}$90.9$ & \cellcolor{yellow!12.5}$24.5$ & \cellcolor{yellow!12.5}$2.8$ & \cellcolor{yellow!12.5}$35.1$ & \cellcolor{yellow!12.5}$20.4$ & \cellcolor{yellow!12.5}$31.7$ & \cellcolor{yellow!12.5}$49.5$ & \cellcolor{yellow!12.5}$0.0$ & \cellcolor{yellow!12.5}$85.5$ & \cellcolor{yellow!12.5}$23.4$ & \cellcolor{yellow!12.5}$67.5$ & \cellcolor{yellow!12.5}$1.3$ & \cellcolor{yellow!12.5}$85.0$ & \cellcolor{yellow!12.5}$46.0$ & \cellcolor{yellow!12.5}$84.1$ & \cellcolor{yellow!12.5}$49.1$ & \cellcolor{yellow!12.5}$70.3$ & \cellcolor{yellow!12.5}$55.0$ & \cellcolor{yellow!12.5}$40.6$
\\\cmidrule{3-23}
& & MeanTeacher~\cite{MeanTeacher} & $45.4$ & $91.2$ & $13.2$ & $5.4$ & $47.3$ & $14.5$ & $29.0$ & $37.3$ & $0.0$ & $86.8$ & $22.6$ & $70.3$ & \cellcolor{blue!9}$1.2$ & $86.7$ & $45.4$ & \cellcolor{blue!9}$84.7$ & $59.4$ & \cellcolor{blue!9}$70.9$ & $55.8$ & $40.8$
\\
& & CBST~\cite{CBST} & $48.8$ & \cellcolor{blue!9}$92.4$ & $16.3$ & $6.4$ & \cellcolor{blue!9}$61.9$ & \cellcolor{blue!9}$27.0$ & $35.7$ & $49.4$ & $0.0$ & $88.9$ & $29.4$ & $73.2$ & $0.7$ & \cellcolor{blue!9}$89.1$ & $49.5$ & $83.9$ & $51.4$ & $68.1$ & \cellcolor{blue!9}$59.8$ & $44.0$
\\
& & CPS~\cite{CPS} & $46.7$ & $92.0$ & $13.5$ & $7.1$ & $37.8$ & $12.7$ & $33.0$ & \cellcolor{blue!9}$54.5$ & $0.0$ & \cellcolor{blue!9}$89.8$ & $25.0$ & \cellcolor{blue!9}$73.8$ & $0.0$ & $88.8$ & \cellcolor{blue!9}$50.1$ & $83.6$ & $57.4$ & $67.8$ & $58.2$ & $42.1$
\\\cmidrule{3-23}
& & \textbf{LaserMix} & \cellcolor{blue!9}$50.6$ & $91.8$ & \cellcolor{blue!9}$35.7$ & \cellcolor{blue!9}$19.8$ & $37.5$ & $25.6$ & \cellcolor{blue!9}$53.6$ & $45.7$ & \cellcolor{blue!9}$2.5$ & $87.8$ & \cellcolor{blue!9}$33.5$ & $71.3$ & $0.7$ & $87.3$ & $43.8$ & $84.6$ & \cellcolor{blue!9}$62.7$ & $69.3$ & \cellcolor{blue!9}$59.8$ & \cellcolor{blue!9}$47.6$

\\\midrule\midrule

\multirow{14}{*}{$10\%$} & \multirow{7}{*}{\rotatebox{90}{Range View}} & \textit{Sup.-only} & \cellcolor{yellow!12.5}$52.2$ & \cellcolor{yellow!12.5}$90.4$ & \cellcolor{yellow!12.5}$34.2$ & \cellcolor{yellow!12.5}$22.6$ & \cellcolor{yellow!12.5}$48.2$ & \cellcolor{yellow!12.5}$24.5$ & \cellcolor{yellow!12.5}$59.7$ & \cellcolor{yellow!12.5}$60.9$ & \cellcolor{yellow!12.5}$0.0$ & \cellcolor{yellow!12.5}$92.2$ & \cellcolor{yellow!12.5}$31.8$ & \cellcolor{yellow!12.5}$78.1$ & \cellcolor{yellow!12.5}$0.5$ & \cellcolor{yellow!12.5}$85.7$ & \cellcolor{yellow!12.5}$47.9$ & \cellcolor{yellow!12.5}$83.9$ & \cellcolor{yellow!12.5}$59.3$ & \cellcolor{yellow!12.5}$69.3$ & \cellcolor{yellow!12.5}$59.0$ & \cellcolor{yellow!12.5}$44.2$
\\\cmidrule{3-23}
& & MeanTeacher~\cite{MeanTeacher} & $53.1$ & $91.1$ & $30.8$ & $23.1$ & $58.9$ & $27.5$ & $60.1$ & $57.9$ & $0.0$ & $92.9$ & $34.7$ & $78.7$ & \cellcolor{blue!9}$0.9$ & $87.3$ & $53.5$ & $83.3$ & $59.6$ & $66.9$ & $57.0$ & $44.1$
\\
& & CBST~\cite{CBST} & $53.4$ & $91.7$ & $33.7$ & $28.9$ & $62.0$ & $29.7$ & $57.9$ & $55.2$ & $0.0$ & $92.9$ & $32.5$ & $78.7$ & $0.8$ & $87.1$ & \cellcolor{blue!9}$53.7$ & $83.5$ & $59.4$ & $68.1$ & $56.7$ & $42.4$
\\
& & CutMix-Seg~\cite{CutMix-Seg} & $54.3$ & $90.9$ & $34.9$ & $37.2$ & $57.4$ & $31.7$ & $56.1$ & $63.9$ & $0.0$ & $92.9$ & $34.5$ & $78.6$ & $0.5$ & $87.0$ & $52.3$ & $83.6$ & $58.8$ & $68.8$ & $55.2$ & $44.7$
\\
& & CPS~\cite{CPS} & $52.3$ & $90.2$ & $32.8$ & $19.7$ & $54.0$ & $23.8$ & $56.8$ & $50.5$ & $0.0$ & $92.7$ & $36.3$ & $79.5$ & $0.4$ & \cellcolor{blue!9}$87.6$ & $52.0$ & \cellcolor{blue!9}$85.7$ & $59.4$ & \cellcolor{blue!9}$69.2$ & $58.6$ & $45.1$
\\\cmidrule{3-23}
& & \textbf{LaserMix} & \cellcolor{blue!9}$58.8$ & \cellcolor{blue!9}$92.0$ & \cellcolor{blue!9}$43.5$ & \cellcolor{blue!9}$50.4$ & \cellcolor{blue!9}$76.1$ & \cellcolor{blue!9}$37.1$ & \cellcolor{blue!9}$69.9$ & \cellcolor{blue!9}$74.3$ & $0.0$ & \cellcolor{blue!9}$93.4$ & \cellcolor{blue!9}$38.8$ & \cellcolor{blue!9}$80.1$ & $0.6$ & $87.1$ & $53.3$ & $84.2$ & \cellcolor{blue!9}$63.2$ & $68.3$ & \cellcolor{blue!9}$58.8$ & \cellcolor{blue!9}$45.3$
\\\cmidrule{2-23}
& \multirow{6}{*}{\rotatebox{90}{Voxel}} & \textit{Sup.-only} & \cellcolor{yellow!12.5}$56.1$ & \cellcolor{yellow!12.5}$93.4$ & \cellcolor{yellow!12.5}$38.4$ & \cellcolor{yellow!12.5}$47.7$ & \cellcolor{yellow!12.5}$65.7$ & \cellcolor{yellow!12.5}$31.0$ & \cellcolor{yellow!12.5}$61.9$ & \cellcolor{yellow!12.5}$64.9$ & \cellcolor{yellow!12.5}$0.0$ & \cellcolor{yellow!12.5}$90.7$ & \cellcolor{yellow!12.5}$37.7$ & \cellcolor{yellow!12.5}$75.3$ & \cellcolor{yellow!12.5}$0.9$ & \cellcolor{yellow!12.5}$89.2$ & \cellcolor{yellow!12.5}$50.5$ & \cellcolor{yellow!12.5}$86.4$ & \cellcolor{yellow!12.5}$56.0$ & \cellcolor{yellow!12.5}$73.9$ & \cellcolor{yellow!12.5}$56.2$ & \cellcolor{yellow!12.5}$46.0$
\\\cmidrule{3-23}
& & MeanTeacher~\cite{MeanTeacher} & $57.1$ & \cellcolor{blue!9}$94.1$ & $40.5$ & \cellcolor{blue!9}$58.4$ & $56.0$ & $38.0$ & \cellcolor{blue!9}$66.5$ & $75.6$ & $0.0$ & $88.4$ & $22.7$ & $72.0$ & $1.5$ & $87.9$ & $49.3$ & $86.7$ & $66.1$ & $74.2$ & $58.0$ & $49.2$
\\
& & CBST~\cite{CBST} & $58.3$ & $93.6$ & $40.3$ & $43.5$ & \cellcolor{blue!9}$80.4$ & $33.8$ & $57.6$ & $78.1$ & $0.0$ & $91.6$ & $36.3$ & $76.6$ & \cellcolor{blue!9}$5.1$ & $89.2$ & $51.1$ & $86.3$ & $61.9$ & $71.2$ & $61.3$ & $49.7$
\\
& & CPS~\cite{CPS} & $58.7$ & $94.0$ & $38.7$ & $51.0$ & $60.3$ & \cellcolor{blue!9}$39.8$ & $65.7$ & $80.0$ & $0.0$ & $91.4$ & $33.2$ & $76.4$ & $2.9$ & \cellcolor{blue!9}$89.8$ & \cellcolor{blue!9}$53.8$ & \cellcolor{blue!9}$87.2$ & $65.7$ & \cellcolor{blue!9}$74.6$ & \cellcolor{blue!9}$61.5$ & $50.0$
\\\cmidrule{3-23}
& & \textbf{LaserMix} & \cellcolor{blue!9}$60.0$ & $93.8$ & \cellcolor{blue!9}$44.9$ & \cellcolor{blue!9}$58.4$ & $65.6$ & $39.4$ & $65.8$ & \cellcolor{blue!9}$80.9$ & \cellcolor{blue!9}$0.2$ & \cellcolor{blue!9}$92.0$ & \cellcolor{blue!9}$44.2$ & \cellcolor{blue!9}$77.1$ & $3.9$ & $89.1$ & $49.0$ & $86.2$ & \cellcolor{blue!9}$66.8$ & $72.3$ & $58.4$ & \cellcolor{blue!9}$51.2$

\\\midrule\midrule

\multirow{14}{*}{$20\%$} & \multirow{7}{*}{\rotatebox{90}{Range View}} & \textit{Sup.-only} & \cellcolor{yellow!12.5}$55.9$ & \cellcolor{yellow!12.5}$92.2$ & \cellcolor{yellow!12.5}$38.4$ & \cellcolor{yellow!12.5}$34.9$ & \cellcolor{yellow!12.5}$68.8$ & \cellcolor{yellow!12.5}$35.1$ & \cellcolor{yellow!12.5}$63.1$ & \cellcolor{yellow!12.5}$69.4$ & \cellcolor{yellow!12.5}$0.0$ & \cellcolor{yellow!12.5}$93.1$ & \cellcolor{yellow!12.5}$33.8$ & \cellcolor{yellow!12.5}$79.0$ & \cellcolor{yellow!12.5}$1.1$ & \cellcolor{yellow!12.5}$86.6$ & \cellcolor{yellow!12.5}$50.4$ & \cellcolor{yellow!12.5}$84.1$ & \cellcolor{yellow!12.5}$60.9$ & \cellcolor{yellow!12.5}$69.2$ & \cellcolor{yellow!12.5}$56.9$ & \cellcolor{yellow!12.5}$45.3$
\\\cmidrule{3-23}
& & MeanTeacher~\cite{MeanTeacher} & $56.1$ & \cellcolor{blue!9}$93.2$ & $33.1$ & $36.3$ & $67.3$ & $39.1$ & $64.9$ & $66.8$ & $0.0$ & $93.3$ & $36.7$ & $79.8$ & $1.0$ & \cellcolor{blue!9}$87.6$ & $54.0$ & $83.9$ & $60.7$ & $67.7$ & $56.5$ & $43.7$
\\
& & CBST~\cite{CBST} & $56.1$ & $92.8$ & $33.2$ & $33.9$ & $64.9$ & $38.9$ & $66.6$ & $69.1$ & $0.0$ & $93.2$ & $36.9$ & $79.7$ & $1.7$ & $87.3$ & $53.6$ & $84.5$ & $60.7$ & $69.1$ & $55.1$ & $44.9$
\\
& & CutMix-Seg~\cite{CutMix-Seg} & $56.6$ & $91.5$ & $42.8$ & $39.8$ & $60.6$ & $32.9$ & $64.3$ & $71.6$ & $0.0$ & $93.1$ & $39.8$ & $79.3$ & $0.6$ & $87.1$ & $53.8$ & \cellcolor{blue!9}$85.0$ & $61.6$ & \cellcolor{blue!9}$71.0$ & $56.1$ & $45.4$
\\
& & CPS~\cite{CPS} & $56.3$ & $90.8$ & \cellcolor{blue!9}$44.0$ & $40.7$ & $67.9$ & $30.7$ & $65.5$ & $58.0$ & $0.0$ & $93.3$ & $39.1$ & $79.5$ & $1.1$ & $87.5$ & \cellcolor{blue!9}$55.6$ & $83.8$ & $60.4$ & $67.9$ & $56.8$ & \cellcolor{blue!9}$46.7$
\\\cmidrule{3-23}
& & \textbf{LaserMix} & \cellcolor{blue!9}$59.4$ & $92.5$ & $43.3$ & \cellcolor{blue!9}$51.5$ & \cellcolor{blue!9}$73.1$ & \cellcolor{blue!9}$45.8$ & \cellcolor{blue!9}$69.4$ & \cellcolor{blue!9}$74.7$ & $0.0$ & \cellcolor{blue!9}$94.0$ & \cellcolor{blue!9}$40.4$ & \cellcolor{blue!9}$80.4$ & \cellcolor{blue!9}$5.0$ & $87.3$ & $53.7$ & $83.8$ & \cellcolor{blue!9}$64.1$ & $66.7$ & \cellcolor{blue!9}$58.0$ & $44.6$
\\\cmidrule{2-23}
& \multirow{6}{*}{\rotatebox{90}{Voxel}} & \textit{Sup.-only} & \cellcolor{yellow!12.5}$57.8$ & \cellcolor{yellow!12.5}$94.0$ & \cellcolor{yellow!12.5}$31.6$ & \cellcolor{yellow!12.5}$47.3$ & \cellcolor{yellow!12.5}$89.5$ & \cellcolor{yellow!12.5}$38.3$ & \cellcolor{yellow!12.5}$57.9$ & \cellcolor{yellow!12.5}$79.1$ & \cellcolor{yellow!12.5}$0.0$ & \cellcolor{yellow!12.5}$91.6$ & \cellcolor{yellow!12.5}$29.6$ & \cellcolor{yellow!12.5}$76.1$ & \cellcolor{yellow!12.5}$0.9$ & \cellcolor{yellow!12.5}$87.8$ & \cellcolor{yellow!12.5}$43.6$ & \cellcolor{yellow!12.5}$86.6$ & \cellcolor{yellow!12.5}$63.7$ & \cellcolor{yellow!12.5}$72.5$ & \cellcolor{yellow!12.5}$61.8$ & \cellcolor{yellow!12.5}$47.5$
\\\cmidrule{3-23}
& & MeanTeacher~\cite{MeanTeacher} & $59.2$ & \cellcolor{blue!9}$94.4$ & $38.7$ & $52.5$ & \cellcolor{blue!9}$81.2$ & $45.8$ & $64.2$ & $78.0$ & $0.0$ & $90.9$ & $35.2$ & $75.7$ & $1.8$ & $89.2$ & $49.8$ & $86.3$ & $65.6$ & $72.6$ & $56.0$ & $47.6$
\\
& & CBST~\cite{CBST} & $59.4$ & $94.2$ & $41.8$ & $51.4$ & $77.7$ & $39.8$ & $65.4$ & $79.8$ & $0.0$ & $91.7$ & $29.8$ & $76.3$ & $3.5$ & $89.2$ & $49.7$ & \cellcolor{blue!9}$87.1$ & $66.1$ & \cellcolor{blue!9}$74.2$ & \cellcolor{blue!9}$60.1$ & $51.3$
\\
& & CPS~\cite{CPS} & $59.6$ & $94.2$ & $41.8$ & $52.9$ & $78.2$ & $39.6$ & $66.1$ & $80.6$ & $0.0$ & $91.9$ & $30.2$ & $76.4$ & \cellcolor{blue!9}$3.7$ & $89.2$ & $50.0$ & $87.0$ & $66.6$ & $73.7$ & $60.0$ & $51.1$
\\\cmidrule{3-23}
& & \textbf{LaserMix} & \cellcolor{blue!9}$61.9$ & \cellcolor{blue!9}$94.4$ & \cellcolor{blue!9}$46.0$ & \cellcolor{blue!9}$68.0$ & $74.3$ & \cellcolor{blue!9}$47.6$ & \cellcolor{blue!9}$68.1$ & \cellcolor{blue!9}$83.7$ & \cellcolor{blue!9}$0.2$ & \cellcolor{blue!9}$92.6$ & \cellcolor{blue!9}$42.7$ & \cellcolor{blue!9}$78.0$ & $1.9$ & \cellcolor{blue!9}$89.7$ & \cellcolor{blue!9}$52.9$ & $86.0$ & \cellcolor{blue!9}$69.3$ & $70.6$ & $59.2$ & \cellcolor{blue!9}$51.7$

\\\midrule\midrule

\multirow{14}{*}{$50\%$} & \multirow{7}{*}{\rotatebox{90}{Range View}} & \textit{Sup.-only} & \cellcolor{yellow!12.5}$57.2$ & \cellcolor{yellow!12.5}$91.3$ & \cellcolor{yellow!12.5}$41.1$ & \cellcolor{yellow!12.5}$47.7$ & \cellcolor{yellow!12.5}$70.2$ & \cellcolor{yellow!12.5}$41.2$ & \cellcolor{yellow!12.5}$66.0$ & \cellcolor{yellow!12.5}$74.4$ & \cellcolor{yellow!12.5}$0.0$ & \cellcolor{yellow!12.5}$93.0$ & \cellcolor{yellow!12.5}$39.2$ & \cellcolor{yellow!12.5}$79.2$ & \cellcolor{yellow!12.5}$2.0$ & \cellcolor{yellow!12.5}$86.0$ & \cellcolor{yellow!12.5}$44.2$ & \cellcolor{yellow!12.5}$83.4$ & \cellcolor{yellow!12.5}$59.3$ & \cellcolor{yellow!12.5}$68.6$ & \cellcolor{yellow!12.5}$55.5$ & \cellcolor{yellow!12.5}$45.1$
\\\cmidrule{3-23}
& & MeanTeacher~\cite{MeanTeacher} & $57.4$ & \cellcolor{blue!9}$93.1$ & $38.6$ & $42.4$ & $61.0$ & $45.0$ & $65.7$ & $73.9$ & $0.0$ & $93.1$ & $38.1$ & $79.4$ & $2.1$ & \cellcolor{blue!9}$87.5$ & $53.8$ & $85.0$ & $60.3$ & $71.5$ & $53.6$ & $47.2$
\\
& & CBST~\cite{CBST} & $56.9$ & $91.5$ & $40.0$ & $42.9$ & $66.1$ & $41.7$ & $64.8$ & $74.2$ & $0.0$ & $93.0$ & $34.9$ & $79.2$ & $1.2$ & $87.0$ & $48.7$ & $83.7$ & $59.6$ & $68.9$ & $55.3$ & $47.1$
\\
& & CutMix-Seg~\cite{CutMix-Seg} & $57.6$ & $92.0$ & $43.3$ & $48.9$ & $44.6$ & $40.7$ & $67.4$ & $78.5$ & $0.0$ & $93.3$ & $39.1$ & $79.7$ & $3.0$ & $87.2$ & \cellcolor{blue!9}$54.2$ & \cellcolor{blue!9}$86.0$ & $61.6$ & \cellcolor{blue!9}$74.8$ & $55.1$ & $44.8$
\\
& & CPS~\cite{CPS} & $57.4$ & $92.1$ & $38.5$ & $44.3$ & $69.6$ & $45.2$ & $66.5$ & $71.0$ & $0.0$ & $93.5$ & $36.6$ & $80.1$ & $1.7$ & $87.0$ & $48.0$ & $83.9$ & $62.3$ & $68.0$ & \cellcolor{blue!9}$58.0$ & $43.7$
\\\cmidrule{3-23}
& & \textbf{LaserMix} & \cellcolor{blue!9}$61.4$ & $92.5$ & \cellcolor{blue!9}$45.6$ & \cellcolor{blue!9}$58.8$ & \cellcolor{blue!9}$73.0$ & \cellcolor{blue!9}$53.2$ & \cellcolor{blue!9}$71.2$ & \cellcolor{blue!9}$82.4$ & $0.0$ & \cellcolor{blue!9}$93.7$ & \cellcolor{blue!9}$43.2$ & \cellcolor{blue!9}$80.7$ & \cellcolor{blue!9}$5.5$ & \cellcolor{blue!9}$87.5$ & $52.6$ & $85.4$ & \cellcolor{blue!9}$64.0$ & $71.9$ & $57.9$ & \cellcolor{blue!9}$47.9$
\\\cmidrule{2-23}
& \multirow{6}{*}{\rotatebox{90}{Voxel}} & \textit{Sup.-only} & \cellcolor{yellow!12.5}$58.7$ & \cellcolor{yellow!12.5}$93.9$ & \cellcolor{yellow!12.5}$40.4$ & \cellcolor{yellow!12.5}$48.0$ & \cellcolor{yellow!12.5}$81.4$ & \cellcolor{yellow!12.5}$33.7$ & \cellcolor{yellow!12.5}$65.7$ & \cellcolor{yellow!12.5}$79.7$ & \cellcolor{yellow!12.5}$0.0$ & \cellcolor{yellow!12.5}$91.9$ & \cellcolor{yellow!12.5}$32.6$ & \cellcolor{yellow!12.5}$76.7$ & \cellcolor{yellow!12.5}$1.3$ & \cellcolor{yellow!12.5}$89.0$ & \cellcolor{yellow!12.5}$51.8$ & \cellcolor{yellow!12.5}$87.2$ & \cellcolor{yellow!12.5}$61.4$ & \cellcolor{yellow!12.5}$72.5$ & \cellcolor{yellow!12.5}$58.7$ & \cellcolor{yellow!12.5}$48.7$
\\\cmidrule{3-23}
& & MeanTeacher~\cite{MeanTeacher} & $60.0$ & $94.1$ & $41.3$ & $57.7$ & $64.6$ & $39.5$ & $65.3$ & $86.8$ & $0.0$ & $91.3$ & $32.8$ & $75.2$ & \cellcolor{blue!9}$3.5$ & $89.7$ & $48.6$ & $85.4$ & $65.9$ & $70.6$ & $58.7$ & $49.1$
\\
& & CBST~\cite{CBST} & $59.7$ & \cellcolor{blue!9}$94.9$ & $40.9$ & $54.4$ & $75.3$ & $43.8$ & $67.3$ & $86.8$ & $0.0$ & $91.5$ & $33.3$ & $75.7$ & $2.6$ & $89.3$ & $50.7$ & $86.7$ & $63.9$ & $72.4$ & $56.4$ & $48.8$
\\
& & CPS~\cite{CPS} & $60.5$ & $94.6$ & $43.3$ & $55.3$ & \cellcolor{blue!9}$80.5$ & $42.5$ & $67.9$ & $84.6$ & $0.0$ & $92.0$ & $34.3$ & $76.9$ & $2.2$ & $89.8$ & $52.3$ & $86.0$ & $67.4$ & $71.1$ & $59.5$ & $49.4$
\\\cmidrule{3-23}
& & \textbf{LaserMix} & \cellcolor{blue!9}$62.3$ & $94.7$ & \cellcolor{blue!9}$48.4$ & \cellcolor{blue!9}$64.7$ & $65.2$ & \cellcolor{blue!9}$44.5$ & \cellcolor{blue!9}$71.0$ & \cellcolor{blue!9}$88.3$ & \cellcolor{blue!9}$2.1$ & \cellcolor{blue!9}$92.7$ & \cellcolor{blue!9}$43.0$ & \cellcolor{blue!9}$78.4$ & $2.0$ & \cellcolor{blue!9}$90.3$ & \cellcolor{blue!9}$54.9$ & \cellcolor{blue!9}$88.1$ & \cellcolor{blue!9}$68.1$ & \cellcolor{blue!9}$75.3$ & \cellcolor{blue!9}$66.6$ & \cellcolor{blue!9}$51.7$
\\\bottomrule
\end{tabular}}
\label{table:class-SemKitti-val}
\end{table*}

\newpage
\begin{table*}[t]
\caption{\textbf{Class-wise IoU scores} of different SSL algorithms on the \textit{val} set of \textbf{ScribbleKITTI} \cite{ScribbleKITTI} (the same as SemanticKITTI \cite{SemanticKITTI}). All IoU scores are given in percentage ($\%$). The \textcolor{yellow!95}{\textit{\textbf{sup.-only}}} and the \textcolor{blue!50}{\textbf{best}} scores for each semantic class within each split are highlighted in \textcolor{yellow!95}{\textbf{yellow}} and \textcolor{blue!50}{\textbf{blue}}, respectively.}
\vspace{-0.1cm}
\centering\scalebox{0.629}{
\begin{tabular}{c|c|c|c|ccccccccccccccccccc}
\toprule
Split & Repr. & Method & \rotatebox{0}{mIoU} & \rotatebox{0}{car} & \rotatebox{0}{bicy} & \rotatebox{0}{moto} & \rotatebox{0}{truck} & \rotatebox{0}{bus} & \rotatebox{0}{ped} & \rotatebox{0}{b.cyc} & \rotatebox{0}{m.cyc} & \rotatebox{0}{road} & \rotatebox{0}{park} & \rotatebox{0}{walk} & \rotatebox{0}{o.gro} & \rotatebox{0}{build} & \rotatebox{0}{fence} & \rotatebox{0}{veg} & \rotatebox{0}{trunk} & \rotatebox{0}{terr} & \rotatebox{0}{pole} & \rotatebox{0}{sign}
\\\midrule\midrule
\multirow{14}{*}{$1\%$} & \multirow{7}{*}{\rotatebox{90}{Range View}} & \textit{Sup.-only} & \cellcolor{yellow!12.5}$33.1$ & \cellcolor{yellow!12.5}$81.3$ & \cellcolor{yellow!12.5}$2.6$ & \cellcolor{yellow!12.5}$0.4$ & \cellcolor{yellow!12.5}$11.7$ & \cellcolor{yellow!12.5}$8.3$ & \cellcolor{yellow!12.5}$11.5$ & \cellcolor{yellow!12.5}$8.7$ & \cellcolor{yellow!12.5}$0.0$ & \cellcolor{yellow!12.5}$76.7$ & \cellcolor{yellow!12.5}$10.9$ & \cellcolor{yellow!12.5}$61.8$ & \cellcolor{yellow!12.5}$0.1$ & \cellcolor{yellow!12.5}$75.8$ & \cellcolor{yellow!12.5}$26.3$ & \cellcolor{yellow!12.5}$73.8$ & \cellcolor{yellow!12.5}$40.7$ & \cellcolor{yellow!12.5}$56.1$ & \cellcolor{yellow!12.5}$48.9$ & \cellcolor{yellow!12.5}$32.5$
\\\cmidrule{3-23}
& & MeanTeacher~\cite{MeanTeacher} & $34.2$ & $82.3$ & $1.7$ & $0.1$ & $10.4$ & $6.7$ & $6.1$ & $4.7$ & $0.0$ & $78.6$ & $13.4$ & $67.8$ & $0.1$ & $80.7$ & $31.3$ & $76.1$ & $43.1$ & $60.0$ & $53.3$ & $32.8$
\\
& & CBST~\cite{CBST} & $35.7$ & $84.8$ & $1.6$ & $0.4$ & $11.7$ & \cellcolor{blue!9}$10.6$ & $14.9$ & \cellcolor{blue!9}$8.0$ & $0.0$ & $83.6$ & $13.4$ & $68.1$ & $0.1$ & $79.5$ & $32.4$ & $77.1$ & $44.6$ & $60.5$ & $53.0$ & $34.4$
\\
& & CutMix-Seg~\cite{CutMix-Seg} & $36.7$ & $84.7$ & $0.9$ & $0.0$ & $5.5$ & $0.9$ & $18.7$ & $1.9$ & $0.0$ & $89.3$ & \cellcolor{blue!9}$25.1$ & $74.6$ & $0.1$ & $82.6$ & $27.0$ & $77.7$ & $52.1$ & \cellcolor{blue!9}$65.0$ & $54.7$ & $35.8$
\\
& & CPS~\cite{CPS} & $33.7$ & $82.7$ & $0.1$ & $0.0$ & $0.9$ & $0.1$ & $2.9$ & $4.1$ & $0.0$ & $85.9$ & $8.9$ & $70.8$ & $0.0$ & $81.2$ & \cellcolor{blue!9}$47.3$ & \cellcolor{blue!9}$78.1$ & $36.0$ & $61.2$ & $51.9$ & $27.5$
\\\cmidrule{3-23}
& & \textbf{LaserMix} & \cellcolor{blue!9}$38.3$ & \cellcolor{blue!9}$86.5$ & \cellcolor{blue!9}$1.9$ & \cellcolor{blue!9}$0.9$ & \cellcolor{blue!9}$12.8$ & $2.9$ & \cellcolor{blue!9}$25.9$ & $2.6$ & $0.0$ & \cellcolor{blue!9}$90.8$ & $25.0$ & \cellcolor{blue!9}$75.8$ & \cellcolor{blue!9}$1.0$ & \cellcolor{blue!9}$83.9$ & $26.4$ & $77.8$ & \cellcolor{blue!9}$55.5$ & $63.9$ & \cellcolor{blue!9}$56.7$ & \cellcolor{blue!9}$38.2$
\\\cmidrule{2-23}
& \multirow{6}{*}{\rotatebox{90}{Voxel}} & \textit{Sup.-only} & \cellcolor{yellow!12.5}$39.2$ & \cellcolor{yellow!12.5}$83.2$ & \cellcolor{yellow!12.5}$13.8$ & \cellcolor{yellow!12.5}$3.4$ & \cellcolor{yellow!12.5}$26.3$ & \cellcolor{yellow!12.5}$11.8$ & \cellcolor{yellow!12.5}$28.0$ & \cellcolor{yellow!12.5}$25.2$ & \cellcolor{yellow!12.5}$0.0$ & \cellcolor{yellow!12.5}$72.5$ & \cellcolor{yellow!12.5}$13.0$ & \cellcolor{yellow!12.5}$59.5$ & \cellcolor{yellow!12.5}$0.2$ & \cellcolor{yellow!12.5}$86.6$ & \cellcolor{yellow!12.5}$33.7$ & \cellcolor{yellow!12.5}$78.7$ & \cellcolor{yellow!12.5}$55.7$ & \cellcolor{yellow!12.5}$58.4$ & \cellcolor{yellow!12.5}$54.0$ & \cellcolor{yellow!12.5}$40.3$
\\\cmidrule{3-23}
& & MeanTeacher~\cite{MeanTeacher} & $41.0$ & $82.3$ & $15.8$ & $7.1$ & \cellcolor{blue!9}$32.0$ & $15.4$ & $23.7$ & $36.3$ & $0.0$ & \cellcolor{blue!9}$75.0$ & $12.6$ & $61.4$ & $0.9$ & $85.3$ & $30.0$ & $80.1$ & $57.0$ & $67.0$ & $56.1$ & $41.3$
\\
& & CBST~\cite{CBST} & $41.5$ & \cellcolor{blue!9}$83.7$ & $22.1$ & $5.9$ & $28.3$ & $13.4$ & $27.1$ & $34.7$ & $0.0$ & $74.0$ & $14.4$ & $61.7$ & $0.2$ & \cellcolor{blue!9}$88.1$ & \cellcolor{blue!9}$36.6$ & $80.3$ & $58.7$ & $60.4$ & \cellcolor{blue!9}$57.1$ & $41.4$
\\
& & CPS~\cite{CPS} & $41.4$ & $82.8$ & $18.2$ & $11.4$ & $20.9$ & $15.1$ & $22.5$ & $35.5$ & $0.0$ & $74.7$ & \cellcolor{blue!9}$15.7$ & $61.6$ & $0.4$ & $86.0$ & $34.2$ & \cellcolor{blue!9}$82.2$ & $58.4$ & \cellcolor{blue!9}$69.9$ & $56.7$ & $40.0$
\\\cmidrule{3-23}
& & \textbf{LaserMix} & \cellcolor{blue!9}$44.2$ & $82.6$ & \cellcolor{blue!9}$25.5$ & \cellcolor{blue!9}$18.8$ & $29.0$ & \cellcolor{blue!9}$19.8$ & \cellcolor{blue!9}$41.1$ & \cellcolor{blue!9}$47.2$ & \cellcolor{blue!9}$0.6$ & $71.5$ & $10.5$ & \cellcolor{blue!9}$64.2$ & \cellcolor{blue!9}$2.2$ & $85.1$ & $33.5$ & $82.0$ & \cellcolor{blue!9}$59.9$ & $65.8$ & $54.5$ & \cellcolor{blue!9}$45.2$

\\\midrule\midrule

\multirow{14}{*}{$10\%$} & \multirow{7}{*}{\rotatebox{90}{Range View}} & \textit{Sup.-only} & \cellcolor{yellow!12.5}$47.7$ & \cellcolor{yellow!12.5}$85.1$ & \cellcolor{yellow!12.5}$30.2$ & \cellcolor{yellow!12.5}$20.4$ & \cellcolor{yellow!12.5}$40.4$ & \cellcolor{yellow!12.5}$20.9$ & \cellcolor{yellow!12.5}$54.4$ & \cellcolor{yellow!12.5}$55.9$ & \cellcolor{yellow!12.5}$0.0$ & \cellcolor{yellow!12.5}$82.8$ & \cellcolor{yellow!12.5}$21.6$ & \cellcolor{yellow!12.5}$68.4$ & \cellcolor{yellow!12.5}$0.5$ & \cellcolor{yellow!12.5}$84.2$ & \cellcolor{yellow!12.5}$40.5$ & \cellcolor{yellow!12.5}$80.3$ & \cellcolor{yellow!12.5}$58.6$ & \cellcolor{yellow!12.5}$61.9$ & \cellcolor{yellow!12.5}$56.0$ & \cellcolor{yellow!12.5}$45.2$
\\\cmidrule{3-23}
& & MeanTeacher~\cite{MeanTeacher} & $49.8$ & $83.5$ & $30.0$ & $22.7$ & $62.2$ & $31.1$ & $59.1$ & $52.4$ & $0.0$ & $77.9$ & $17.6$ & $70.5$ & $2.0$ & \cellcolor{blue!9}$86.8$ & $42.6$ & $82.0$ & \cellcolor{blue!9}$61.2$ & $62.6$ & $58.4$ & $46.5$
\\
& & CBST~\cite{CBST} & $50.7$ & \cellcolor{blue!9}$90.3$ & $27.2$ & $18.1$ & $53.1$ & $24.6$ & $60.3$ & $56.5$ & $0.0$ & \cellcolor{blue!9}$90.3$ & \cellcolor{blue!9}$32.4$ & \cellcolor{blue!9}$76.0$ & $0.7$ & $86.1$ & $46.0$ & $81.1$ & $58.7$ & $64.5$ & $51.7$ & $45.0$
\\
& & CutMix-Seg~\cite{CutMix-Seg} & $50.7$ & $87.4$ & $28.1$ & $25.9$ & $60.5$ & $24.5$ & $58.4$ & $57.7$ & $0.0$ & $85.5$ & $27.5$ & $72.1$ & \cellcolor{blue!9}$1.3$ & $84.7$ & $39.4$ & $82.4$ & $58.8$ & $68.3$ & $56.4$ & $44.4$
\\
& & CPS~\cite{CPS} & $50.0$ & $85.8$ & $26.7$ & $17.4$ & $54.5$ & $20.5$ & $54.4$ & $53.7$ & $0.0$ & $88.9$ & $29.2$ & $74.4$ & $0.6$ & $86.5$ & \cellcolor{blue!9}$48.6$ & $82.4$ & $58.6$ & $65.2$ & $57.3$ & $45.1$
\\\cmidrule{3-23}
& & \textbf{LaserMix} & \cellcolor{blue!9}$54.4$ & $87.1$ & \cellcolor{blue!9}$35.4$ & \cellcolor{blue!9}$44.4$ & \cellcolor{blue!9}$62.5$ & \cellcolor{blue!9}$36.4$ & \cellcolor{blue!9}$66.9$ & \cellcolor{blue!9}$72.6$ & $0.0$ & $80.8$ & $27.8$ & $73.7$ & $0.6$ & $85.2$ & $35.2$ & \cellcolor{blue!9}$83.9$ & $60.6$ & \cellcolor{blue!9}$70.0$ & \cellcolor{blue!9}$59.3$ & \cellcolor{blue!9}$51.6$
\\\cmidrule{2-23}
& \multirow{6}{*}{\rotatebox{90}{Voxel}} & \textit{Sup.-only} & \cellcolor{yellow!12.5}$48.0$ & \cellcolor{yellow!12.5}$85.7$ & \cellcolor{yellow!12.5}$25.6$ & \cellcolor{yellow!12.5}$21.3$ & \cellcolor{yellow!12.5}$52.8$ & \cellcolor{yellow!12.5}$29.9$ & \cellcolor{yellow!12.5}$46.5$ & \cellcolor{yellow!12.5}$47.2$ & \cellcolor{yellow!12.5}$0.1$ & \cellcolor{yellow!12.5}$79.5$ & \cellcolor{yellow!12.5}$15.4$ & \cellcolor{yellow!12.5}$63.8$ & \cellcolor{yellow!12.5}$0.3$ & \cellcolor{yellow!12.5}$85.4$ & \cellcolor{yellow!12.5}$39.6$ & \cellcolor{yellow!12.5}$84.8$ & \cellcolor{yellow!12.5}$59.7$ & \cellcolor{yellow!12.5}$71.5$ & \cellcolor{yellow!12.5}$57.7$ & \cellcolor{yellow!12.5}$45.8$
\\\cmidrule{3-23}
& & MeanTeacher~\cite{MeanTeacher} & $50.1$ & $83.7$ & $32.6$ & $45.1$ & $41.0$ & $34.7$ & $56.0$ & $59.2$ & $0.0$ & $75.9$ & $14.0$ & $64.0$ & $0.7$ & $85.6$ & $37.9$ & $83.3$ & $62.6$ & $68.2$ & $59.7$ & $47.0$
\\
& & CBST~\cite{CBST} & $50.6$ & \cellcolor{blue!9}$85.8$ & $31.4$ & $30.5$ & \cellcolor{blue!9}$58.5$ & $24.4$ & $55.1$ & $58.8$ & $0.0$ & \cellcolor{blue!9}$82.6$ & $15.3$ & $67.8$ & $0.5$ & $87.7$ & \cellcolor{blue!9}$40.0$ & $82.8$ & $62.5$ & $65.0$ & \cellcolor{blue!9}$62.0$ & \cellcolor{blue!9}$50.8$
\\
& & CPS~\cite{CPS} & $51.8$ & $84.6$ & \cellcolor{blue!9}$34.9$ & \cellcolor{blue!9}$47.1$ & $37.5$ & $29.5$ & $60.1$ & $69.1$ & $0.0$ & $79.8$ & $16.5$ & $67.3$ & \cellcolor{blue!9}$2.7$ & \cellcolor{blue!9}$88.0$ & $39.2$ & \cellcolor{blue!9}$84.5$ & $64.5$ & \cellcolor{blue!9}$71.0$ & $60.4$ & $47.9$
\\\cmidrule{3-23}
& & \textbf{LaserMix} & \cellcolor{blue!9}$53.7$ & \cellcolor{blue!9}$85.8$ & $34.7$ & $45.6$ & $54.9$ & \cellcolor{blue!9}$35.8$ & \cellcolor{blue!9}$63.2$ & \cellcolor{blue!9}$73.6$ & \cellcolor{blue!9}$1.3$ & $79.8$ & \cellcolor{blue!9}$25.0$ & \cellcolor{blue!9}$68.2$ & $1.8$ & $87.7$ & $35.4$ & $84.0$ & \cellcolor{blue!9}$65.8$ & $70.8$ & $59.4$ & $48.2$

\\\midrule\midrule

\multirow{14}{*}{$20\%$} & \multirow{7}{*}{\rotatebox{90}{Range View}} & \textit{Sup.-only} & \cellcolor{yellow!12.5}$49.9$ & \cellcolor{yellow!12.5}$86.3$ & \cellcolor{yellow!12.5}$32.2$ & \cellcolor{yellow!12.5}$23.8$ & \cellcolor{yellow!12.5}$49.5$ & \cellcolor{yellow!12.5}$30.3$ & \cellcolor{yellow!12.5}$60.5$ & \cellcolor{yellow!12.5}$58.4$ & \cellcolor{yellow!12.5}$0.0$ & \cellcolor{yellow!12.5}$83.6$ & \cellcolor{yellow!12.5}$22.4$ & \cellcolor{yellow!12.5}$69.5$ & \cellcolor{yellow!12.5}$1.1$ & \cellcolor{yellow!12.5}$85.1$ & \cellcolor{yellow!12.5}$40.6$ & \cellcolor{yellow!12.5}$80.9$ & \cellcolor{yellow!12.5}$59.9$ & \cellcolor{yellow!12.5}$62.3$ & \cellcolor{yellow!12.5}$55.9$ & \cellcolor{yellow!12.5}$46.4$
\\\cmidrule{3-23}
& & MeanTeacher~\cite{MeanTeacher} & $51.6$ & $82.9$ & $27.7$ & \cellcolor{blue!9}$43.1$ & $59.5$ & $32.8$ & $59.5$ & $60.7$ & $0.0$ & $80.8$ & $25.7$ & $70.3$ & $0.7$ & $85.3$ & $41.6$ & $82.1$ & $60.5$ & $66.0$ & $55.6$ & $45.7$
\\
& & CBST~\cite{CBST} & $52.7$ & \cellcolor{blue!9}$90.0$ & $33.1$ & $30.2$ & $53.6$ & $33.8$ & $60.0$ & $60.4$ & $0.0$ & \cellcolor{blue!9}$89.3$ & \cellcolor{blue!9}$30.3$ & \cellcolor{blue!9}$75.8$ & $0.6$ & $85.6$ & $44.8$ & $83.5$ & $58.6$ & \cellcolor{blue!9}$70.3$ & $54.7$ & $47.1$
\\
& & CutMix-Seg~\cite{CutMix-Seg} & $52.9$ & $86.9$ & $30.0$ & $35.6$ & $64.8$ & $35.7$ & $60.9$ & $63.6$ & $0.0$ & $88.3$ & $29.0$ & $74.7$ & $0.9$ & $85.2$ & $40.3$ & $82.0$ & $59.4$ & $65.2$ & $56.5$ & $45.5$
\\
& & CPS~\cite{CPS} & $52.8$ & $86.3$ & $35.4$ & $28.1$ & $67.1$ & $27.7$ & $59.5$ & $59.2$ & $0.0$ & $89.0$ & $28.0$ & $75.0$ & $0.8$ & \cellcolor{blue!9}$86.7$ & \cellcolor{blue!9}$47.3$ & $83.1$ & $61.0$ & $66.9$ & $58.1$ & $44.6$
\\\cmidrule{3-23}
& & \textbf{LaserMix} & \cellcolor{blue!9}$55.6$ & $87.3$ & \cellcolor{blue!9}$36.0$ & $34.3$ & \cellcolor{blue!9}$69.5$ & \cellcolor{blue!9}$40.6$ & \cellcolor{blue!9}$66.3$ & \cellcolor{blue!9}$70.6$ & $0.0$ & $84.2$ & $27.2$ & $72.3$ & \cellcolor{blue!9}$2.4$ & $86.4$ & $44.6$ & \cellcolor{blue!9}$84.1$ & \cellcolor{blue!9}$62.8$ & $69.8$ & \cellcolor{blue!9}$59.4$ & \cellcolor{blue!9}$57.7$
\\\cmidrule{2-23}
& \multirow{6}{*}{\rotatebox{90}{Voxel}} & \textit{Sup.-only} & \cellcolor{yellow!12.5}$52.1$ & \cellcolor{yellow!12.5}$86.9$ & \cellcolor{yellow!12.5}$38.0$ & \cellcolor{yellow!12.5}$39.5$ & \cellcolor{yellow!12.5}$67.3$ & \cellcolor{yellow!12.5}$29.7$ & \cellcolor{yellow!12.5}$56.5$ & \cellcolor{yellow!12.5}$69.9$ & \cellcolor{yellow!12.5}$0.0$ & \cellcolor{yellow!12.5}$79.0$ & \cellcolor{yellow!12.5}$16.0$ & \cellcolor{yellow!12.5}$66.0$ & \cellcolor{yellow!12.5}$0.3$ & \cellcolor{yellow!12.5}$87.0$ & \cellcolor{yellow!12.5}$38.6$ & \cellcolor{yellow!12.5}$84.3$ & \cellcolor{yellow!12.5}$60.6$ & \cellcolor{yellow!12.5}$66.2$ & \cellcolor{yellow!12.5}$58.8$ & \cellcolor{yellow!12.5}$45.2$
\\\cmidrule{3-23}
& & MeanTeacher~\cite{MeanTeacher} & $52.8$ & $85.9$ & $27.9$ & $41.5$ & $55.5$ & $33.0$ & \cellcolor{blue!9}$64.1$ & $72.0$ & \cellcolor{blue!9}$1.2$ & $81.0$ & $22.5$ & $67.8$ & $1.2$ & $89.1$ & $39.9$ & $82.9$ & $63.7$ & $66.9$ & \cellcolor{blue!9}$60.5$ & $46.7$
\\
& & CBST~\cite{CBST} & $53.3$ & $86.6$ & $36.8$ & $40.9$ & \cellcolor{blue!9}$72.9$ & $28.3$ & $58.0$ & $69.5$ & $0.0$ & $81.1$ & $18.3$ & $68.2$ & $0.7$ & $88.7$ & \cellcolor{blue!9}$44.3$ & $83.6$ & $63.3$ & $64.4$ & $60.3$ & $47.5$
\\
& & CPS~\cite{CPS} & $53.9$ & $85.4$ & $37.2$ & $44.7$ & $58.9$ & $32.9$ & $63.5$ & $71.0$ & $0.0$ & \cellcolor{blue!9}$81.6$ & \cellcolor{blue!9}$23.1$ & $69.2$ & \cellcolor{blue!9}$1.9$ & $88.4$ & $38.2$ & $83.8$ & $65.7$ & $69.2$ & $60.2$ & $48.9$
\\\cmidrule{3-23}
& & \textbf{LaserMix} & \cellcolor{blue!9}$55.1$ & \cellcolor{blue!9}$88.0$ & \cellcolor{blue!9}$38.8$ & \cellcolor{blue!9}$51.3$ & $54.8$ & \cellcolor{blue!9}$36.6$ & $60.2$ & \cellcolor{blue!9}$73.9$ & $0.0$ & $78.8$ & $22.7$ & \cellcolor{blue!9}$71.9$ & $1.5$ & \cellcolor{blue!9}$90.3$ & $43.3$ & \cellcolor{blue!9}$85.3$ & \cellcolor{blue!9}$66.5$ & \cellcolor{blue!9}$70.9$ & $60.3$ & \cellcolor{blue!9}$51.6$

\\\midrule\midrule

\multirow{14}{*}{$50\%$} & \multirow{7}{*}{\rotatebox{90}{Range View}} & \textit{Sup.-only} & \cellcolor{yellow!12.5}$52.5$ & \cellcolor{yellow!12.5}$86.7$ & \cellcolor{yellow!12.5}$35.9$ & \cellcolor{yellow!12.5}$40.2$ & \cellcolor{yellow!12.5}$55.9$ & \cellcolor{yellow!12.5}$30.1$ & \cellcolor{yellow!12.5}$63.2$ & \cellcolor{yellow!12.5}$62.9$ & \cellcolor{yellow!12.5}$0.1$ & \cellcolor{yellow!12.5}$83.6$ & \cellcolor{yellow!12.5}$25.3$ & \cellcolor{yellow!12.5}$70.8$ & \cellcolor{yellow!12.5}$1.1$ & \cellcolor{yellow!12.5}$85.1$ & \cellcolor{yellow!12.5}$40.0$ & \cellcolor{yellow!12.5}$82.9$ & \cellcolor{yellow!12.5}$60.4$ & \cellcolor{yellow!12.5}$69.0$ & \cellcolor{yellow!12.5}$56.3$ & \cellcolor{yellow!12.5}$48.3$
\\\cmidrule{3-23}
& & MeanTeacher~\cite{MeanTeacher} & $53.3$ & $86.9$ & $31.9$ & $37.5$ & $58.6$ & $36.3$ & $63.3$ & $62.0$ & $0.0$ & $87.6$ & $29.5$ & $74.1$ & $1.0$ & $86.4$ & $40.7$ & $82.6$ & $61.3$ & $68.9$ & $58.0$ & $47.0$
\\
& & CBST~\cite{CBST} & $54.6$ & \cellcolor{blue!9}$90.1$ & $36.0$ & $36.6$ & $64.7$ & $41.6$ & $61.2$ & $66.7$ & $0.0$ & \cellcolor{blue!9}$90.4$ & \cellcolor{blue!9}$33.8$ & \cellcolor{blue!9}$76.8$ & \cellcolor{blue!9}$3.8$ & $84.5$ & \cellcolor{blue!9}$44.3$ & $83.7$ & $57.6$ & $70.2$ & $48.4$ & $47.8$
\\
& & CutMix-Seg~\cite{CutMix-Seg} & $54.3$ & $88.1$ & $35.3$ & $40.0$ & $68.8$ & $39.3$ & $62.4$ & $69.8$ & $0.0$ & $88.0$ & $32.0$ & $74.3$ & $0.9$ & $85.1$ & $38.4$ & $82.4$ & $59.3$ & $67.4$ & $56.3$ & $44.5$
\\
& & CPS~\cite{CPS} & $54.6$ & $87.1$ & $35.0$ & $41.0$ & $66.1$ & $40.8$ & $63.2$ & $65.5$ & $0.0$ & $87.9$ & $30.0$ & $74.6$ & $1.4$ & $86.1$ & $42.4$ & $82.7$ & $60.9$ & $67.9$ & $57.5$ & $48.2$
\\\cmidrule{3-23}
& & \textbf{LaserMix} & \cellcolor{blue!9}$58.7$ & $88.2$ & \cellcolor{blue!9}$37.1$ & \cellcolor{blue!9}$56.0$ & \cellcolor{blue!9}$80.9$ & \cellcolor{blue!9}$51.8$ & \cellcolor{blue!9}$70.8$ & \cellcolor{blue!9}$75.0$ & $0.0$ & $87.0$ & $31.8$ & $74.7$ & $0.8$ & \cellcolor{blue!9}$86.6$ & $41.3$ & \cellcolor{blue!9}$84.6$ & \cellcolor{blue!9}$62.1$ & \cellcolor{blue!9}$72.9$ & \cellcolor{blue!9}$59.8$ & \cellcolor{blue!9}$53.7$
\\\cmidrule{2-23}
& \multirow{6}{*}{\rotatebox{90}{Voxel}} & \textit{Sup.-only} & \cellcolor{yellow!12.5}$53.8$ & \cellcolor{yellow!12.5}$87.5$ & \cellcolor{yellow!12.5}$37.2$ & \cellcolor{yellow!12.5}$41.3$ & \cellcolor{yellow!12.5}$71.4$ & \cellcolor{yellow!12.5}$29.6$ & \cellcolor{yellow!12.5}$58.8$ & \cellcolor{yellow!12.5}$80.4$ & \cellcolor{yellow!12.5}$0.0$ & \cellcolor{yellow!12.5}$81.1$ & \cellcolor{yellow!12.5}$16.7$ & \cellcolor{yellow!12.5}$67.5$ & \cellcolor{yellow!12.5}$0.4$ & \cellcolor{yellow!12.5}$88.4$ & \cellcolor{yellow!12.5}$39.4$ & \cellcolor{yellow!12.5}$83.1$ & \cellcolor{yellow!12.5}$64.4$ & \cellcolor{yellow!12.5}$65.5$ & \cellcolor{yellow!12.5}$61.8$ & \cellcolor{yellow!12.5}$47.5$
\\\cmidrule{3-23}
& & MeanTeacher~\cite{MeanTeacher} & $53.9$ & $86.9$ & $33.6$ & $46.2$ & $48.9$ & $33.2$ & $62.8$ & $77.7$ & $0.0$ & $82.7$ & $22.8$ & $68.6$ & \cellcolor{blue!9}$3.2$ & $89.2$ & $38.6$ & $83.8$ & $66.4$ & $68.0$ & $62.3$ & $48.5$
\\
& & CBST~\cite{CBST} & $54.5$ & $87.6$ & $39.5$ & $36.7$ & $65.9$ & $35.7$ & $62.8$ & $78.1$ & $0.0$ & $82.4$ & $20.4$ & $69.6$ & $0.1$ & $88.8$ & \cellcolor{blue!9}$42.3$ & $84.2$ & $64.0$ & $67.4$ & $60.1$ & $50.1$
\\
& & CPS~\cite{CPS} & $54.8$ & $85.1$ & $35.2$ & $45.2$ & \cellcolor{blue!9}$68.6$ & $32.0$ & \cellcolor{blue!9}$65.7$ & $77.9$ & $0.2$ & $81.2$ & $21.7$ & $69.0$ & $1.6$ & $89.2$ & $40.2$ & $84.5$ & $65.1$ & $70.1$ & $60.9$ & $48.5$
\\\cmidrule{3-23}
& & \textbf{LaserMix} & \cellcolor{blue!9}$56.8$ & \cellcolor{blue!9}$88.0$ & \cellcolor{blue!9}$40.8$ & \cellcolor{blue!9}$51.6$ & $63.1$ & \cellcolor{blue!9}$38.4$ & $61.7$ & \cellcolor{blue!9}$79.9$ & \cellcolor{blue!9}$2.0$ & \cellcolor{blue!9}$83.1$ & \cellcolor{blue!9}$26.1$ & \cellcolor{blue!9}$71.2$ & $2.8$ & \cellcolor{blue!9}$90.1$ & $41.7$ & \cellcolor{blue!9}$85.9$ & \cellcolor{blue!9}$69.5$ & \cellcolor{blue!9}$70.5$ & \cellcolor{blue!9}$63.0$ & \cellcolor{blue!9}$51.6$
\\\bottomrule
\end{tabular}}
\label{table:class-ScribbleKitti-val}
\end{table*}

\newpage
\begin{table*}[t]
\caption{Class-wise IoU scores for \textbf{ranularity studies} on the \textbf{\textit{range view} representation} (under $10\%$ split on the \textit{val} set of nuScenes \cite{nuScenes}). All scores are given in percentage ($\%$). The best score for each semantic class is highlighted in \textbf{bold}.}
\vspace{-0.1cm}
\centering\scalebox{0.629}{
\begin{tabular}{c|c|c|cccccccccccccccc}
\toprule
Method & Illustr. & \rotatebox{0}{mIoU} & \rotatebox{0}{barr} & \rotatebox{0}{bicy} & \rotatebox{0}{bus} & \rotatebox{0}{car} & \rotatebox{0}{const} & \rotatebox{0}{moto} & \rotatebox{0}{ped} & \rotatebox{0}{cone} & \rotatebox{0}{trail} & \rotatebox{0}{truck} & \rotatebox{0}{driv} & \rotatebox{0}{othe} & \rotatebox{0}{walk} & \rotatebox{0}{terr} & \rotatebox{0}{manm} & \rotatebox{0}{veg}
\\\midrule\midrule
Baseline & \begin{minipage}[b]{0.138\columnwidth}\centering\raisebox{-.4\height}{\includegraphics[width=\linewidth]{figs/ablation-baseline.png}}\end{minipage} & $60.4$ & $69.0$ & $12.5$ & $67.0$ & $83.6$ & $27.2$ & $22.0$ & $63.7$ & $55.0$ & $40.4$ & $58.8$ & $95.0$ & $63.8$ & $67.2$ & $71.3$ & $85.6$ & $84.6$
\\\midrule
($1\alpha$, $2\phi$) & \begin{minipage}[b]{0.138\columnwidth}\centering\raisebox{-.4\height}{\includegraphics[width=\linewidth]{figs/ablation-col1row2.png}}\end{minipage} & $63.5$ & $70.8$ & $17.8$ & $65.3$ & $84.9$ & $26.9$ & $44.7$ & $65.8$ & $59.2$ & $\mathbf{46.6}$ & $62.2$ & $95.5$ & $64.3$ & $69.2$ & $72.5$ & $86.1$ & $84.9$
\\\midrule
($1\alpha$, $3\phi$) & \begin{minipage}[b]{0.138\columnwidth}\centering\raisebox{-.4\height}{\includegraphics[width=\linewidth]{figs/ablation-col1row3.png}}\end{minipage} & $65.2$ & $72.3$ & $21.5$ & $67.1$ & $85.1$ & $26.2$ & $57.1$ & $70.4$ & $59.3$ & $45.8$ & $60.7$ & $95.6$ & $65.4$ & $69.3$ & $73.7$ & $87.0$ & $85.9$
\\\midrule
($1\alpha$, $4\phi$) & \begin{minipage}[b]{0.138\columnwidth}\centering\raisebox{-.4\height}{\includegraphics[width=\linewidth]{figs/ablation-col1row4.png}}\end{minipage} & $66.5$ & $\textbf{73.7}$ & $22.4$ & $72.9$ & $87.0$ & $26.3$ & $59.4$ & $70.2$ & $60.3$ & $44.7$ & $\mathbf{64.7}$ & $\mathbf{95.8}$ & $67.8$ & $\mathbf{70.9}$ & $74.2$ & $87.0$ & $85.9$
\\\midrule
($1\alpha$, $5\phi$) & \begin{minipage}[b]{0.138\columnwidth}\centering\raisebox{-.4\height}{\includegraphics[width=\linewidth]{figs/ablation-col1row5.png}}\end{minipage} & $66.2$ & $72.8$ & $24.1$ & $74.0$ & $85.7$ & $\mathbf{36.3}$ & $47.8$ & $71.5$ & $60.8$ & $45.8$ & $64.5$ & $95.7$ & $64.8$ & $69.9$ & $73.3$ & $\mathbf{87.1}$ & $85.9$
\\\midrule
($1\alpha$, $6\phi$) & \begin{minipage}[b]{0.138\columnwidth}\centering\raisebox{-.4\height}{\includegraphics[width=\linewidth]{figs/ablation-col1row6.png}}\end{minipage} & $65.4$ & $72.6$ & $25.2$ & $69.8$ & $84.6$ & $33.8$ & $48.3$ & $70.1$ & $60.5$ & $44.8$ & $61.9$ & $95.4$ & $65.3$ & $68.9$ & $73.3$ & $86.8$ & $85.5$
\\\midrule
($2\alpha$, $1\phi$) & \begin{minipage}[b]{0.138\columnwidth}\centering\raisebox{-.4\height}{\includegraphics[width=\linewidth]{figs/ablation-col2row1.png}}\end{minipage} & $61.5$ & $68.4$ & $19.1$ & $67.1$ & $83.5$ & $28.1$ & $26.1$ & $64.8$ & $57.6$ & $41.5$ & $59.0$ & $95.1$ & $64.4$ & $68.1$ & $72.0$ & $85.4$ & $84.3$
\\\midrule
($2\alpha$, $2\phi$) & \begin{minipage}[b]{0.138\columnwidth}\centering\raisebox{-.4\height}{\includegraphics[width=\linewidth]{figs/ablation-col2row2.png}}\end{minipage} & $63.3$ & $70.8$ & $11.9$ & $64.8$ & $84.4$ & $27.3$ & $51.4$ & $69.2$ & $58.3$ & $41.8$ & $59.5$ & $95.6$ & $63.8$ & $69.8$ & $73.2$ & $86.2$ & $85.3$
\\\midrule
($2\alpha$, $3\phi$) & \begin{minipage}[b]{0.138\columnwidth}\centering\raisebox{-.4\height}{\includegraphics[width=\linewidth]{figs/ablation-col2row3.png}}\end{minipage} & $65.9$ & $71.9$ & $24.2$ & $69.1$ & $86.3$ & $28.2$ & $58.5$ & $71.5$ & $60.1$ & $44.7$ & $63.4$ & $95.7$ & $65.3$ & $70.1$ & $73.4$ & $86.9$ & $85.9$
\\\midrule
($2\alpha$, $4\phi$) & \begin{minipage}[b]{0.138\columnwidth}\centering\raisebox{-.4\height}{\includegraphics[width=\linewidth]{figs/ablation-col2row4.png}}\end{minipage} & $66.1$ & $72.9$ & $\mathbf{27.8}$ & $70.4$ & $86.4$ & $34.2$ & $54.1$ & $71.4$ & $61.5$ & $42.9$ & $61.3$ & $95.4$ & $64.6$ & $69.0$ & $73.3$ & $86.9$ & $85.8$
\\\midrule
($2\alpha$, $5\phi$) & \begin{minipage}[b]{0.138\columnwidth}\centering\raisebox{-.4\height}{\includegraphics[width=\linewidth]{figs/ablation-col2row5.png}}\end{minipage} & $\mathbf{66.7}$ & $73.1$ & $20.5$ & $69.6$ & $\mathbf{87.5}$ & $27.1$ & $67.8$ & $71.1$ & $61.0$ & $43.4$ & $64.6$ & $95.6$ & $\mathbf{69.1}$ & $70.1$ & $74.1$ & $87.0$ & $85.7$
\\\midrule
($2\alpha$, $6\phi$) & \begin{minipage}[b]{0.138\columnwidth}\centering\raisebox{-.4\height}{\includegraphics[width=\linewidth]{figs/ablation-col2row6.png}}\end{minipage} & $65.3$ & $71.9$ & $23.5$ & $68.6$ & $85.3$ & $32.2$ & $51.0$ & $69.6$ & $60.2$ & $45.2$ & $63.6$ & $95.4$ & $63.6$ & $69.2$ & $73.3$ & $86.4$ & $85.1$
\\\midrule
($3\alpha$, $1\phi$) & \begin{minipage}[b]{0.138\columnwidth}\centering\raisebox{-.4\height}{\includegraphics[width=\linewidth]{figs/ablation-col3row1.png}}\end{minipage} & $60.9$ & $67.4$ & $14.7$ & $65.3$ & $82.9$ & $25.2$ & $39.5$ & $63.6$ & $57.0$ & $34.7$ & $55.2$ & $95.0$ & $64.6$ & $67.1$ & $71.4$ & $85.6$ & $84.7$
\\\midrule
($3\alpha$, $2\phi$) & \begin{minipage}[b]{0.138\columnwidth}\centering\raisebox{-.4\height}{\includegraphics[width=\linewidth]{figs/ablation-col3row2.png}}\end{minipage} & $64.2$ & $71.4$ & $15.5$ & $66.3$ & $86.2$ & $25.8$ & $54.4$ & $68.4$ & $60.4$ & $44.3$ & $63.2$ & $95.5$ & $63.3$ & $69.3$ & $72.6$ & $86.3$ & $85.1$
\\\midrule
($3\alpha$, $3\phi$) & \begin{minipage}[b]{0.138\columnwidth}\centering\raisebox{-.4\height}{\includegraphics[width=\linewidth]{figs/ablation-col3row3.png}}\end{minipage} & $65.9$ & $73.1$ & $15.7$ & $\mathbf{74.4}$ & $85.3$ & $28.0$ & $59.7$ & $69.0$ & $60.6$ & $45.3$ & $63.4$ & $95.6$ & $68.8$ & $70.5$ & $\mathbf{74.3}$ & $86.5$ & $85.2$
\\\midrule
($3\alpha$, $4\phi$) & \begin{minipage}[b]{0.138\columnwidth}\centering\raisebox{-.4\height}{\includegraphics[width=\linewidth]{figs/ablation-col3row4.png}}\end{minipage} & $66.3$ & $72.1$ & $16.7$ & $73.1$ & $86.2$ & $34.2$ & $66.6$ & $67.3$ & $58.4$ & $45.5$ & $62.4$ & $95.7$ & $65.6$ & $70.5$ & $74.0$ & $86.7$ & $85.6$
\\\midrule
($3\alpha$, $5\phi$) & \begin{minipage}[b]{0.138\columnwidth}\centering\raisebox{-.4\height}{\includegraphics[width=\linewidth]{figs/ablation-col3row5.png}}\end{minipage} & $66.0$ & $72.1$ & $18.4$ & $72.7$ & $86.0$ & $31.9$ & $\mathbf{68.8}$ & $66.0$ & $58.3$ & $42.9$ & $61.9$ & $95.6$ & $65.7$ & $70.0$ & $73.9$ & $86.5$ & $85.4$
\\\midrule
($3\alpha$, $6\phi$) & \begin{minipage}[b]{0.138\columnwidth}\centering\raisebox{-.4\height}{\includegraphics[width=\linewidth]{figs/ablation-col3row6.png}}\end{minipage} & $65.2$ & $71.0$ & $26.4$ & $68.6$ & $85.6$ & $29.3$ & $54.7$ & $69.1$ & $59.6$ & $43.5$ & $62.1$ & $95.3$ & $63.9$ & $68.7$ & $73.4$ & $86.3$ & $85.1$
\\\midrule
($4\alpha$, $1\phi$) & \begin{minipage}[b]{0.138\columnwidth}\centering\raisebox{-.4\height}{\includegraphics[width=\linewidth]{figs/ablation-col4row1.png}}\end{minipage} & $60.9$ & $69.1$ & $16.7$ & $66.2$ & $83.7$ & $23.5$ & $26.8$ & $64.4$ & $56.2$ & $40.6$ & $60.4$ & $95.0$ & $63.7$ & $67.0$ & $71.5$ & $85.6$ & $84.6$
\\\midrule
($4\alpha$, $2\phi$) & \begin{minipage}[b]{0.138\columnwidth}\centering\raisebox{-.4\height}{\includegraphics[width=\linewidth]{figs/ablation-col4row2.png}}\end{minipage} & $64.7$ & $71.0$ & $16.7$ & $66.4$ & $84.7$ & $28.4$ & $59.4$ & $70.6$ & $60.7$ & $42.1$ & $60.3$ & $95.4$ & $65.4$ & $69.6$ & $73.2$ & $86.4$ & $85.5$
\\\midrule
($4\alpha$, $3\phi$) & \begin{minipage}[b]{0.138\columnwidth}\centering\raisebox{-.4\height}{\includegraphics[width=\linewidth]{figs/ablation-col4row3.png}}\end{minipage} & $65.3$ & $72.0$ & $24.9$ & $68.5$ & $85.6$ & $25.1$ & $52.7$ & $71.3$ & $59.3$ & $45.2$ & $62.3$ & $95.5$ & $66.9$ & $69.6$ & $73.1$ & $86.8$ & $85.8$ 
\\\midrule
($4\alpha$, $4\phi$) & \begin{minipage}[b]{0.138\columnwidth}\centering\raisebox{-.4\height}{\includegraphics[width=\linewidth]{figs/ablation-col4row4.png}}\end{minipage} & $65.6$ & $71.9$ & $21.3$ & $71.3$ & $86.4$ & $32.8$ & $44.0$ & $69.3$ & $\mathbf{61.8}$ & $45.5$ & $64.0$ & $95.7$ & $67.7$ & $70.2$ & $73.9$ & $\mathbf{87.1}$ & $\mathbf{86.2}$
\\\midrule
($4\alpha$, $5\phi$) & \begin{minipage}[b]{0.138\columnwidth}\centering\raisebox{-.4\height}{\includegraphics[width=\linewidth]{figs/ablation-col4row5.png}}\end{minipage} & $65.7$ & $71.9$ & $24.8$ & $72.1$ & $84.5$ & $31.6$ & $52.2$ & $\mathbf{71.6}$ & $60.6$ & $43.6$ & $60.4$ & $95.5$ & $65.1$ & $70.1$ & $73.8$ & $\mathbf{87.1}$ & $86.1$
\\\midrule
($4\alpha$, $6\phi$) & \begin{minipage}[b]{0.138\columnwidth}\centering\raisebox{-.4\height}{\includegraphics[width=\linewidth]{figs/ablation-col4row6.png}}\end{minipage} & $65.2$ & $72.1$ & $18.8$ & $69.9$ & $83.7$ & $32.5$ & $57.2$ & $69.0$ & $60.2$ & $43.2$ & $59.1$ & $95.5$ & $66.1$ & $69.8$ & $73.5$ & $87.0$ & $86.1$
\\\midrule
\end{tabular}}
\label{table:class-ablation-inclination-rv}
\end{table*}

\newpage
\begin{table*}[t]
\caption{Class-wise IoU scores for \textbf{granularity studies} on the \textbf{\textit{voxel} representation} (under $10\%$ split on the \textit{val} set of nuScenes \cite{nuScenes}). All scores are given in percentage ($\%$). The best score for each semantic class is highlighted in \textbf{bold}.}
\vspace{-0.1cm}
\centering\scalebox{0.629}{
\begin{tabular}{c|c|c|cccccccccccccccc}
\toprule
Method & Illustr. & \rotatebox{0}{mIoU} & \rotatebox{0}{barr} & \rotatebox{0}{bicy} & \rotatebox{0}{bus} & \rotatebox{0}{car} & \rotatebox{0}{const} & \rotatebox{0}{moto} & \rotatebox{0}{ped} & \rotatebox{0}{cone} & \rotatebox{0}{trail} & \rotatebox{0}{truck} & \rotatebox{0}{driv} & \rotatebox{0}{othe} & \rotatebox{0}{walk} & \rotatebox{0}{terr} & \rotatebox{0}{manm} & \rotatebox{0}{veg}
\\\midrule\midrule
Baseline & \begin{minipage}[b]{0.138\columnwidth}\centering\raisebox{-.4\height}{\includegraphics[width=\linewidth]{figs/ablation-baseline.png}}\end{minipage} & $66.0$ & $71.1$ & $19.7$ & $85.1$ & $83.3$ & $42.0$ & $43.5$ & $64.0$ & $54.9$ & $45.6$ & $73.7$ & $95.3$ & $66.8$ & $69.8$ & $69.6$ & $86.7$ & $84.9$
\\\midrule
($1\alpha$, $2\phi$) &  \begin{minipage}[b]{0.138\columnwidth}\centering\raisebox{-.4\height}{\includegraphics[width=\linewidth]{figs/ablation-col1row2.png}}\end{minipage} & $68.7$ & $\mathbf{72.4}$ & $20.7$ & $87.0$ & $83.1$ & $38.3$ & $66.7$ & $65.8$ & $\mathbf{57.5}$ & $54.9$ & $76.6$ & $95.4$ & $67.1$ & $70.3$ & $71.6$ & $86.9$ & $84.6$
\\\midrule
($1\alpha$, $3\phi$) & \begin{minipage}[b]{0.138\columnwidth}\centering\raisebox{-.4\height}{\includegraphics[width=\linewidth]{figs/ablation-col1row3.png}}\end{minipage} & $69.0$ & $71.5$ & $21.5$ & $87.0$ & $83.7$ & $39.5$ & $68.1$ & $66.1$ & $\mathbf{57.5}$ & $56.6$ & $\mathbf{77.2}$ & $95.5$ & $66.5$ & $70.7$ & $71.9$ & $86.9$ & $84.6$
\\\midrule
($1\alpha$, $4\phi$) & \begin{minipage}[b]{0.138\columnwidth}\centering\raisebox{-.4\height}{\includegraphics[width=\linewidth]{figs/ablation-col1row4.png}}\end{minipage} & $69.4$ & $71.8$ & $24.1$ & $87.2$ & $84.5$ & $40.8$ & $69.6$ & $66.7$ & $57.1$ & $55.4$ & $76.1$ & $95.5$ & $\mathbf{67.2}$ & $70.7$ & $70.9$ & $86.7$ & $85.4$
\\\midrule
($1\alpha$, $5\phi$) & \begin{minipage}[b]{0.138\columnwidth}\centering\raisebox{-.4\height}{\includegraphics[width=\linewidth]{figs/ablation-col1row5.png}}\end{minipage} & $69.6$ & $71.1$ & $24.4$ & $\mathbf{88.1}$ & $83.0$ & $\mathbf{42.1}$ & $72.2$ & $66.4$ & $57.4$ & $57.7$ & $75.2$ & $95.4$ & $\mathbf{67.2}$ & $70.5$ & $70.7$ & $86.9$ & $85.5$
\\\midrule
($1\alpha$, $6\phi$) & \begin{minipage}[b]{0.138\columnwidth}\centering\raisebox{-.4\height}{\includegraphics[width=\linewidth]{figs/ablation-col1row6.png}}\end{minipage} & $69.3$ & $70.3$ & $23.1$ & $87.3$ & $83.5$ & $38.6$ & $74.1$ & $65.8$ & $56.5$ & $57.2$ & $76.7$ & $95.5$ & $65.7$ & $70.8$ & $71.0$ & $86.9$ & $85.7$
\\\midrule
($2\alpha$, $1\phi$) & \begin{minipage}[b]{0.138\columnwidth}\centering\raisebox{-.4\height}{\includegraphics[width=\linewidth]{figs/ablation-col2row1.png}}\end{minipage} & $67.2$ & $70.0$ & $19.7$ & $84.3$ & $86.3$ & $39.6$ & $65.7$ & $62.6$ & $52.1$ & $50.7$ & $73.4$ & $95.2$ & $64.4$ & $69.2$ & $71.4$ & $86.6$ & $84.6$
\\\midrule
($2\alpha$, $2\phi$) & \begin{minipage}[b]{0.138\columnwidth}\centering\raisebox{-.4\height}{\includegraphics[width=\linewidth]{figs/ablation-col2row2.png}}\end{minipage} & $67.7$ & $70.7$ & $\mathbf{25.9}$ & $84.7$ & $84.7$ & $37.4$ & $65.3$ & $63.5$ & $52.6$ & $53.6$ & $71.9$ & $95.3$ & $65.8$ & $69.8$ & $71.6$ & $86.6$ & $84.7$
\\\midrule
($2\alpha$, $3\phi$) & \begin{minipage}[b]{0.138\columnwidth}\centering\raisebox{-.4\height}{\includegraphics[width=\linewidth]{figs/ablation-col2row3.png}}\end{minipage} & $68.5$ & $70.3$ & $19.8$ & $86.2$ & $86.2$ & $38.1$ & $70.9$ & $64.0$ & $55.7$ & $55.7$ & $74.7$ & $95.3$ & $65.9$ & $69.8$ & $\mathbf{72.0}$ & $87.0$ & $84.7$
\\\midrule
($2\alpha$, $4\phi$) & \begin{minipage}[b]{0.138\columnwidth}\centering\raisebox{-.4\height}{\includegraphics[width=\linewidth]{figs/ablation-col2row4.png}}\end{minipage} & $69.6$ & $72.3$ & $24.2$ & $86.3$ & $85.0$ & $41.8$ & $72.1$ & $66.2$ & $56.6$ & $56.5$ & $76.4$ & $95.4$ & $66.7$ & $70.6$ & $71.4$ & $86.9$ & $85.5$
\\\midrule
($2\alpha$, $5\phi$) & \begin{minipage}[b]{0.138\columnwidth}\centering\raisebox{-.4\height}{\includegraphics[width=\linewidth]{figs/ablation-col2row5.png}}\end{minipage} & $69.3$ & $\mathbf{72.4}$ & $21.7$ & $86.3$ & $84.8$ & $40.5$ & $68.7$ & $66.9$ & $56.6$ & $\mathbf{58.0}$ & $76.8$ & $95.3$ & $67.0$ & $70.3$ & $71.5$ & $86.9$ & $85.5$
\\\midrule
($2\alpha$, $6\phi$) & \begin{minipage}[b]{0.138\columnwidth}\centering\raisebox{-.4\height}{\includegraphics[width=\linewidth]{figs/ablation-col2row6.png}}\end{minipage} & $69.1$ & $71.9$ & $20.5$ & $87.7$ & $84.2$ & $41.6$ & $69.3$ & $66.5$ & $56.9$ & $55.0$ & $76.5$ & $95.4$ & $66.4$ & $70.7$ & $71.1$ & $86.9$ & $85.5$
\\\midrule
($3\alpha$, $1\phi$) & \begin{minipage}[b]{0.138\columnwidth}\centering\raisebox{-.4\height}{\includegraphics[width=\linewidth]{figs/ablation-col3row1.png}}\end{minipage} & $67.3$ & $70.1$ & $20.7$ & $82.8$ & $86.3$ & $34.4$ & $66.5$ & $62.5$ & $53.6$ & $55.2$ & $74.5$ & $95.1$ & $64.4$ & $69.1$ & $71.4$ & $86.6$ & $84.4$
\\\midrule
($3\alpha$, $2\phi$) & \begin{minipage}[b]{0.138\columnwidth}\centering\raisebox{-.4\height}{\includegraphics[width=\linewidth]{figs/ablation-col3row2.png}}\end{minipage} & $68.3$ & $71.9$ & $16.6$ & $85.9$ & $83.0$ & $39.5$ & $66.3$ & $66.0$ & $57.0$ & $56.6$ & $76.3$ & $95.4$ & $64.3$ & $70.5$ & $71.7$ & $86.7$ & $84.6$
\\\midrule
($3\alpha$, $3\phi$) & \begin{minipage}[b]{0.138\columnwidth}\centering\raisebox{-.4\height}{\includegraphics[width=\linewidth]{figs/ablation-col3row3.png}}\end{minipage} & $68.6$ & $70.9$ & $20.2$ & $86.7$ & $86.4$ & $39.1$ & $68.0$ & $66.2$ & $56.7$ & $53.3$ & $74.5$ & $95.3$ & $66.2$ & $70.0$ & $71.8$ & $87.1$ & $85.0$
\\\midrule
($3\alpha$, $4\phi$) & \begin{minipage}[b]{0.138\columnwidth}\centering\raisebox{-.4\height}{\includegraphics[width=\linewidth]{figs/ablation-col3row4.png}}\end{minipage} & $68.5$ & $70.6$ & $19.9$ & $86.7$ & $86.1$ & $39.3$ & $67.6$ & $66.2$ & $56.8$ & $54.1$ & $74.2$ & $95.3$ & $65.6$ & $70.1$ & $71.9$ & $87.1$ & $84.9$
\\\midrule
($3\alpha$, $5\phi$) & \begin{minipage}[b]{0.138\columnwidth}\centering\raisebox{-.4\height}{\includegraphics[width=\linewidth]{figs/ablation-col3row5.png}}\end{minipage} & $\mathbf{69.8}$ & $72.2$ & $24.3$ & $87.3$ & $84.8$ & $41.2$ & $73.6$ & $\mathbf{67.0}$ & $57.1$ & $55.9$ & $76.9$ & $95.4$ & $67.1$ & $70.4$ & $71.2$ & $86.9$ & $85.6$
\\\midrule
($3\alpha$, $6\phi$) & \begin{minipage}[b]{0.138\columnwidth}\centering\raisebox{-.4\height}{\includegraphics[width=\linewidth]{figs/ablation-col3row6.png}}\end{minipage} & $69.1$ & $71.9$ & $20.5$ & $87.7$ & $84.2$ & $41.6$ & $69.3$ & $66.5$ & $56.9$ & $55.0$ & $76.5$ & $95.4$ & $66.4$ & $70.7$ & $71.1$ & $86.9$ & $85.5$
\\\midrule
($4\alpha$, $1\phi$) & \begin{minipage}[b]{0.138\columnwidth}\centering\raisebox{-.4\height}{\includegraphics[width=\linewidth]{figs/ablation-col4row1.png}}\end{minipage} & $66.7$ & $70.5$ & $15.4$ & $85.1$ & $86.5$ & $35.2$ & $67.5$ & $62.5$ & $51.3$ & $51.0$ & $72.9$ & $95.1$ & $63.9$ & $68.5$ & $71.6$ & $86.2$ & $84.5$
\\\midrule
($4\alpha$, $2\phi$) & \begin{minipage}[b]{0.138\columnwidth}\centering\raisebox{-.4\height}{\includegraphics[width=\linewidth]{figs/ablation-col4row2.png}}\end{minipage} & $68.4$ & $70.9$ & $18.6$ & $86.0$ & $85.3$ & $39.5$ & $69.9$ & $65.6$ & $56.4$ & $54.2$ & $73.6$ & $95.4$ & $65.5$ & $70.0$ & $71.7$ & $87.0$ & $84.8$
\\\midrule
($4\alpha$, $3\phi$) & \begin{minipage}[b]{0.138\columnwidth}\centering\raisebox{-.4\height}{\includegraphics[width=\linewidth]{figs/ablation-col4row3.png}}\end{minipage} & $68.7$ & $70.3$ & $22.1$ & $86.5$ & $\mathbf{86.6}$ & $39.6$ & $67.8$ & $66.0$ & $57.1$ & $53.2$ & $75.4$ & $95.3$ & $66.0$ & $69.9$ & $71.9$ & $\mathbf{87.2}$ & $85.0$
\\\midrule
($4\alpha$, $4\phi$) & \begin{minipage}[b]{0.138\columnwidth}\centering\raisebox{-.4\height}{\includegraphics[width=\linewidth]{figs/ablation-col4row4.png}}\end{minipage} & $68.9$ & $71.5$ & $23.8$ & $86.9$ & $83.1$ & $38.6$ & $\mathbf{74.5}$ & $65.8$ & $56.5$ & $50.8$ & $75.1$ & $\mathbf{95.6}$ & $65.8$ & $70.8$ & $70.4$ & $87.1$ & $85.6$
\\\midrule
($4\alpha$, $5\phi$) & \begin{minipage}[b]{0.138\columnwidth}\centering\raisebox{-.4\height}{\includegraphics[width=\linewidth]{figs/ablation-col4row5.png}}\end{minipage} & $69.0$ & $71.0$ & $23.3$ & $87.7$ & $83.5$ & $40.3$ & $73.7$ & $66.1$ & $56.9$ & $51.6$ & $75.1$ & $\mathbf{95.6}$ & $65.3$ & $70.7$ & $70.4$ & $87.1$ & $85.6$
\\\midrule
($4\alpha$, $6\phi$) & \begin{minipage}[b]{0.138\columnwidth}\centering\raisebox{-.4\height}{\includegraphics[width=\linewidth]{figs/ablation-col4row6.png}}\end{minipage} & $69.4$ & $70.9$ & $25.2$ & $87.9$ & $83.5$ & $40.8$ & $73.0$ & $66.5$ & $57.2$ & $54.0$ & $76.4$ & $\mathbf{95.6}$ & $65.1$ & $\mathbf{70.9}$ & $70.6$ & $87.1$ & $\mathbf{85.8}$
\\\midrule
\end{tabular}}
\label{table:class-ablation-inclination-voxel}
\end{table*}

\newpage

\newpage

\newpage

\newpage

\newpage

\newpage

\clearpage
{\small
\bibliographystyle{ieee_fullname}
\bibliography{egbib}
}

\end{document}